\documentclass{article}

\PassOptionsToPackage{numbers,square,sort&compress}{natbib}

\usepackage[final]{neurips_2022}

\usepackage[utf8]{inputenc} 
\usepackage[T1]{fontenc}    
\usepackage{amsmath}
\usepackage{wrapfig}
\usepackage{graphicx}
\usepackage{subfig}
\usepackage{caption}
\usepackage{hyperref}       
\usepackage{cleveref}       
\usepackage{url}            
\usepackage{booktabs}       
\usepackage{amsfonts}       
\usepackage{nicefrac}       
\usepackage{microtype}      
\usepackage[dvipsnames]{xcolor}         
\usepackage{placeins}
\usepackage{chemformula}
\usepackage{siunitx}
\usepackage{todonotes}

\definecolor{AltairBlue}{HTML}{1f77b4}
\definecolor{AltairOrange}{HTML}{ff7f0e}

\DeclareMathOperator*{\argmin}{arg\,min}

\crefname{section}{\S}{\S}

\graphicspath{{fig/}}

\title{Modeling the Machine Learning Multiverse}

\author{%
    Samuel J.~Bell$^1$ \quad Onno P.~Kampman$^2$ \quad Jesse Dodge$^3$ \quad Neil D.~Lawrence$^1$ \\
    $^1$Computer Laboratory, University of Cambridge \\ $^2$Department of Psychology, University of Cambridge \\ $^3$Allen Institute for AI \\
    \texttt{\{sjb326,opk20,ndl21\}@cam.ac.uk} \quad \texttt{jessed@allenai.org}
}

\begin{document}

\maketitle

\begin{abstract}
Amid mounting concern about the reliability and credibility of machine learning research,
we present a principled framework for making robust and generalizable claims: the \emph{multiverse analysis}.
Our framework builds upon the multiverse analysis \citep{Steegen2016} introduced in response to psychology's own reproducibility crisis.
To efficiently explore high-dimensional and often continuous ML search spaces, we \emph{model} the multiverse with a Gaussian Process surrogate and apply Bayesian experimental design.
Our framework is designed to facilitate drawing robust scientific conclusions about model performance, and thus our approach focuses on exploration rather than conventional optimization.
In the first of two case studies, we investigate disputed claims about the relative merit of adaptive optimizers. 
Second, we synthesize conflicting research on the effect of learning rate on the large batch training generalization gap.
For the machine learning community, a multiverse analysis is a simple and effective technique for identifying robust claims, for increasing transparency, and a step toward improved reproducibility.
\end{abstract}

\section{Introduction}\label{sec:introduction}

Machine learning research faces mounting concern about the reliability of our results and the credibility of our claims \citep{Gundersen2018, Bouthillier2019, Lipton2019, Sculley2018, Cooper2021, Raff2019, Forde2019, Rahimi2017, Henderson2017, Agarwal2021, Lucic2018, Dacrema2019, Schmidt2021, Oakden-Rayner2019, Melis2018, Reimers2017, Gorman2019, Marie2021, Narang2021}.
The field of psychology has faced a similar crisis \citep{Bell2021}, and confrontation with its shortcomings has sparked practical innovations \citep{John2012, Wagenmakers2012, Chambers2013, Klein2014, Steegen2016}.
One such innovation is of direct relevance to the machine learning community: the \emph{multiverse analysis} \citep{Steegen2016}.

Throughout any investigation, scientists make decisions about how to perform their work.
In psychology, as in other disciplines, there are a plethora of different ways to conduct and analyze experiments.
From just a handful of choices, we reach such a large decision tree that researchers can repeatedly try different paths until they chance upon a positive result \citep{Simmons2011}.
Even without conscious manipulation, these decision points pose a fundamental problem: what if the psychologist had chosen an alternate---and perfectly reasonable---route through this garden of forking paths \citep{Gelman2014}?
Would their results and conclusions still stand?

In their replication of a contentious study \citep{Durante2013} on the effect of menstrual cycle and relationship status on women's political preferences, \citet{Steegen2016} identify both decisions taken and reasonable alternatives.\footnote{E.g.\ outlier selection; variable discretization; and how to estimate menstrual onset.}
They name the Cartesian product of alternatives a \emph{multiverse}, a set of parallel universes each containing a slightly different study.
Introducing the \emph{multiverse analysis}, \citeauthor{Steegen2016} re-run the study as if inside each universe, finding that the alleged effect of relationship status and fertility is highly sensitive to different choices, in particular the definition of ``single''.
Multiverse analyses have subsequently been used to evaluate the robustness of claims across a variety of psychological and neuroscientific settings (e.g.\ \citep{Donnelly2019, Kalokerinos2019, Modecki2020,Lonsdorf2022,Dafflon2022}).

In a machine learning context, we present the multiverse analysis as a principled framework for analyzing robustness and generality.
Consider an example of some modification to a model, say batch normalization \citep{Ioffe2015}.
To verify batch norm's efficacy, one needs a test bed including model architecture, optimization method, dataset, evaluation metric, and so on.
Regardless of these specific choices, we would like that batch norm be effective in general.
With a multiverse analysis, we can systematically explore the effect of each choice, and understand the circumstances in which a claim holds true.

Our primary contribution is to introduce the multiverse analysis to ML, which we use to draw more robust conclusions about model performance.
To efficiently explore the high-dimensional and often continuous ML search space, we model the multiverse with a Gaussian Process (GP) surrogate and use Bayesian experimental design (\cref{sec:intro-methods}).
We present motivating evidence that choosing exploration over optimization---the essence of a multiverse analysis---is essential when we seek proper understanding of our claims and their generality (\cref{sec:motivating-example}).
In the first of two case studies, we use a multiverse analysis over hyperparameters to replicate an experiment on adaptive optimizers~\citep{Wilson2017}, finding that claims on the relative merits of optimizers are highly sensitive to learning rate (\cref{sec:case-study-1}).
Second, we perform a more extensive multiverse analysis to synthesize divergent research on the large batch ``generalization gap''~\citep{Keskar2017}, finding an interaction effect of batch size and learning rate (\cref{sec:case-study-2}).
We conclude by discussing the limitations of our approach and directions for future work (\cref{sec:discussion}).

\section{Efficient multiverse exploration}\label{sec:intro-methods}

Our approach to the multiverse analysis packages ML researcher decisions into a simple framework, requiring only a choice of \textbf{search space} \(\mathcal{X}\) and an \textbf{evaluation function} \(\ell\), which together define our multiverse.
This search space defines the set of \emph{reasonable} choices.
For example, we might consider only a few directly relevant hyperparameters (\cref{sec:case-study-1}); include purportedly irrelevant choices such as dataset (\cref{sec:case-study-2}); or conduct even more expansive analyses (\cref{sec:discussion}).
The evaluation function should pertain to the hypothesis being tested.
Here, we define it as model test accuracy (\cref{sec:case-study-2}) or difference in test accuracy (\cref{sec:case-study-1}).

A barrier to analyzing any ML multiverse is the size of the search space and the presence of continuous dimensions.
This makes exhaustive search intractable.
To overcome this, we use a GP surrogate---modeling \(\ell\) as a function of \(\mathcal{X}\)---as a stand-in for the multiverse.
Using the surrogate, we iteratively explore the space using Bayesian experimental design \citep{Chaloner1995}:
\begin{enumerate}
    \item Sample an initial design, \(X_0 \sim \mathcal{X}\), and evaluate \(\ell\) at each point, \(Y_0 = \ell(X_0)\);
    \item Fit a GP model \(f\) to the sampled points \(X_0\) and corresponding results \(Y_0\);
    \item Use an acquisition function \(a\) on \(f\) to sample and evaluate a new batch (\(X_i, Y_i\));
    \item Repeat steps 2--3 until an appropriate stopping criterion.
\end{enumerate}
In step 1, our initial design is drawn from a Sobol sequence, a low discrepancy sequence that achieves improved coverage over uniform random sampling in higher dimensions \citep{Bergstra2012}.
The search space and the evaluation function are specific to the analysis at hand and described later.

In step 2, we model the output of our evaluation function as a noisy function of the inputs,
\begin{align}
    \mathbf{y}_i &= f(\mathbf{x}_i) + \epsilon_i \, , \quad  \epsilon_i \sim \mathcal{N}(0, \Sigma) \, .
\end{align}
Placing a GP prior over $f$ with zero mean and positive definite kernel function $k$,
\begin{align}
    f &\sim \text{GP}(0, k) \, ,
\end{align}
we obtain the posterior mean $\mu$ and variance $\sigma^{2}$ \citep[chap.~2]{Williams2006}:
\begin{align}
    \mathbf{y}_{i+1} &\sim \mathcal{N}(\mu(\mathbf{x}_{i+1}), \sigma^{2}(\mathbf{x}_{i+1})) \, , \\
    \mu(\mathbf{x}_{i+1}) &= \mathbf{k}^T K^{-1} Y_{i} \, , \\ 
    \sigma^{2}(\mathbf{x}_{i+1}) &= k(\mathbf{x}_{i+1}, \mathbf{x}_{i+1}) - \mathbf{k}^T (K + \Sigma)^{-1} \mathbf{k} \, ,
\end{align}
where $K$ is the kernel matrix and $\mathbf{k} = [k(\mathbf{x}_{i+1}, \mathbf{x}_{0}), \ldots, k(\mathbf{x}_{i+1}, \mathbf{x}_{i})]$.
We use a Mat\'{e}rn 5/2 kernel with automatic relevance determination \citep{Snoek2012}.

In step 3, we use integrated variance reduction (IVR) as our acquisition function \citep{Sacks1989} (motivated in~\cref{sec:motivating-example}).
IVR calculates the change in \emph{total} variance if we were to acquire a candidate point \(\textbf{x}_{i+1}\)
\begin{align}
    a(\textbf{x}_{i+1}; X_i, Y_i) = \int_{\mathcal{X}} \sigma^{2}(\mathbf{p}; X_{i+1}, Y_{i+1}) - \sigma^{2}(\mathbf{p}; X_{i}, Y_{i}) \,d\mathbf{p} \, ,
\end{align}
where \(X_i, Y_i\) excludes the candidate point and \(X_{i+1}, Y_{i+1}\) includes it.

Using a Monte Carlo approximation\footnote{In particularly high-dimensional search spaces, it may be more appropriate to use a quasi-Monte Carlo method to estimate the variance reduction over a sample of points with improved coverage of the space.} for the intractable integral, the most informative point \(\mathbf{x}*\) to sample next is
\begin{align}
    \mathbf{x}* = \argmin_{\mathbf{x}_{i+1}} a(\mathbf{x}_{i+1}; X_i, Y_i) \, .
\end{align}


A key difference between standard hyperparameter optimization and our framework is that when optimizing, all models except the best performing are discarded, whereas we use all available information to draw more robust conclusions.
To do this, we continue to use our surrogate model.
First, we visualize the multiverse using the surrogate's posterior predictive mean, and visualize the extent to which we have explored through the surrogate's posterior variance.
Second, we test for interaction effects by comparing a GP with a shared kernel \(M_{\textrm{shared}}\) against an additive kernel \(M_{\textrm{additive}}\) (see \cref{fig:kernel-examples}).
We compare their fit using the Bayes factor \(K\),
\begin{equation}\label{eq:bayes-factor}
	K = \frac{P(X,Y|M_{\textrm{additive}})}{P(X,Y|M_{\textrm{shared}})} \, ,
\end{equation}
where \(P(X,Y|M)\) is the marginal likelihood of the observations given a model.
\(K > 1\) indicates that \(M_{\textrm{additive}}\) better explains the data than \(M_{\textrm{shared}}\), indicating the absence of an interaction.
This also provides an appropriate termination condition for step 4: we continue to sample until we reach a conclusive Bayes factor.
Finally, we perform a Monte Carlo sensitivity analysis \citep{Sobol2001, Saltelli2002} to assess how much a change in one of the parameters would affect the model outputs.

For GP modeling we use GPy~\citep{GPy2012} with EmuKit \citep{Emukit2019} for experimental design and sensitivity analysis. We use TorchVision's~\citep{Paszke2017} off-the-shelf deep learning model architectures.

\section{Motivating example: SVM hyperparameters}\label{sec:motivating-example}

\begin{figure}[t]
    \centering
    \subfloat[UCB]{
        \includegraphics[trim={0 0 0 0.3cm},clip,width=0.33\textwidth]{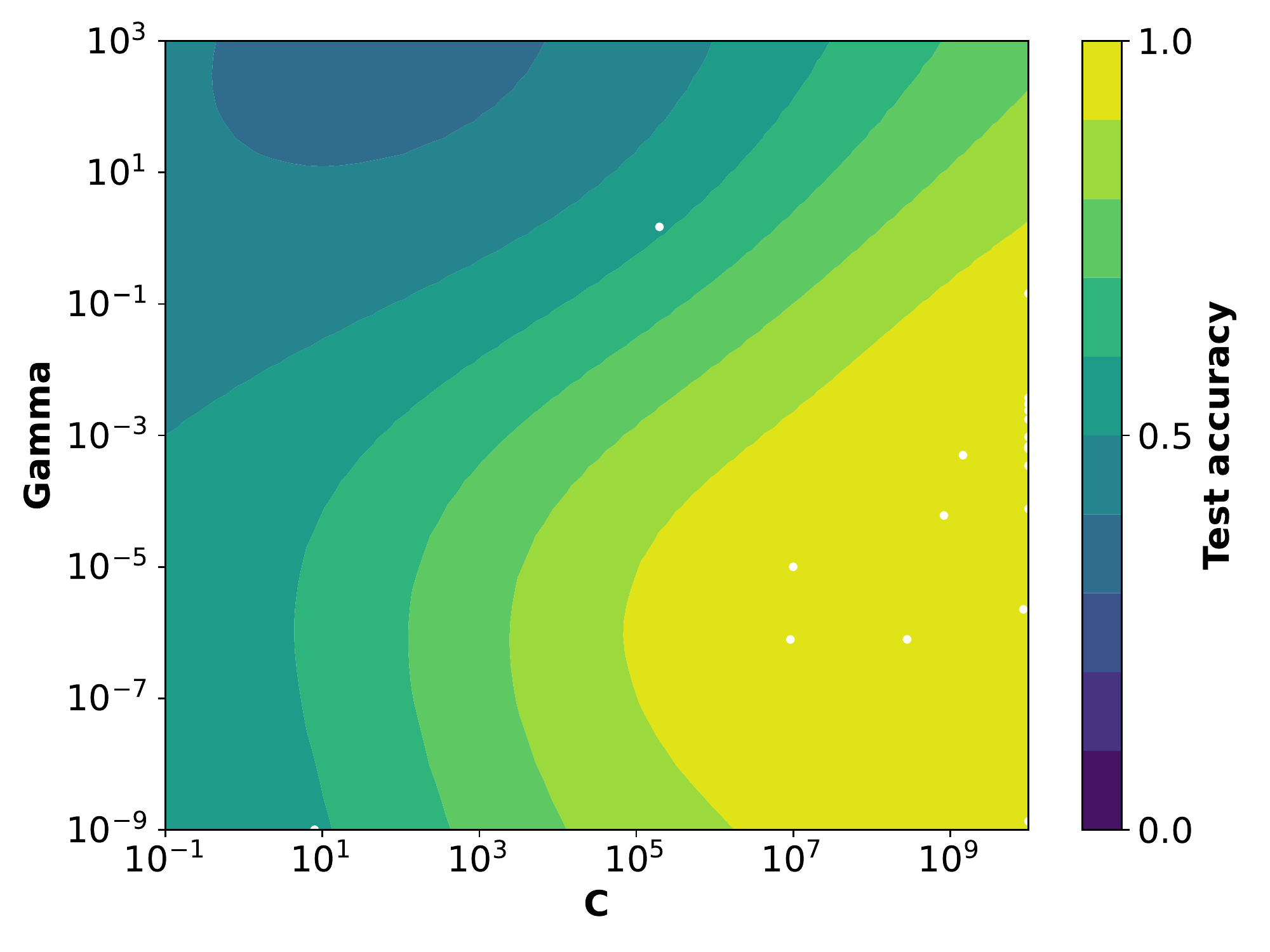}
    }
    \subfloat[IVR]{
        \includegraphics[trim={0 0 0 0.3cm},clip,width=0.33\textwidth]{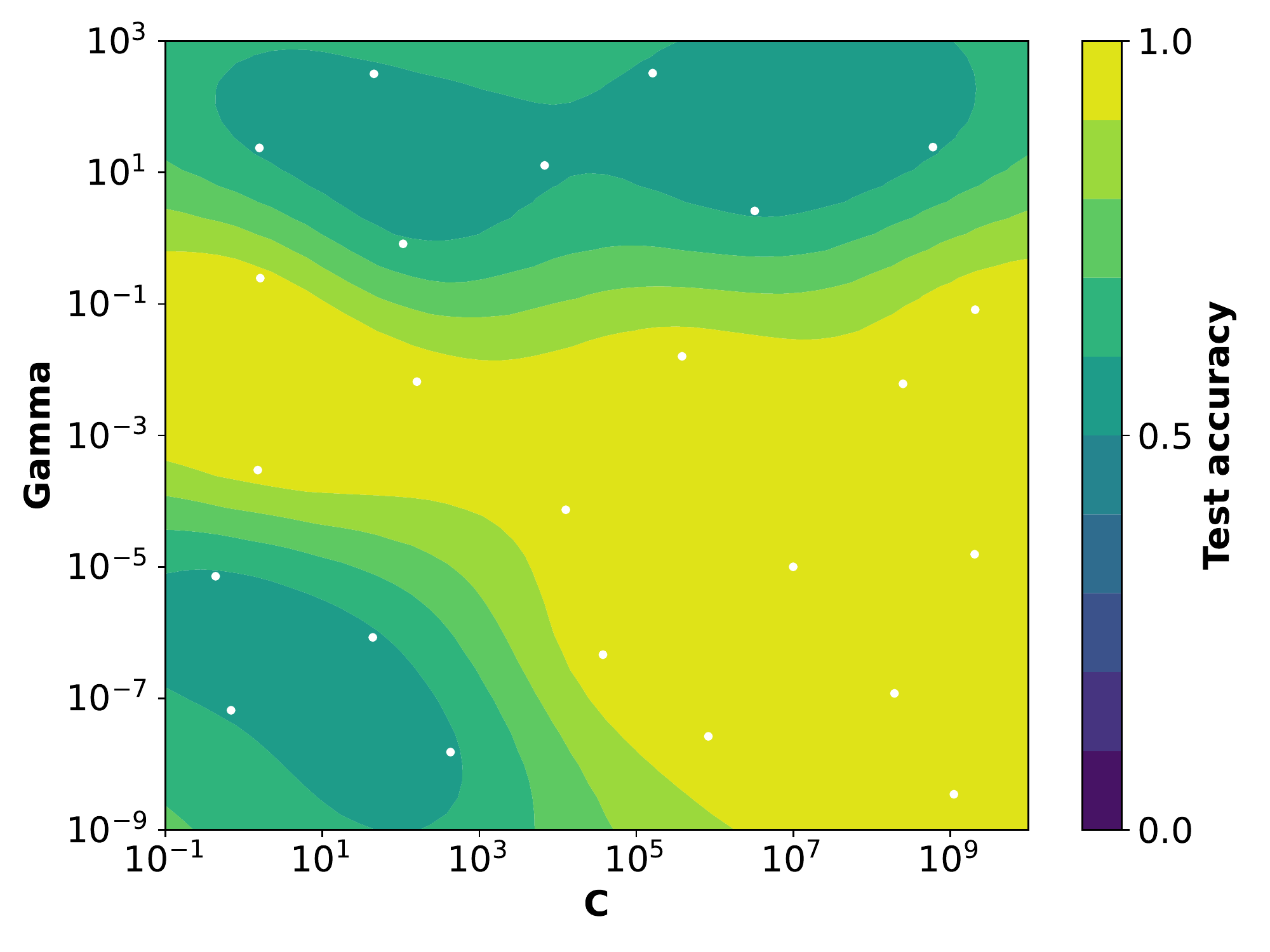}
    }
    \subfloat[IVR]{
        \includegraphics[trim={0 0 0 0.3cm},clip,width=0.33\textwidth]{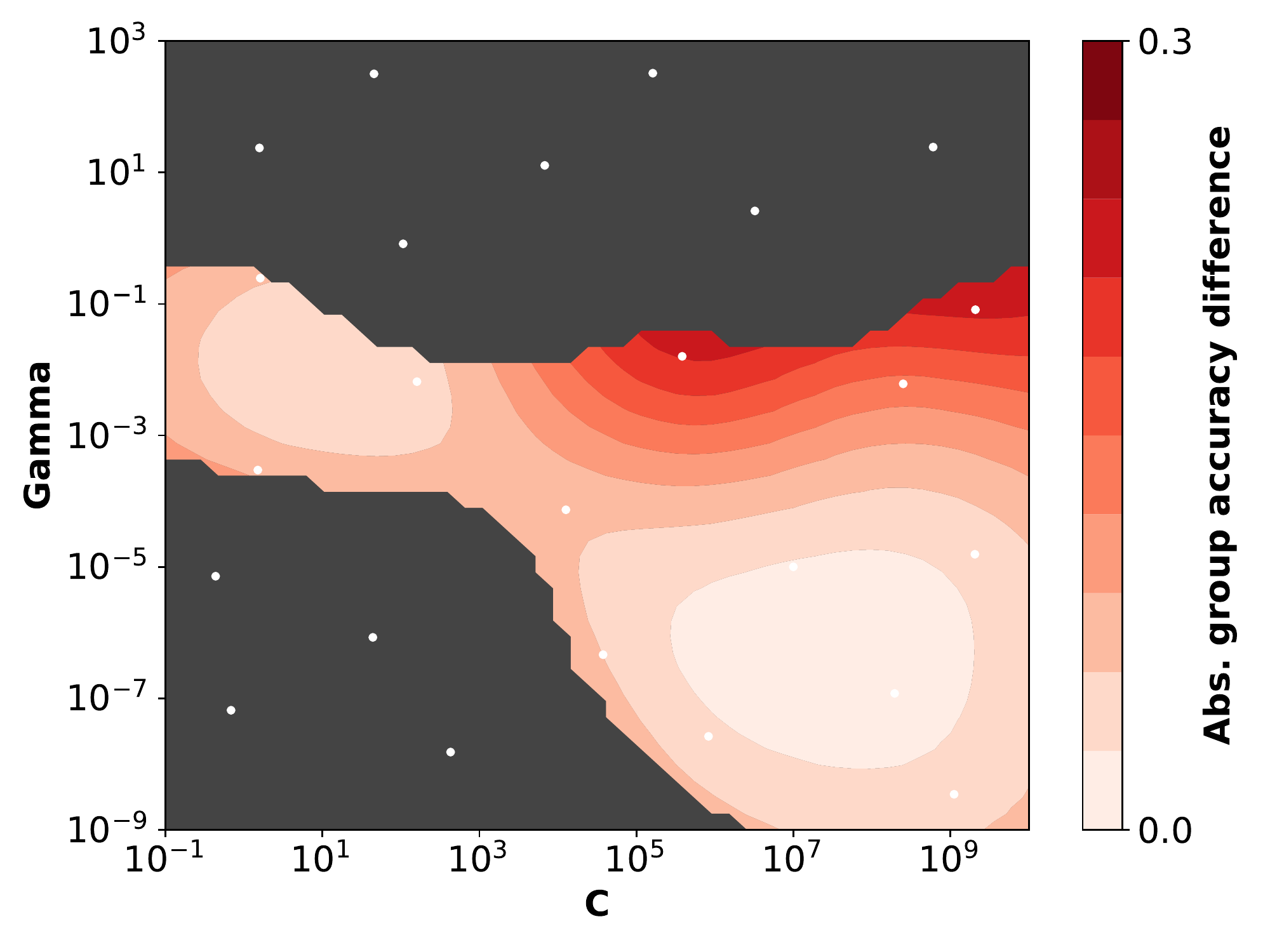}
    }
    \caption{Contour plot of GP-predicted mean test accuracy over search space of \(C\) and \(\gamma\) (Gamma) as explored by \textbf{(a)} UCB and \textbf{(b)} IVR acquisition functions. \textbf{(c)} Secondary objectives, e.g.\ minimizing group-level outcome differences, may vary along the IVR-revealed plateau.}\label{fig:svm}
\end{figure}

Here we motivate the multiverse analysis and our focus on exploration with an analysis of hyperparameter optimization.
To both aid replication and assess the effect of hyperparameter tuning on final results, there are calls for increased transparency around search space and tuning method (e.g.~\citep{Pineau2020}).
Going one step further, we argue that \emph{premature} optimization---even if transparently reported---hinders our understanding.
We illustrate this with a simple example, tuning two hyperparameters of an SVM classifier to show that optimizing leads to a distorted view of hyperparameter space.
We compare optimization and exploration using the upper confidence bound (UCB) \citep{Cox1992} and IVR acquisition functions.

In this multiverse analysis, we define the \textbf{search space} as the SVM's regularization coefficient \(C\) and the lengthscale \(\gamma\) of its RBF kernel.
Our \textbf{evaluation function} is the test accuracy of the SVM on the Wisconsin Breast Cancer Dataset \citep{Wolberg1995} of samples of suspected cancer to be classified as benign or malignant.
Given the same initial sample, we evaluate 23 further configurations for each acquisition function.

If we optimize, we would conclude from \cref{fig:svm}{a} that there is a single region of strong performance.
Conversely, exploring with IVR (\cref{fig:svm}{b}) shows that this region is in fact a plateau, and we are free to use any value of \(C\) as long as we scale \(\gamma\) accordingly.
While best test accuracy is similar in both cases, only by exploring do we learn about the full space of our options.
This knowledge is of vital importance if we properly account for additional real-world objectives, such as minimizing disparity in outcomes across different social groups.
We test this idea by assigning a synthetic majority/minority group label, \(g \sim \mathrm{Bern}(0.4)\), to each datapoint, and show in \cref{fig:svm}{c} how group-level outcomes can vary along the plateau revealed by IVR.
Prematurely optimizing could easily result in selecting a model that introduces group disparity.

There is, of course, an appropriate moment for optimization.
As one moves along the spectrum from research to deployment, so too should one move from exploration to exploitation.
Our aim here is not to critique optimization \emph{per se}, but to highlight that exploration is also of paramount importance.
When conducting scientific research in particular, we argue it is more appropriate to learn and understand as much possible, rather than eke out another marginal improvement.
Here we use a multiverse analysis as a framework for systematic exploration.

\section{Case study 1: When are adaptive optimizers helpful?}\label{sec:case-study-1}

\begin{figure}[b]
    \centering
    \subfloat[Mean surface]{
        \includegraphics[trim={0 0 0 0.3cm},clip,width=0.38\textwidth]{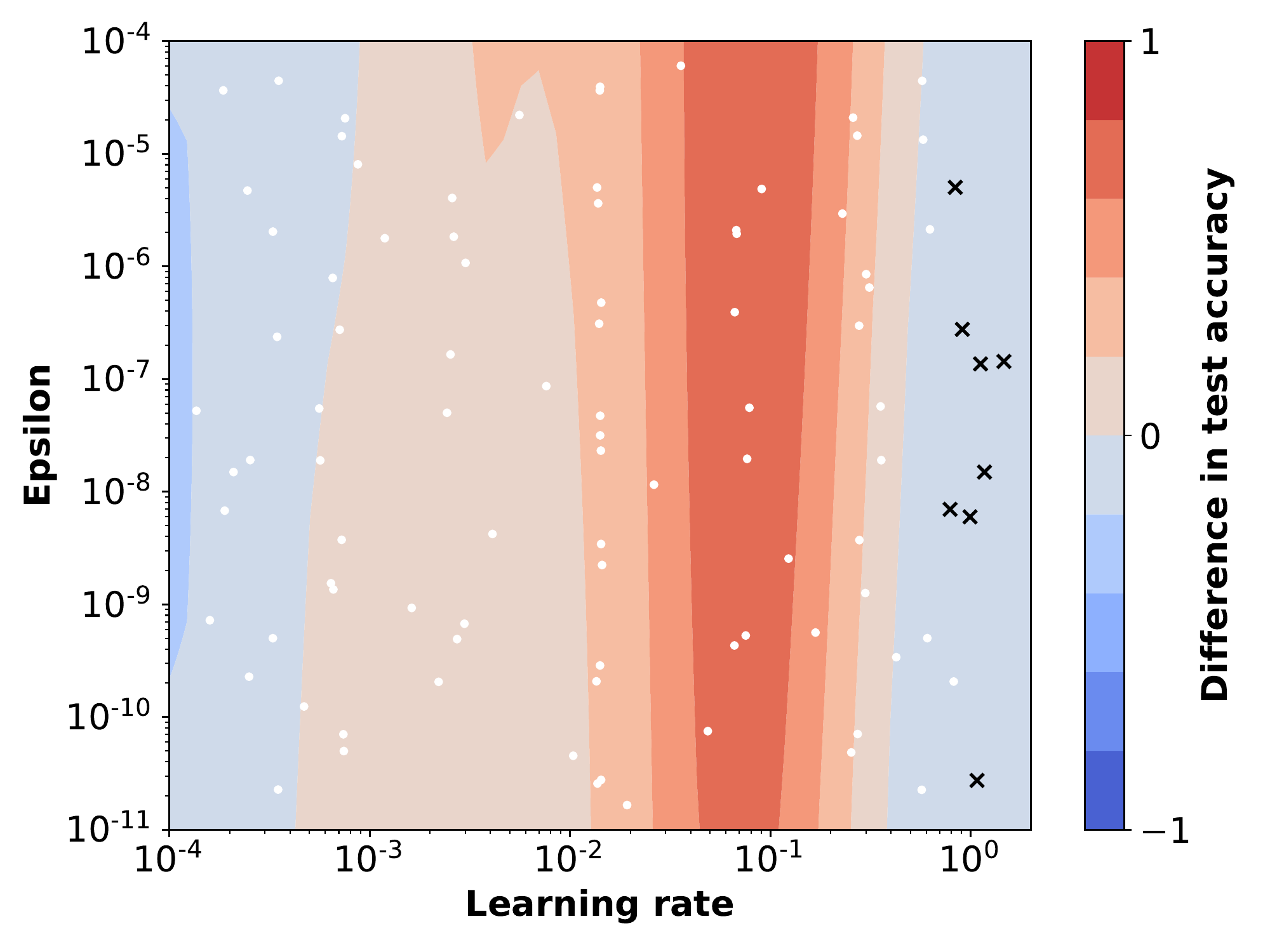}
    }
    \subfloat[Variance surface]{
        \includegraphics[trim={0 0 0 0.3cm},clip,width=0.38\textwidth]{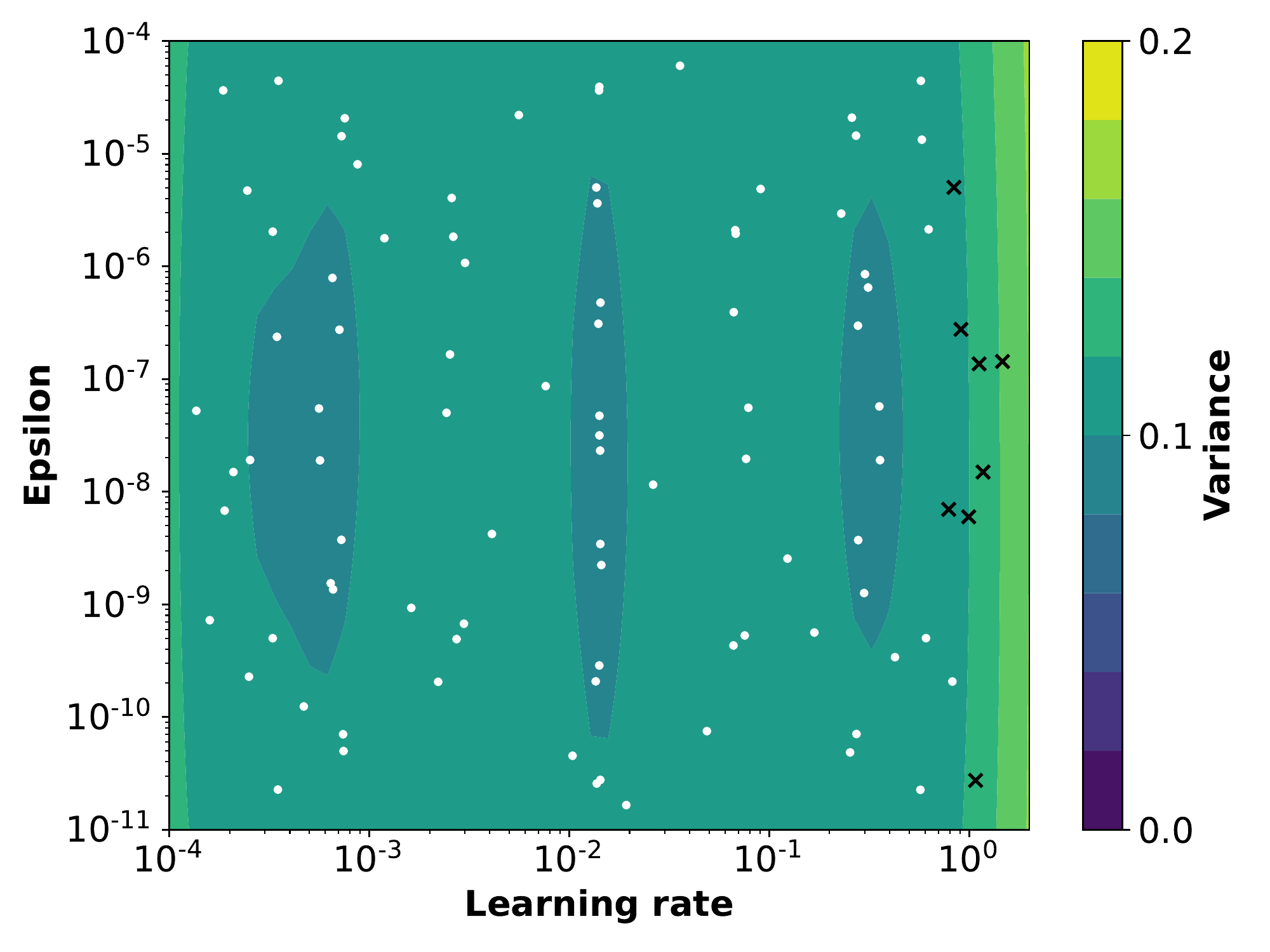}
    }
    \subfloat[Final accuracies]{
        \includegraphics[trim={0 -1cm 0 0},clip,width=0.225\textwidth]{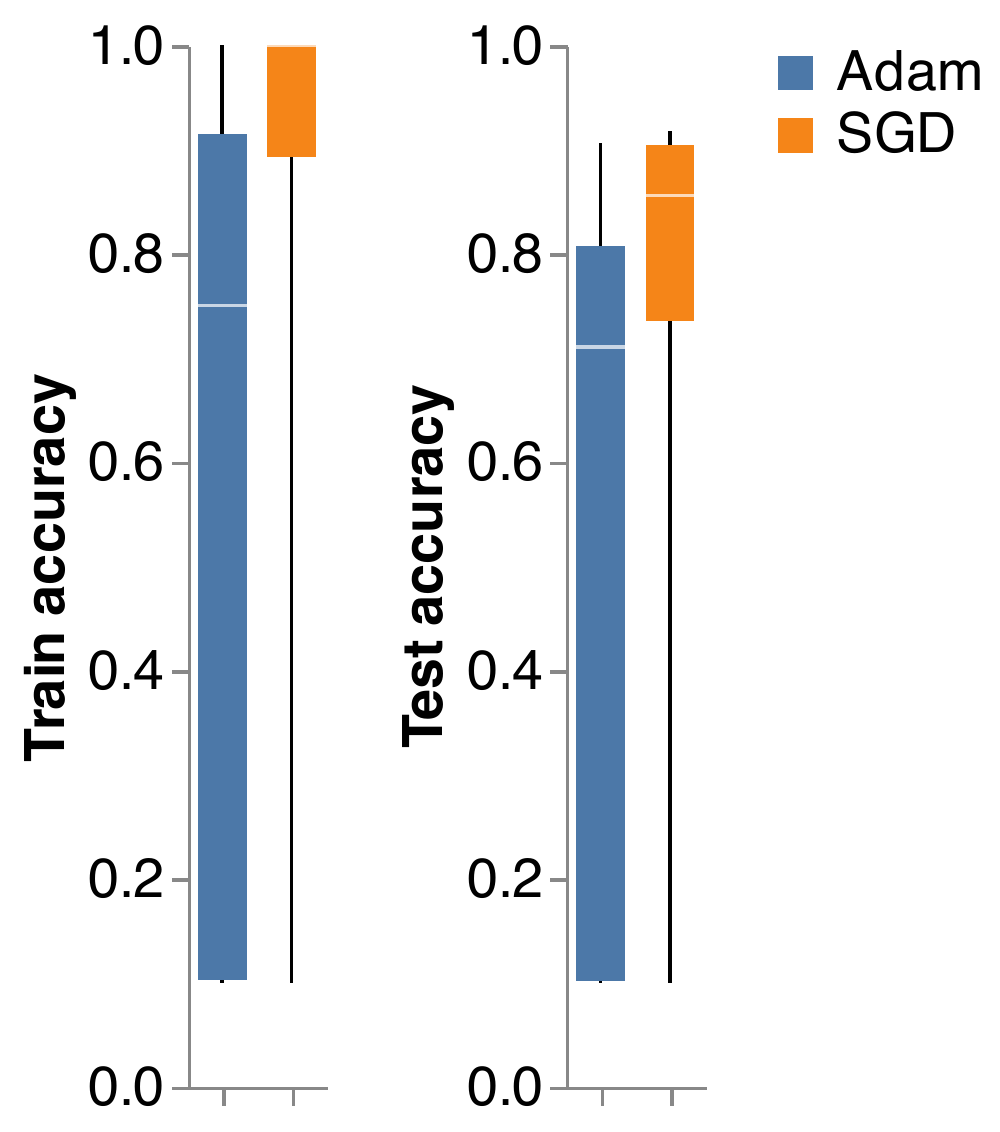}
    }
    \caption{Contour plot of GP-predicted \textbf{(a)} mean difference in test accuracy (SGD - Adam) and \textbf{(b)} variance over the search space of learning rate and \(\epsilon\). Red regions indicate SGD with momentum outperforms Adam. White points are successful trials; black crosses failed. \textbf{(c)} Final train and test accuracies. Whiskers extend to min and max. Note SGD train accuracy has median, UQ and max~\(1.0\).}\label{fig:adaptive-opt-contours}
\end{figure}

Adam \citep{Kingma2014} is a popular adaptive optimizer for training deep neural networks.
Questioning this practice, \citet{Wilson2017} suggest adaptive optimizers offer limited advantages over vanilla stochastic gradient descent (SGD) \citep{Robbins1951}.
They present experiments showing that SGD with momentum \citep{Polyak1964} outperforms all other optimizers across image recognition on CIFAR-10 \citep{Krizhevsky2009}, language modeling, and constituency parsing.

Under certain hyperparameter conditions, however, adaptive optimizers can be considered equivalent to SGD with momentum, highlighting the crucial role of hyperparameter tuning \citep{Choi2019}.
In replications of \citeauthor{Wilson2017}'s experiment with VGG \citep{Simonyan2014} on CIFAR-10, additionally tuning Adam's \(\epsilon\) hyperparameter---introduced solely for numerical stability and typically ignored---eliminates SGD's advantage \citep{Choi2019, Cooper2021}.


Both replications \citep{Choi2019, Cooper2021} can be considered proof by existence.
In order to refute \citeauthor{Wilson2017}'s claim, it suffices to find any point in hyperparameter space where Adam beats SGD with momentum.
However, this approach tells us little about relative optimizer performance \emph{in general}.
In this short case study, we perform a multiverse analysis of \citeauthor{Wilson2017}'s experiment with VGG on CIFAR-10.
We explore a reasonable and relevant search space and analyze how different hyperparameter choices lead to different optimizer recommendations.

\subsection{Multiverse Definition}

Because the claim under scrutiny compares optimizers, we define the \textbf{evaluation function} as the difference between model test accuracy achieved by Adam and SGD with momentum. 
Let \(\mathbf{g}_t = \nabla f(\theta_{t-1})\) be the minibatch-estimated gradient of the loss function $f$ w.r.t.\ the model parameters \(\theta_{t-1}\).
For SGD with momentum, we take a step of size \(\alpha\) in the direction of a decaying sum of recent gradients,
\begin{align*}
    \theta_{t} &= \theta_{t-1} - \alpha \mathbf{d}_t \, , \quad \mathbf{d}_t = \mu \mathbf{d}_{t-1} + \mathbf{g}_t \, ,
\end{align*}
where momentum parameter \(\mu\) controls decay.
In contrast, Adam steps along the gradient normalized by an unbiased estimate of its first and second moments \(\hat{\mathbf{m}}_t\) and \(\hat{\mathbf{v}}_t\):
\begin{align*}
    \mathbf{d}_t &= \frac{\hat{\mathbf{m}}_t}{\sqrt{\hat{\mathbf{v}}_t} + \epsilon} \, , \\
    \hat{\mathbf{m}}_t = \frac{\mathbf{m}_t}{1 - \beta^{t}_{1}} \, , \quad \mathbf{m}_t &= \beta_{1} \mathbf{m}_{t-1} + (1 - \beta_{1}) \mathbf{g}_t \, , \quad \mathbf{m}_0 = 0 \, , \\ 
    \hat{\mathbf{v}}_t = \frac{\mathbf{v}_t}{1 - \beta^{t}_{2}} \, , \quad \mathbf{v}_t &= \beta_{2} \mathbf{v}_{t-1} + (1 - \beta_{2}) \mathbf{g}^{2}_{t} \, , \quad \mathbf{v}_0 = 0 \, .
\end{align*}
Like \citeauthor{Wilson2017}, we use the default of \(\mu = 0.9\) and use the Polyak implementation.
For Adam, we also use default parameters \(\beta_1 = 0.9\) and \(\beta_2 = 0.999\).
The model is VGG-16 with batch normalization \citep{Ioffe2015} and dropout \citep{Srivastava2014}, trained for 300 epochs on CIFAR-10.

We set our \textbf{search space} to learning rate \(\alpha \in [10^{-4}, 10^{0}]\) by \(\epsilon \in [10^{-11}, 10^{-4}]\). \(\epsilon\) is only applied to Adam.
We evaluate 3 batches of 32 points.

\subsection{Results}

\begin{wrapfigure}[13]{R}{0.29\textwidth}
    \vspace{-35pt}
    \centering
    \subfloat[Main]{%
        \includegraphics[trim={0 0 0 0},clip,height=2.7cm]{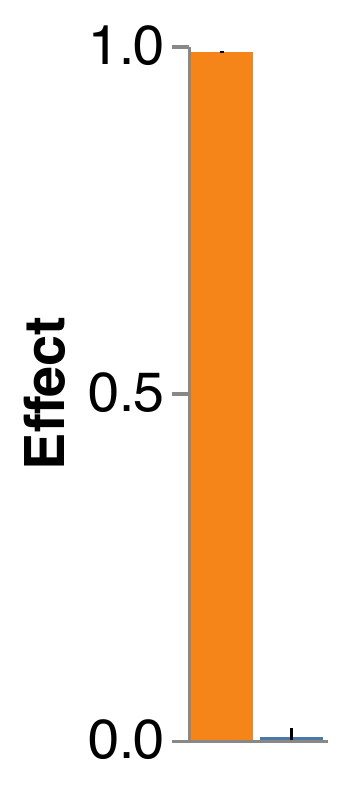}
    }\hspace{-0.8cm}
    \subfloat[Total]{%
        \includegraphics[trim={-2.3cm 0 0 0},clip,height=2.7cm]{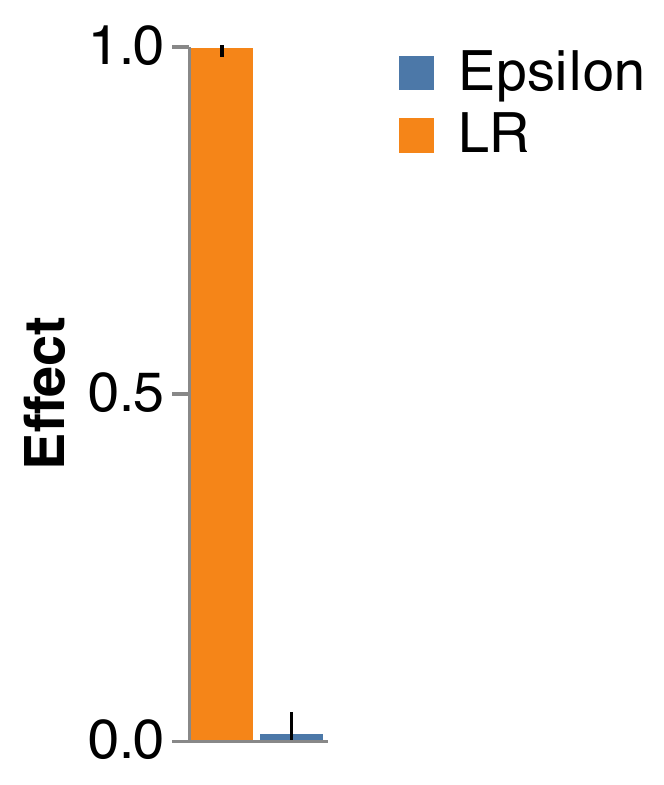}
    }
    \caption{\textbf{(a)} Main and \textbf{(b)} total effects of learning rate and \(\epsilon\). LR drives almost all output variance. Bars are STD.}\label{fig:adaptive-opt-sensitivity-analysis}
\end{wrapfigure}


The contour plot in \cref{fig:adaptive-opt-contours}{a} shows a large region (\(10^{-3} \leq \alpha \leq 10^{\frac{1}{2}}\)) in which SGD outperforms Adam.
However, we also identify regions (approx.\ \(\alpha < 10^{-3}\) or \(\alpha > 10^{\frac{1}{2}}\)) where the opposite is true, though we note the higher uncertainty (\cref{fig:adaptive-opt-contours}{b}) in the high learning rate region.
See \cref{fig:adaptive-opt-scatters} for raw results.

The change in relative performance is described almost entirely by learning rate.
The sensitivity analysis (\cref{fig:adaptive-opt-sensitivity-analysis}) reveals main effects (i.e.\ sensitivity to each variable in isolation) for learning rate and \(\epsilon\) of \(0.990\pm0.014\) and \(-0.004\pm0.0002\), and similarly total effects (i.e.\ sensitivity to each variable including its interaction with others) of \(0.996\pm0.014\) and \(0.008\pm0.031\).
Testing for interaction effects via Bayes factor (see \cref{eq:bayes-factor}), we find \(K = 2635\), thus rejecting the existence of an interaction between learning rate and \(\epsilon\).

In summary, we find that SGD with momentum often, but not always, outperforms Adam when training VGG on CIFAR-10, and that the conclusion we draw is highly sensitive to selected learning rate.
Unlike others \citep{Choi2019, Cooper2021}, we do not find a significant effect of \(\epsilon\), instead finding that learning rate determines which optimizer performs best.
In practice, our results suggest optimizer choice may make little difference to final test error, as long as an appropriate learning rate is used.
If budget for hyperparameter search is limited, \(\epsilon\) is unlikely to be the most efficient hyperparameter to include in a search.\footnote{That said, our results do not preclude the possibility that with a much more extensive and optimization-based search, the resulting estimate of the search space might reveal some (modest) room for improvement from tuning \(\epsilon\). However, as indicated by the sensitivity analysis in \cref{fig:adaptive-opt-sensitivity-analysis}, any effect is likely to be dwarfed by properly tuning learning rate, and is unlikely to materially alter the best choice of optimizer.}
Using a multiverse analysis, we have systematically explored how decisions about learning rate and epsilon impact conclusions about the optimal optimizer.
In doing so, we have shown that reported results may vary according to researcher choices.

\section{Case study 2: Is there a large-batch generalization gap?}\label{sec:case-study-2}

\begin{figure}[t]
    \centering
    \subfloat[AlexNet on CIFAR-10]{
        \includegraphics[trim={0 0 3.5cm 0},clip,width=0.30\textwidth]{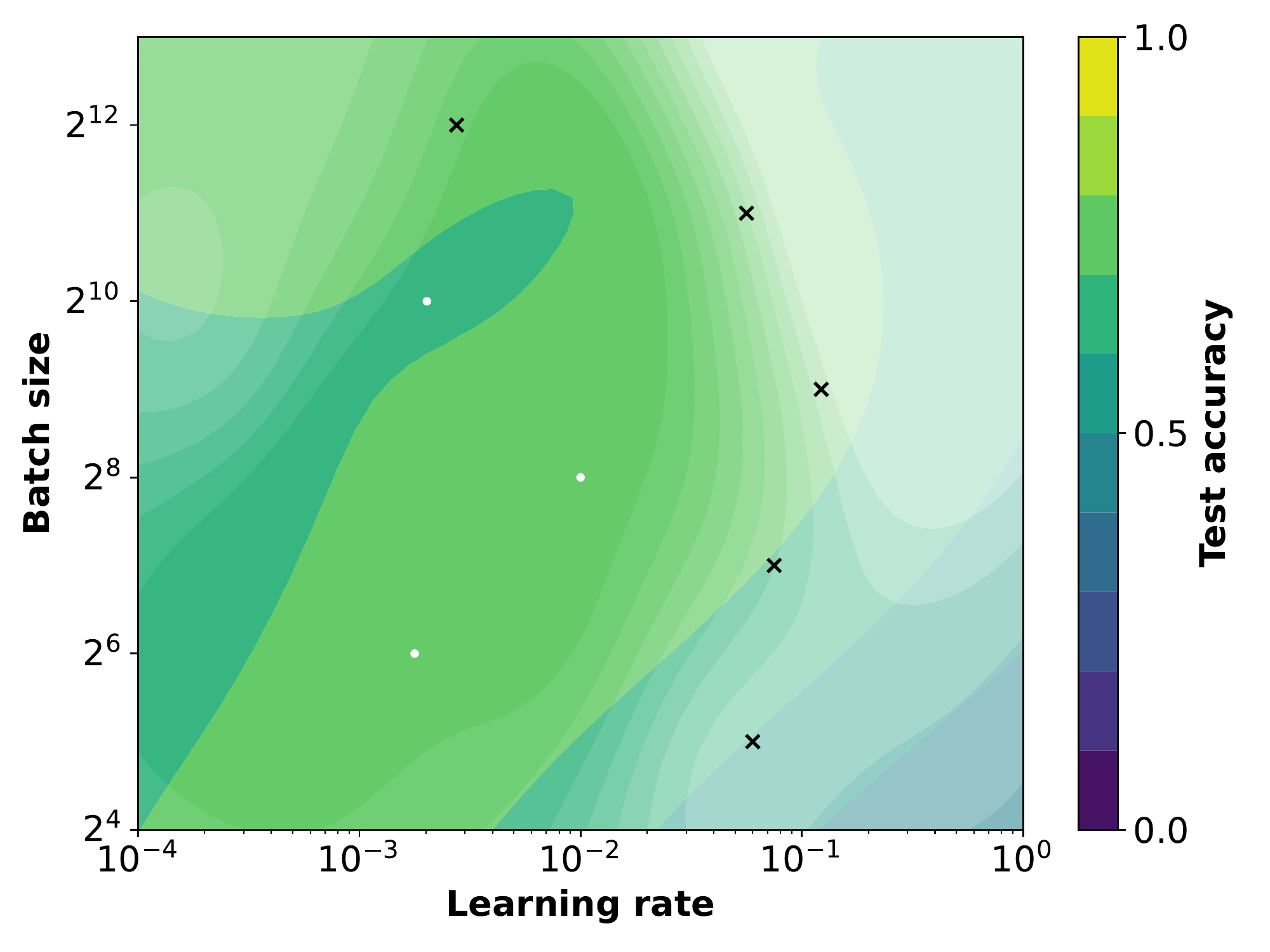}
    }
    \subfloat[ResNet on CIFAR-10]{
        \includegraphics[trim={0 0 3.5cm 0},clip,width=0.30\textwidth]{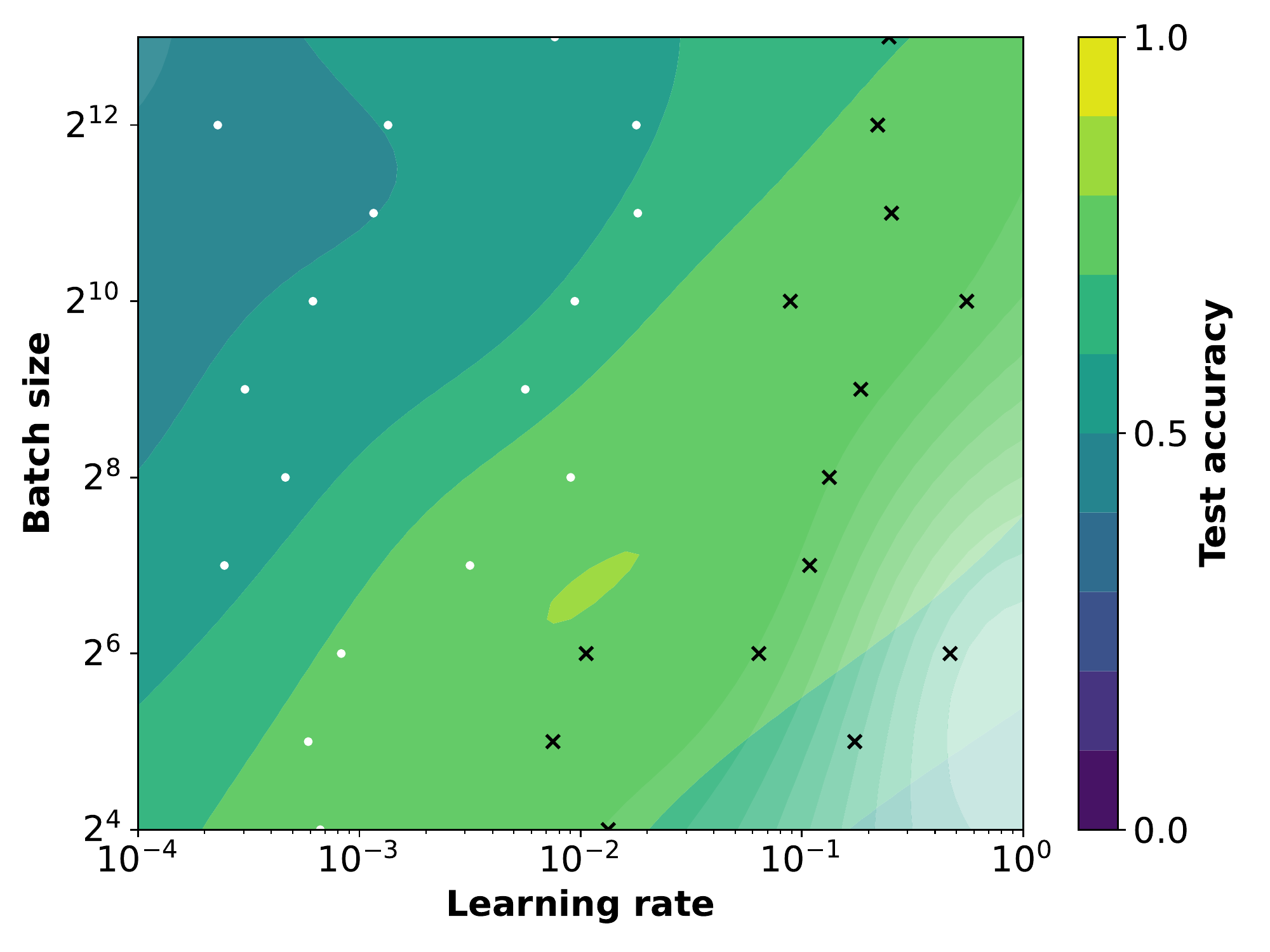}
    }
    \subfloat[VGG on CIFAR-10]{
        \includegraphics[trim={0 0 0 0},clip,width=0.3625\textwidth]{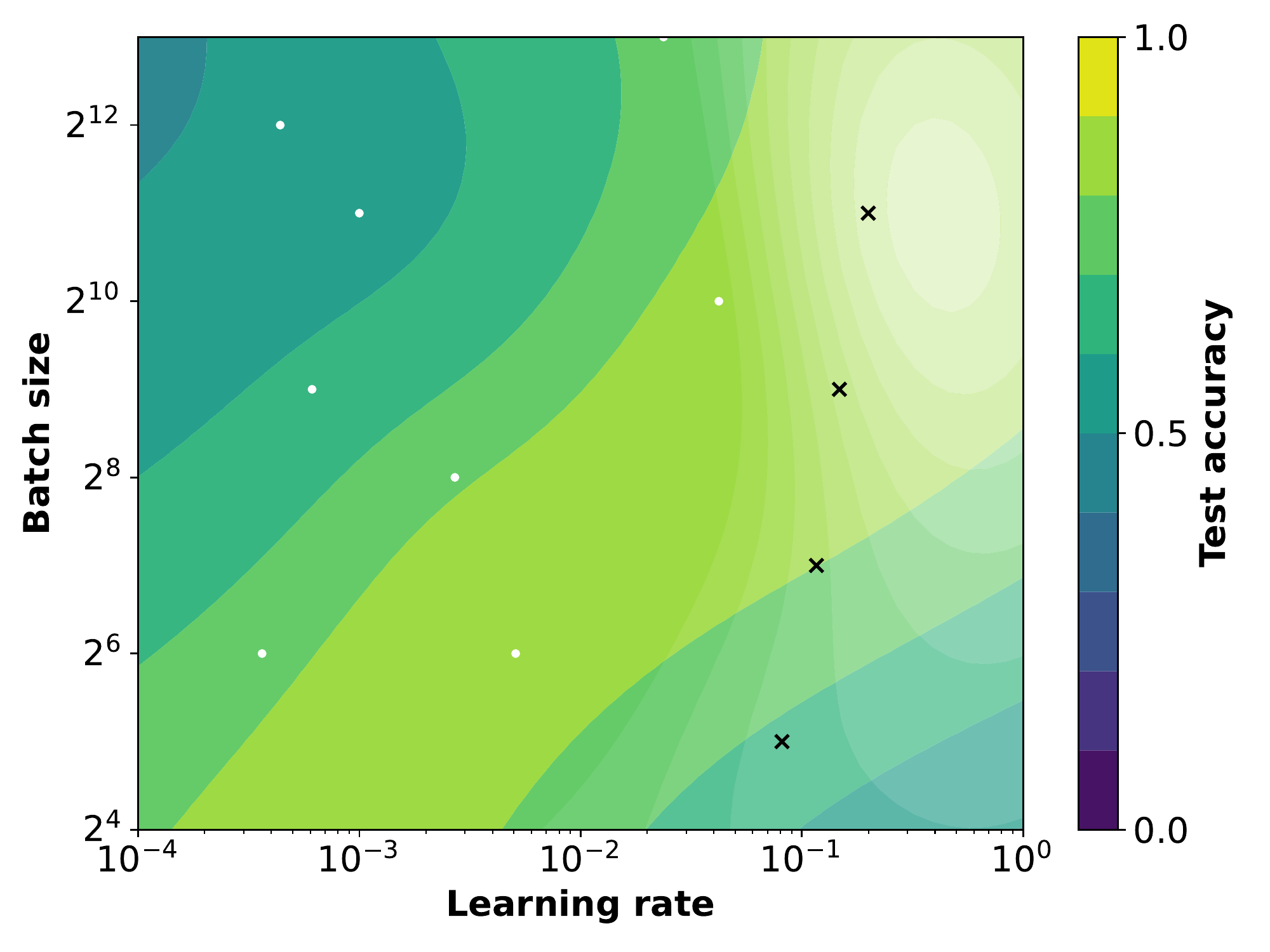}
    } \\ 
    \subfloat[AlexNet on CIFAR-100]{
        \includegraphics[trim={0 0 3.5cm 0},clip,width=0.30\textwidth]{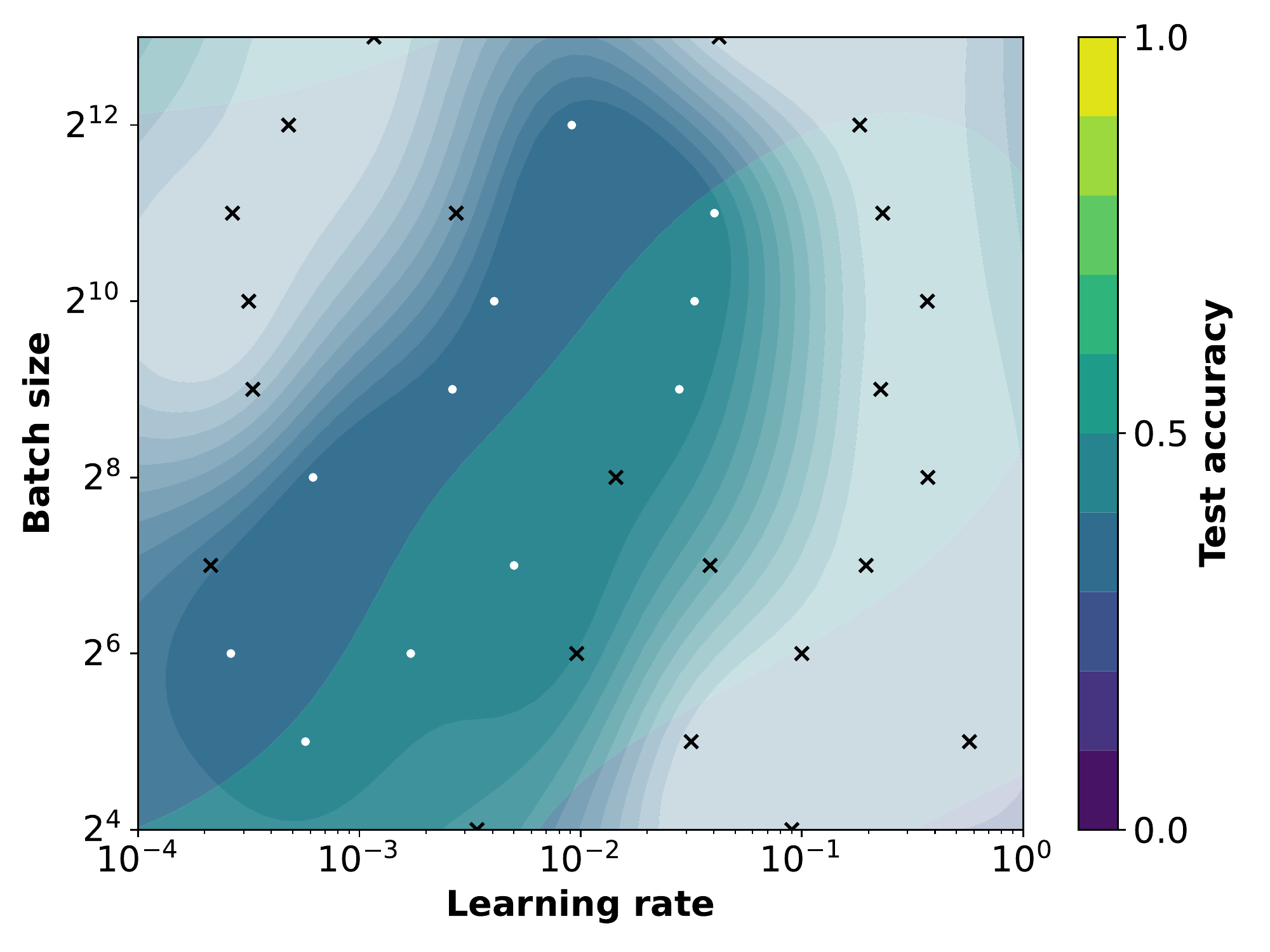}
    }
    \subfloat[ResNet on CIFAR-100]{
        \includegraphics[trim={0 0 3.5cm 0},clip,width=0.30\textwidth]{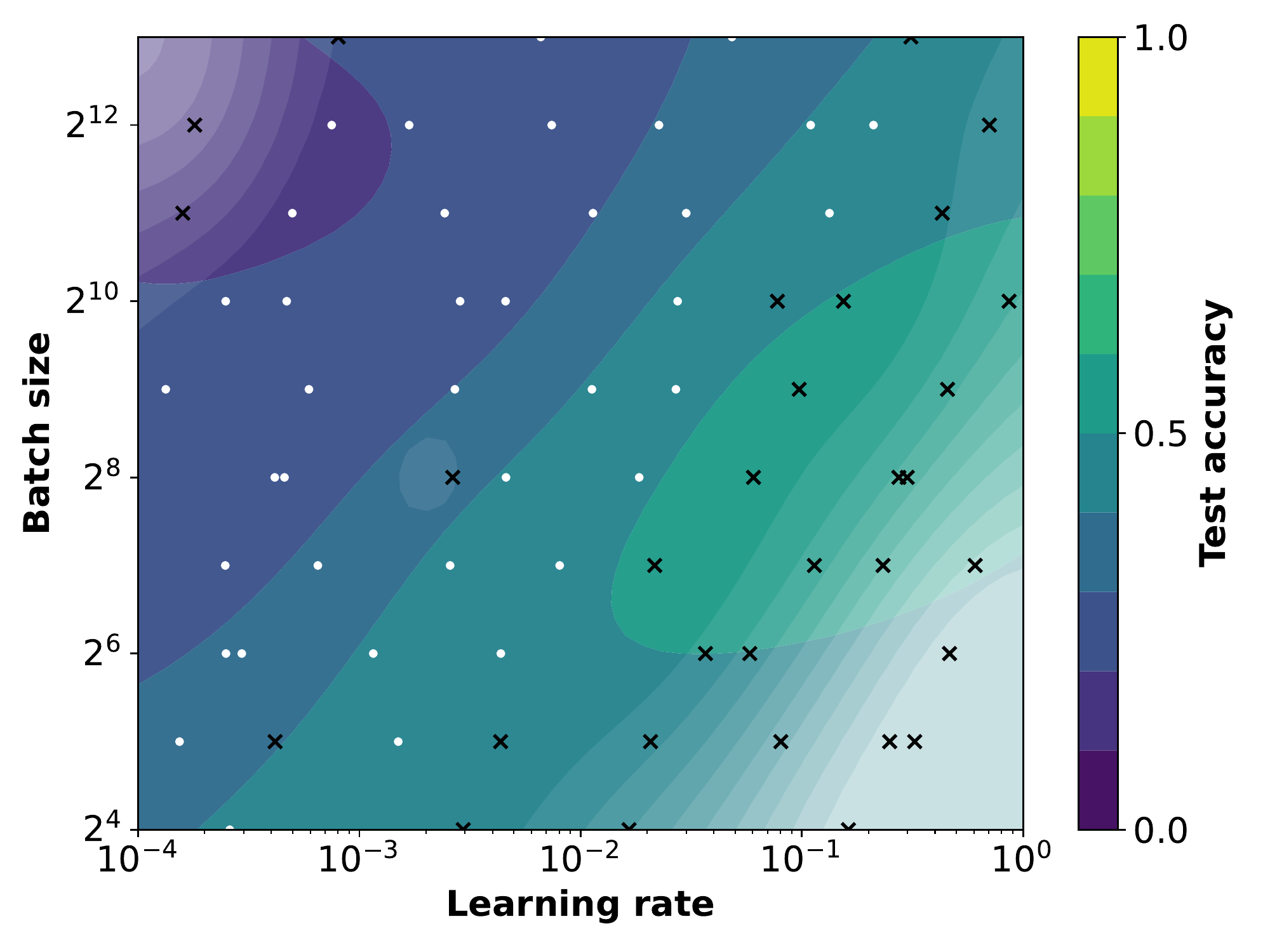}
    }
    \subfloat[VGG on CIFAR-100]{
        \includegraphics[trim={0 0 0 0},clip,width=0.3625\textwidth]{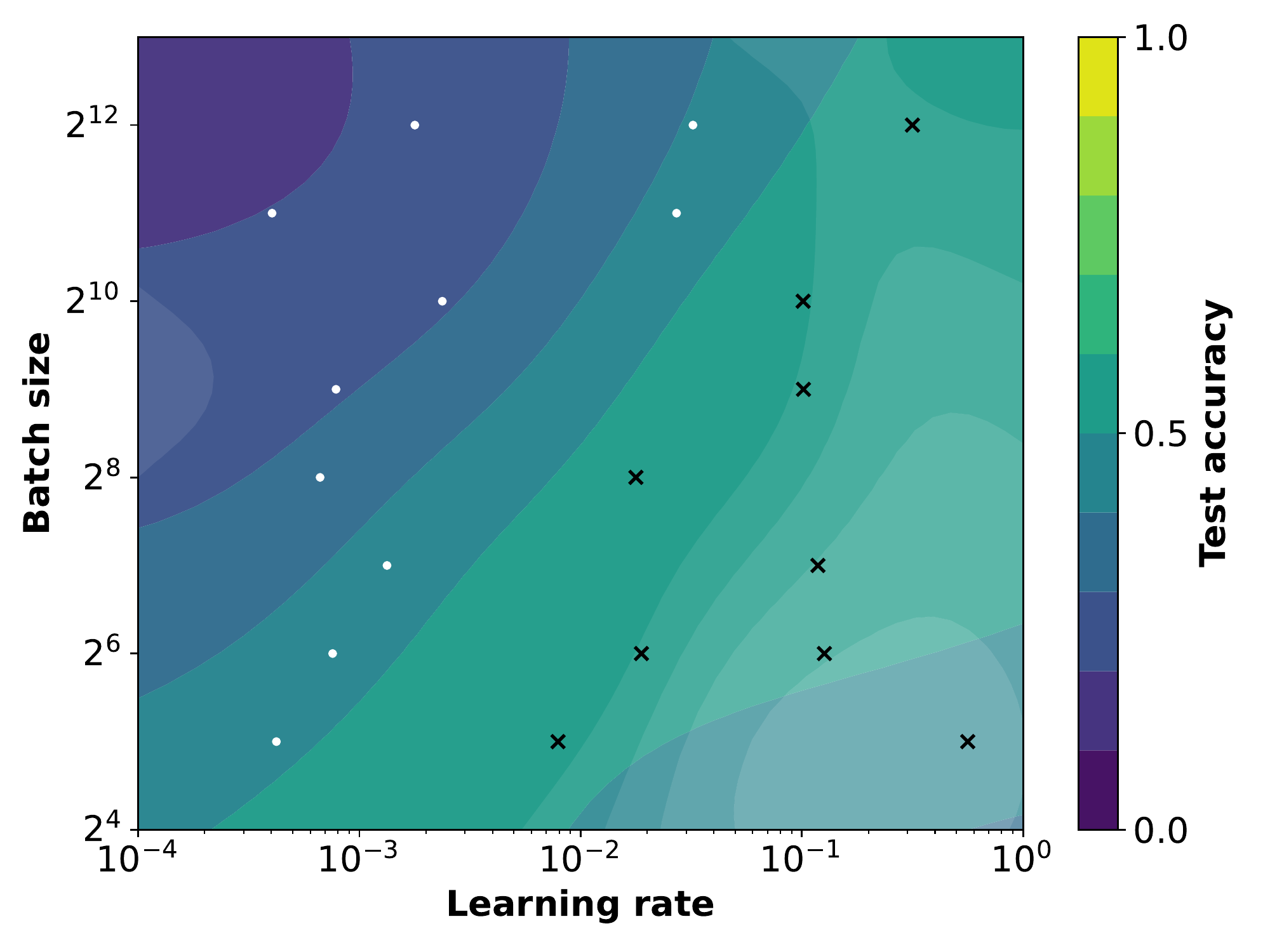}
    }
    \caption{Contour plot of GP-predicted mean test accuracy over the search space of learning rate, batch size, dataset and model. White points are trials with training accuracy \(\geq 0.99\); black crosses were excluded. Overlayed translucent regions indicate high training error. For Tiny ImageNet see \cref{fig:batch-size-contours-tiny-imagenet}; for variance see \cref{fig:batch-size-contours-var}. The discrepancy between contours and data points in \textbf{(a)} is due to the coregionalized model sharing information across functions.}\label{fig:batch-size-contours}
\end{figure}

Larger batch sizes are a prerequisite for distributed neural network training.
However, many researchers have reported a ``generalization gap'', where generalization performance declines as batch size increases \citep{Keskar2017, Golmant2018, Masters2018}, a phenomenon potentially \citep{Hochreiter1997, Li2017, Jastrzkebski2017, Chaudhari2019} but not conclusively \citep{Dinh2017} linked to the sharpness of the resulting minima.


Subsequent work has sought ways to mitigate the generalization gap by, in broad strokes, either scaling up the batch size throughout training \citep[pp.~262-265]{Devarakonda2017, Bottou2018}, or by scaling learning rate proportional to batch size \citep{Hoffer2017, Smith2017, Goyal2017, You2017}.
To make sense of these results, one must cut through myriad researcher choices about datasets; model architectures; termination criteria \citep{Shallue2019}; linear \citep{Goyal2017} or sublinear (e.g.\ square root)~\citep{Hoffer2017} scaling rules; high initial learning rate \citep{Hoffer2017}; learning rate decay or warmup \citep{Goyal2017}; layer-specific learning rates \citep{You2017}; batch norm variants \citep{Hoffer2017}; regularization techniques e.g.\ label smoothing~\citep{Shallue2019}; and so on.
In our second case study, we use a multiverse analysis to synthesize existing research on the relationship between batch size, learning rate and generalization error.

\subsection{Multiverse Definition}

Our \textbf{evaluation function} is the test accuracy of a model trained according to the sampled configuration.
The \textbf{search space} includes learning rate \(\alpha \in [10^{-4}, 10^{-\frac{1}{2}}]\), batch size \(\in \{2^4, \ldots, 2^{13}\}\), model~\(\in \{\)AlexNet \citep{Krizhevsky2012}, VGG \citep{Simonyan2014}, ResNet \citep{He2016}\(\}\), and dataset \(\in \{\)CIFAR-10, CIFAR-100 \citep{Krizhevsky2009}, Tiny ImageNet\citep{TinyImageNet}\(\}\). 
Specifically, we use VGG-16 with batch norm and ResNet-18.
Tiny ImageNet was selected as a substitute for ImageNet \citep{Russakovsky2015} to limit compute expenditure.

For categorical parameters, we treat each pair of model and dataset as its own function with an intrinsic coregionalization model \citep{Helterbrand1994,Alvarez2012}.
Given base kernel \(k\), our multi-output kernel matrix is:
\begin{align*}
	B_{m} &= \textbf{w}_m \textbf{w}_{m}^{\top} + \textrm{diag}(\kappa_{m}) \, , \\ 
	B_{d} &= \textbf{w}_d \textbf{w}_{d}^{\top} + \textrm{diag}(\kappa_{d}) \, , \\ 
	K(X, X) &= B_{m} \otimes B_{d} \otimes k(X, X) \, ,
\end{align*}
where \(B_{d}\) and \(B_{m}\) represent dataset and model respectively.
In interpreting these parameters, the outer product of \(\textbf{w}\) defines how related each output is, whereas \(\kappa\) allows them to vary independently.


We evaluate 6 batches of 32 points. 
Before analysis, we discard all model runs with training accuracy~\(< 0.99\) and re-fit the model.
In addition to previous methods we also analyze the coregionalization parameters to investigate how much dataset and model impact our results.

\subsection{Results}

\begin{wrapfigure}[41]{r}{0.34\textwidth}
    \vspace{-50pt}
	\centering
    \subfloat[Dataset]{
        \centering
        \includegraphics[trim={0.4cm 2.2cm 0 1.4cm},clip,width=0.33\textwidth]{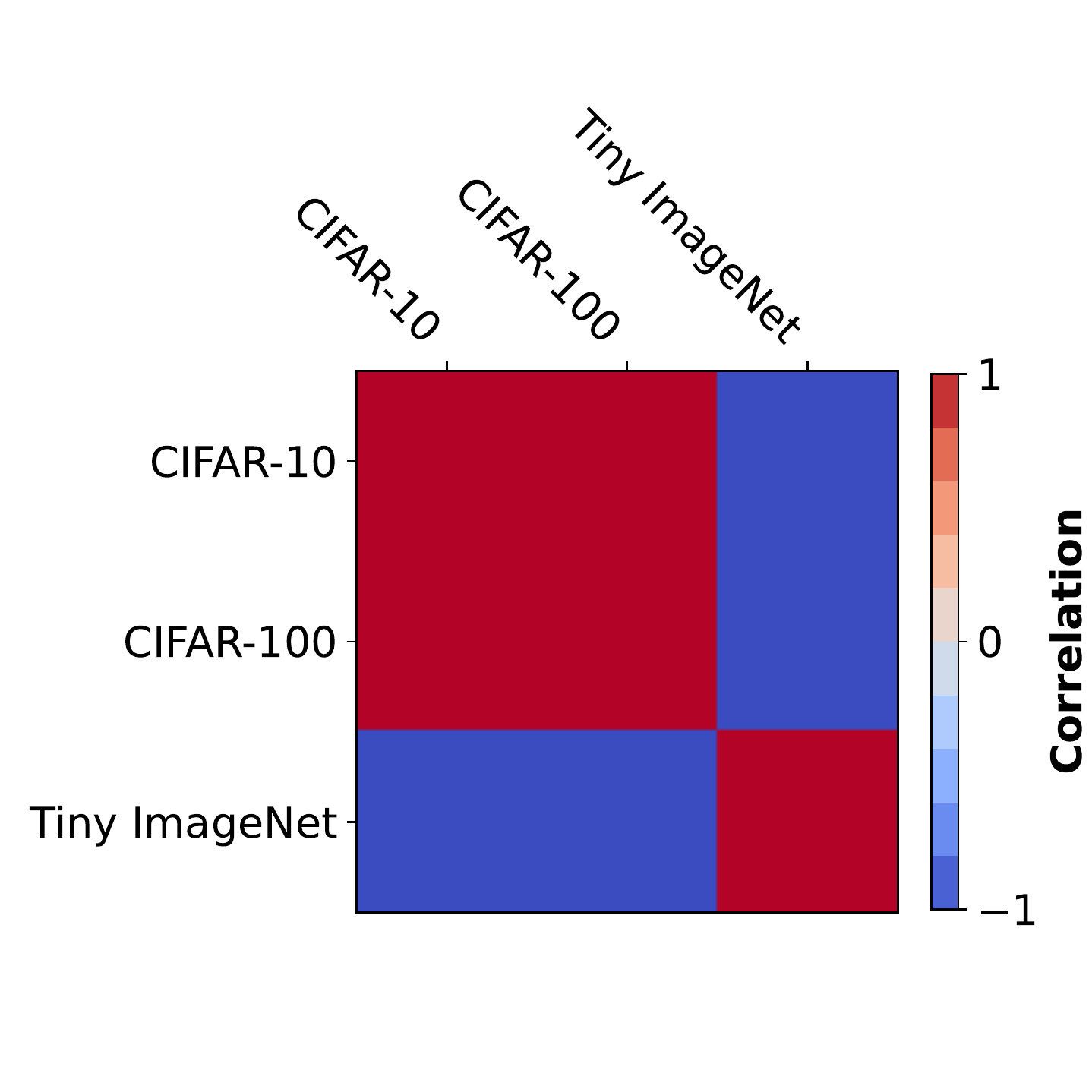}
	}\\
    \subfloat[Model]{
        \centering
        \includegraphics[trim={0 2cm 0 2cm},clip,width=0.33\textwidth]{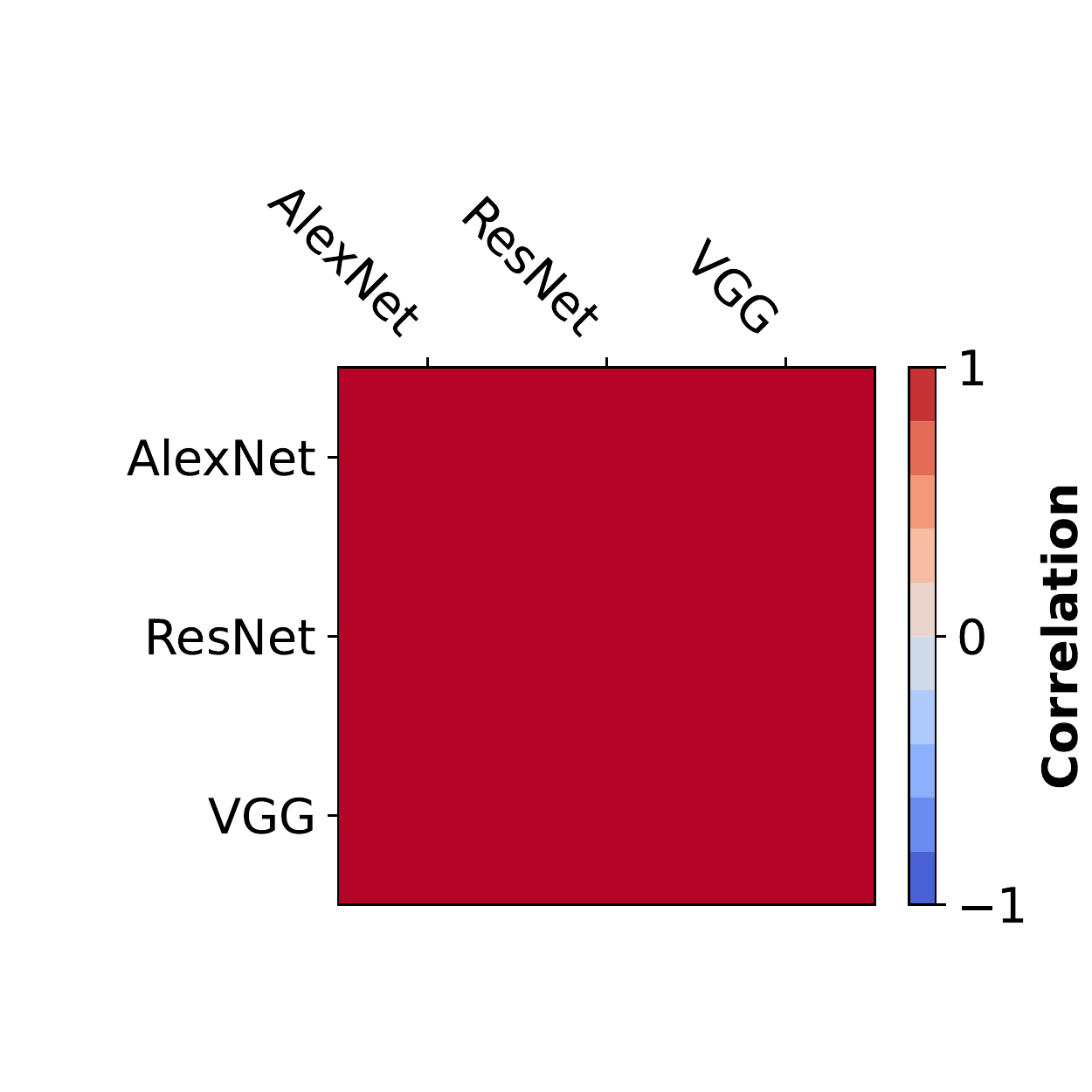}
    }
    \caption{Multi-output GP function correlations. \textbf{(a)} Tiny ImageNet is moderately negatively correlated with CIFAR-10/100. \textbf{(b)} All three model outputs are highly correlated.}\label{fig:batch-size-coregionalization}
    \vspace{10pt}
    \subfloat[Main]{%
        \includegraphics[trim={-1 0 0 0},clip,height=2.7cm]{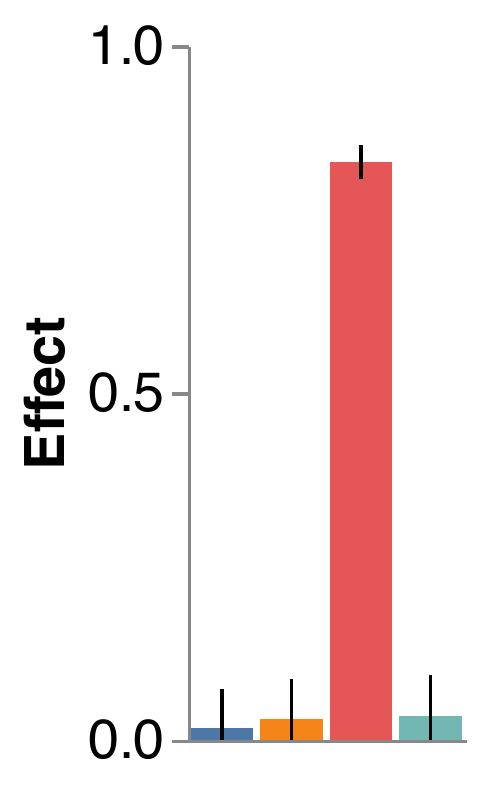}
    }\hspace{-1.2cm}
    \subfloat[Total]{%
        \includegraphics[trim={-3cm 0 0 0},clip,height=2.7cm]{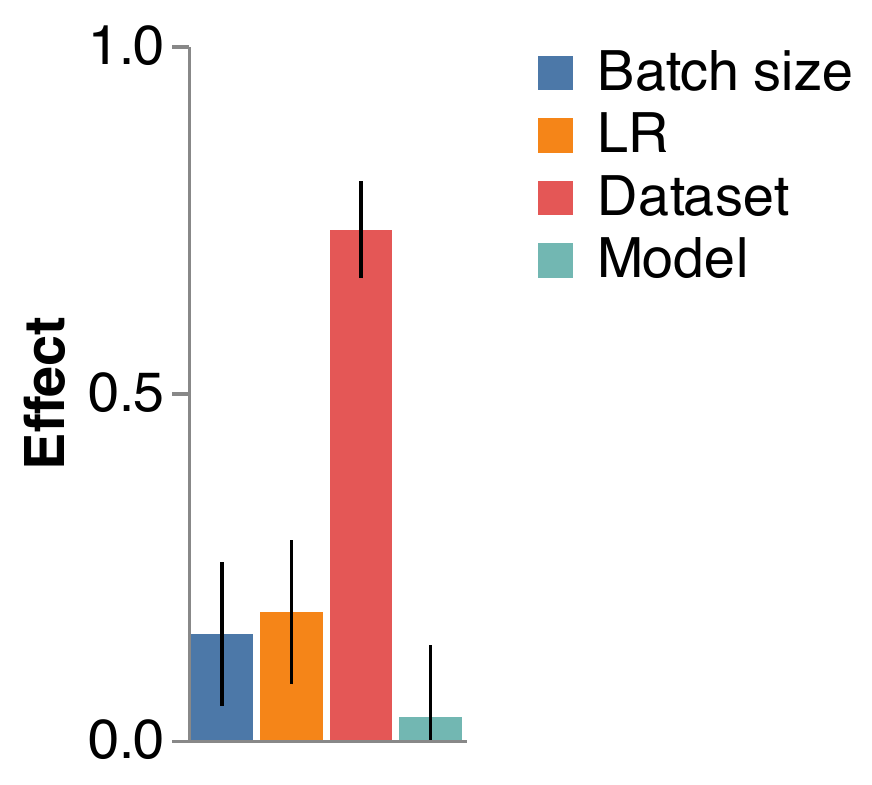}
    }
    \caption{\textbf{(a)} Main and \textbf{(b)} total effects of all parameters. LR drives more output variance than batch size, but dataset is most important. Increased total effect for LR and batch size confirms interaction effect. Bars are STD.}\label{fig:batch-size-sensitivity-analysis}
\end{wrapfigure}

Of 224 trials, 9 failed to converge and a further 115 did not reach training accuracy \(\geq 0.99\).
A preliminary analysis of the raw results (see \cref{fig:batch-size-scatters}) shows test accuracy is barely positively correlated with learning rate (\(\rho = 0.08\)) and slightly negatively correlated with batch size (\(\rho = -0.2\)).
In \cref{fig:batch-size-contours}, however, a consistent plateau emerges across all models on both CIFAR-10 and CIFAR-100:
test accuracy is maintained as long as batch size scales up with learning rate.
On both sides of this plateau, if either learning rate or batch size are too large, generalization error will increase. 

We were unable to successfully train any model on Tiny ImageNet.
Of the 42 models we trained on Tiny ImageNet, 22 reached the \(0.99\) training accuracy threshold, but none of these obtained a test accuracy higher than 0.01, indicating significant overfitting.
See \cref{fig:batch-size-contours-tiny-imagenet} for Tiny ImageNet contours.

Testing for interaction effects via Bayes factor (see \cref{eq:bayes-factor}), we find \(K = 0.58\), suggesting that an additive kernel is not sufficient to explain the data and indicating an interaction effect.
This implies that batch size alone does not explain generalization, but batch size and learning rate \emph{together}.
Our modeling approach gracefully handles the failures on Tiny ImageNet, such that even with Tiny ImageNet trials removed, the interaction effect remains present (\(K = 0.34\)).

The coregionalization kernels explain how model and dataset affect our results.
In \cref{fig:batch-size-coregionalization}{b} all model outputs are highly correlated, indicating that choice of model does not impact the relationship between batch size, learning rate, and generalization error.
In contrast \cref{fig:batch-size-coregionalization}{a} shows highly correlated outputs across CIFAR-10 and CIFAR-100, and moderate negative correlation with Tiny ImageNet.
A strength of our chosen coregionalized model is that it flexibly captures the batch size and learning rate relationship in CIFAR-10 and 100, despite the failures on Tiny ImageNet.
Our results on CIFAR-10 and CIFAR-100 indicate that the relationship may hold across datasets \emph{where training was successful}, including across different dataset complexities, in contrast to previous work \citep{Golmant2018}.

Finally, our sensitivity analysis in \cref{fig:batch-size-sensitivity-analysis} reveals next to no main effect for either learning rate (\(0.03\pm0.06\)) or batch size (\(0.02\pm0.06\)).
However, the larger total effects for both learning rate (\(0.19\pm0.11\)) and batch size (\(0.15\pm0.11\)) again support the existence of an interaction effect.
Both learning rate and batch size are dominated by the main effect of dataset (\(0.83\pm0.03\)), though this is likely due to Tiny ImageNet.

In conclusion, our framework reveals a complex interaction between learning rate, batch size, and generalization error that is consistent across all models and two out of three datasets.
These results support scaling learning rate linearly with batch size \citep{Goyal2017, Jastrzkebski2017, Smith2017}, but do not support a threshold batch size (up to \(2^{13}\)) after which learning rate is no longer corrective \citep{Golmant2018, Shallue2019}.
After accounting for learning rate, we find no consistent evidence of the proposed \citep{Masters2018, Keskar2017} large-batch generalization gap.
Here, a multiverse analysis has proven a useful tool to pull together disparate research on the generalization gap and to surface underlying trends.


\section{Discussion}\label{sec:discussion}

In the three applications of the multiverse analysis presented here, we have demonstrated that decisions taken by the researcher can have a significant effect on the final conclusions drawn.
In our SVM hyperparameter example, we highlight the important role of systematic exploration and show that premature optimization may both limit our understanding and present only a partial truth to downstream practitioners.
Our first case study on optimization demonstrated how varying the learning rate results in fundamentally different conclusions about whether to use SGD with momentum or Adam for optimization.
Conversely, our second case study showed a simple relationship between batch size, learning rate and generalization error irrespective of model and dataset.

\subsection{Choice of search space}

Across these examples we have tried to illustrate differently complex analyses by way of differently sized search spaces.
In case study 1 we inflated a rather small multiverse focusing only on directly relevant hyperparameters, though in case study 2 we built a larger multiverse including parameters (model and dataset) that vary freely between existing research.
An expected critique of multiverse analyses is that choice of search space is itself a choice.\footnote{The same applies to GP surrogate setup, e.g.\ kernel choice, though we suggest our settings used in this work will be a suitable default for most purposes. See \cref{fig:adaptive-opt-kernel-comparison,fig:batch-size-kernel-comparison-matern-ard,fig:batch-size-kernel-comparison-matern-no-ard,fig:batch-size-kernel-comparison-rbf-ard,fig:batch-size-kernel-comparison-rbf-no-ard} for a comparison of different kernels.}
However, choosing dimensions and bounds for a search space is more principled than choosing specific points, and that declaring each dimension makes assumptions of relevance or irrelevance explicit.
That said, it is always possible for subsequent work to critique the choice of search space and to add new dimensions.
Indeed, our first case study \emph{should} be expanded to include each of \citeauthor{Wilson2017}'s experiments with different model architectures, datasets, deep learning frameworks and additional adaptive optimizer variants, though we reinforce that we chose a limited analysis to provide a simple case study.
Our second case study also presents a number of interesting avenues for expansion in subsequent analyses, including additional datasets and model architectures, learning rate schedules and optimizers.
Most interesting would be inclusion of termination criterion and evaluation metric, both previously highlighted as key drivers of divergent findings \citep{Shallue2019}.

\subsection{Compute cost}\label{sec:compute-cost}

Our main contribution is to present the multiverse analysis framework and show how it can be used to draw robust conclusions about model performance, so we allocated our compute budget to showcase illustrative examples.
Our framework is equally appropriate for research pushing state-of-the-art with a larger budget, though such experiments aren't necessary to demonstrate our framework's value.
In this project, we used \(1138\) hours of GPU time, at a rough cost of \$775 with \SI{140}{\kilogram} of \ch{CO2}.\footnote{Calculated using \url{https://mlco2.github.io/impact} \citep{Lacoste2019} assuming A100 GPUs on the University of
Cambridge HPC cluster with carbon efficiency \SI[per-mode=symbol,sticky-per,bracket-unit-denominator=false]{0.307}{\kilogram \ch{CO2}\per\kWh}.} 
We expect costs to scale with multiverse size, requiring a pause for consideration in light of recent critiques of the environmental impacts of ML \citep{Lacoste2019, Strubell2019, Schwartz2020}.
In response, we first note that at least some of the required compute is already taking place as part of existing trial-and-error workflows.
Second, through the introduction of a surrogate model for the multiverse, one can substantially reduce the amount of exploration required. 
Finally, we suggest that \emph{all} resources consumed in producing non-robust results, and those of subsequent work that builds upon them, are wasted by definition.

\subsection{Future work}

It is common practice when reporting model performance to train a number of runs with different random seeds and report the mean and standard deviation.
Number of runs varies according to researcher budget and time.
Appropriate sample size to enable robust inference in the face of noise is rarely considered.
Our approach to the multiverse analysis presents an elegant potential solution to this issue. 
In our examples thus far we train only a single model for each point and assume homoscedasticity for simplicity.
However, research in experimental design suggests that accounting for heteroscedasticity by separately modeling the variance in the surrogate \citep{Goldberg1997, Binois2018} could allow for a principled trade-off between conducting another run and a sampling a new configuration \citep{Binois2019}.


While we introduce the multiverse analysis to ML, it has previously been applied to a handful of studies in human computer interaction \citep{Dragicevic2019}.
In an exciting area for development, the authors also develop interactive visualizations to help the reader explore the effects each choice, up to and including re-rendering written conclusions in an online version of the paper.
Given that the largest multiverse in our work is 4-dimensional, our faceted contour plots are sufficient to communicate key trends. 
However, in more expansive multiverses novel visualization techniques will become essential.

Recent experiments with pre-registration in ML (e.g.\ \citep{PreregistrationWorkshop}) promise to enforce a distinction between exploratory and confirmatory research \citep{Wagenmakers2012, Nosek2018}.
Committing to an experimentation and analysis plan in advance can be a helpful foil for questionable practices such as tweaking the parameters until one reaches a positive result.
However, this commitment is limited to a single analysis---one particular instantiation of a decision set---while the other possible analyses, the remainder of the multiverse, remain unexplored.
We see the multiverse analysis as complementary to pre-registration, in that pre-registering a multiverse analysis both codifies auxiliary hypotheses and allows room for exploration.

\section{Conclusion}

For continued progress in ML, we depend on reproducible results and conclusions that generalize to new settings.
We have introduced the multiverse analysis as a principled framework to explore the impact of researcher decisions and facilitate the drawing of more robust conclusions.
Our first case study unifies conflicting work on the benefit of adaptive optimizers and reinforces that optimizer merit is driven primarily by learning rate.
In our second case study we identified a complex interaction between batch size, learning rate and generalization error, and we dispute the existence of a generalization gap.
We have also presented evidence supporting the practice of scaling learning rate linearly with batch size.

By using a multiverse analysis, both researchers and practitioners gain more robust claims, better understanding of how decisions impact results, and helpful insight into the generality and reproducibility of conclusions.



\begin{ack}
SB is supported by the Biotechnology and Biological Sciences Research Council [grant number BBSRC BB/M011194/1].
NL is supported by a Senior Turing AI Fellowship funded by the UK government’s Office for AI, through UK Research and Innovation (grant reference EP/V030302/1), and delivered by the Alan Turing Institute.
NL's chair is endowed by DeepMind.
\end{ack}


\small
\bibliography{multiverse-main}

\begin{thebibliography}{85}
\expandafter\ifx\csname natexlab\endcsname\relax\def\natexlab#1{#1}\fi
\expandafter\ifx\csname url\endcsname\relax
  \def\url#1{{\tt #1}}\fi

\bibitem[Steegen et~al.(2016)Steegen, Tuerlinckx, Gelman, and
  Vanpaemel]{Steegen2016}
Sara Steegen, Francis Tuerlinckx, Andrew Gelman, and Wolf Vanpaemel.
\newblock Increasing transparency through a multiverse analysis.
\newblock {\em Perspectives on Psychological Science}, 11\penalty0
  (5):\penalty0 702--712, 2016.

\bibitem[Gundersen and Kjensmo(2018)]{Gundersen2018}
Odd~Erik Gundersen and Sigbj{\o}rn Kjensmo.
\newblock State of the art: Reproducibility in artificial intelligence.
\newblock In {\em 32nd AAAI Conference on Artificial Intelligence, AAAI 2018},
  pages 1644--1651, 2018.

\bibitem[Bouthillier et~al.(2019)Bouthillier, Laurent, and
  Vincent]{Bouthillier2019}
Xavier Bouthillier, C{\'{e}}sar Laurent, and Pascal Vincent.
\newblock Unreproducible research is reproducible.
\newblock In {\em Proceedings of the 36th International Conference on Machine
  Learning, ICML}, pages 1150--1159, 2019.

\bibitem[Lipton and Steinhardt(2019)]{Lipton2019}
Zachary~C. Lipton and Jacob Steinhardt.
\newblock Troubling trends in machine-learning scholarship.
\newblock {\em Queue}, 17\penalty0 (1):\penalty0 45–77, 2019.

\bibitem[Sculley et~al.(2018)Sculley, Snoek, Rahimi, and
  Wiltschko]{Sculley2018}
D.~Sculley, Jasper Snoek, Ali Rahimi, and Alex Wiltschko.
\newblock {Winner's curse? On pace, progress, and empirical rigor}.
\newblock In {\em Proceedings of the 6th International Conference on Learning
  Representations, ICLR 2018}, 2018.

\bibitem[Feder~Cooper et~al.(2021)Feder~Cooper, Lu, Forde, and
  De~Sa]{Cooper2021}
A.~Feder~Cooper, Yucheng Lu, Jessica Forde, and Christopher~M. De~Sa.
\newblock Hyperparameter optimization is deceiving us, and how to stop it.
\newblock In {\em Advances in Neural Information Processing Systems},
  volume~34, 2021.

\bibitem[Raff(2019)]{Raff2019}
Edward Raff.
\newblock A step toward quantifying independently reproducible machine learning
  research.
\newblock In {\em Advances in Neural Information Processing Systems},
  volume~32, 2019.

\bibitem[Forde and Paganini(2019)]{Forde2019}
Jessica Forde and Michela Paganini.
\newblock The scientific method in the science of machine learning.
\newblock 2019, arXiv:1904.10922.

\bibitem[Rahimi and Recht(2017)]{Rahimi2017}
Ali Rahimi and Benjamin Recht.
\newblock Reflections on random kitchen sinks, 2017.
\newblock URL \url{https://www.argmin.net/2017/12/05/kitchen-sinks/}.
\newblock Accessed 2022-04-28.

\bibitem[Henderson et~al.(2018)Henderson, Islam, Bachman, Pineau, Precup, and
  Meger]{Henderson2017}
Peter Henderson, Riashat Islam, Philip Bachman, Joelle Pineau, Doina Precup,
  and David Meger.
\newblock Deep reinforcement learning that matters.
\newblock In {\em Proceedings of the AAAI Conference on Artificial
  Intelligence}, volume~32, 2018.

\bibitem[Agarwal et~al.(2021)Agarwal, Schwarzer, Castro, Courville, and
  Bellemare]{Agarwal2021}
Rishabh Agarwal, Max Schwarzer, Pablo~Samuel Castro, Aaron~C. Courville, and
  Marc Bellemare.
\newblock Deep reinforcement learning at the edge of the statistical precipice.
\newblock In {\em Advances in Neural Information Processing Systems},
  volume~34, 2021.

\bibitem[Lu\v{c}i\'{c} et~al.(2018)Lu\v{c}i\'{c}, Kurach, Michalski, Gelly, and
  Bousquet]{Lucic2018}
Mario Lu\v{c}i\'{c}, Karol Kurach, Marcin Michalski, Sylvain Gelly, and Olivier
  Bousquet.
\newblock Are {GANs} created equal? {A} large-scale study.
\newblock In {\em Advances in Neural Information Processing Systems},
  volume~31, 2018.

\bibitem[Dacrema et~al.(2019)Dacrema, Cremonesi, and Jannach]{Dacrema2019}
Maurizio~Ferrari Dacrema, Paolo Cremonesi, and Dietmar Jannach.
\newblock Are we really making much progress? {A} worrying analysis of recent
  neural recommendation approaches.
\newblock In {\em Proceedings of the 13th ACM Conference on Recommender
  Systems, RecSys}, page 101–109, 2019.

\bibitem[Schmidt et~al.(2021)Schmidt, Schneider, and Hennig]{Schmidt2021}
Robin~M. Schmidt, Frank Schneider, and Philipp Hennig.
\newblock Descending through a crowded valley-benchmarking deep learning
  optimizers.
\newblock In {\em Proceedings of the 38th International Conference on Machine
  Learning, ICML}, pages 9367--9376. PMLR, 2021.

\bibitem[Oakden-Rayner et~al.(2019)Oakden-Rayner, Dunnmon, Carneiro, and
  Ré]{Oakden-Rayner2019}
Luke Oakden-Rayner, Jared Dunnmon, Gustavo Carneiro, and Christopher Ré.
\newblock Hidden stratification causes clinically meaningful failures in
  machine learning for medical imaging.
\newblock 2019, arXiv:1909.12475.

\bibitem[Melis et~al.(2018)Melis, Dyer, and Blunsom]{Melis2018}
G{\'{a}}bor Melis, Chris Dyer, and Phil Blunsom.
\newblock On the state of the art of evaluation in neural language models.
\newblock In {\em Proceedings of the 6th International Conference on Learning
  Representations, ICLR 2018}, 2018, arXiv:1707.05589.

\bibitem[Reimers and Gurevych(2017)]{Reimers2017}
Nils Reimers and Iryna Gurevych.
\newblock Reporting score distributions makes a difference: Performance study
  of {LSTM}-networks for sequence tagging.
\newblock 2017, arXiv:1707.09861.

\bibitem[Gorman and Bedrick(2019)]{Gorman2019}
Kyle Gorman and Steven Bedrick.
\newblock We need to talk about standard splits.
\newblock In {\em Proceedings of the 57th Annual Meeting of the Association for
  Computational Linguistics}, pages 2786--2791, 2019.

\bibitem[Marie et~al.(2021)Marie, Fujita, and Rubino]{Marie2021}
Benjamin Marie, Atsushi Fujita, and Raphael Rubino.
\newblock Scientific credibility of machine translation research: A
  meta-evaluation of 769 papers.
\newblock 2021, arXiv:2106.15195.

\bibitem[Narang et~al.(2021)Narang, Chung, Tay, Fedus, Fevry, Matena, Malkan,
  Fiedel, Shazeer, and Lan]{Narang2021}
Sharan Narang, Hyung~Won Chung, Yi~Tay, William Fedus, Thibault Fevry, Michael
  Matena, Karishma Malkan, Noah Fiedel, Noam Shazeer, and Zhenzhong Lan.
\newblock Do transformer modifications transfer across implementations and
  applications?
\newblock 2021, arXiv:2102.11972.

\bibitem[Bell and Kampman(2021)]{Bell2021}
Samuel~J. Bell and Onno~P. Kampman.
\newblock Perspectives on machine learning from psychology's reproducibility
  crisis.
\newblock In {\em Science and Engineering of Deep Learning Workshop, ICLR},
  2021, arXiv:2104.08878.

\bibitem[John et~al.(2012)John, Loewenstein, and Prelec]{John2012}
Leslie~K. John, George Loewenstein, and Drazen Prelec.
\newblock Measuring the prevalence of questionable research practices with
  incentives for truth telling.
\newblock {\em Psychological Science}, 23\penalty0 (5):\penalty0 524--532,
  2012.

\bibitem[Wagenmakers et~al.(2012)Wagenmakers, Wetzels, Borsboom, van~der Maas,
  and Kievit]{Wagenmakers2012}
Eric-Jan Wagenmakers, Ruud Wetzels, Denny Borsboom, Han L.~J. van~der Maas, and
  Rogier~A. Kievit.
\newblock An agenda for purely confirmatory research.
\newblock {\em Perspectives on Psychological Science}, 7\penalty0 (6):\penalty0
  632--638, 2012.

\bibitem[Chambers(2013)]{Chambers2013}
Christopher~D. Chambers.
\newblock Registered reports: A new publishing initiative at {Cortex}.
\newblock {\em Cortex}, 49\penalty0 (3):\penalty0 609--610, 2013.

\bibitem[Klein et~al.(2014)Klein, Ratliff, Vianello, Adams, Bahn{\'{i}}k,
  Bernstein, Bocian, Brandt, Brooks, Brumbaugh, Cemalcilar, Chandler, Cheong,
  Davis, Devos, Eisner, Frankowska, Furrow, Galliani, Hasselman, Hicks,
  Hovermale, Hunt, Huntsinger, Ijzerman, John, Joy-Gaba, Kappes, Krueger,
  Kurtz, Levitan, Mallett, Morris, Nelson, Nier, Packard, Pilati, Rutchick,
  Schmidt, Skorinko, Smith, Steiner, Storbeck, {Van Swol}, Thompson, {Van 'T
  Veer}, Vaughn, Vranka, Wichman, Woodzicka, and Nosek]{Klein2014}
Richard~A. Klein, Kate~A. Ratliff, Michelangelo Vianello, Reginald~B. Adams,
  {\v{S}}t{\v{e}}p{\'{a}}n Bahn{\'{i}}k, Michael~J. Bernstein, Konrad Bocian,
  Mark~J. Brandt, Beach Brooks, Claudia~Chloe Brumbaugh, Zeynep Cemalcilar,
  Jesse Chandler, Winnee Cheong, William~E. Davis, Thierry Devos, Matthew
  Eisner, Natalia Frankowska, David Furrow, Elisa~Maria Galliani, Fred
  Hasselman, Joshua~A. Hicks, James~F. Hovermale, S.~Jane Hunt, Jeffrey~R.
  Huntsinger, Hans Ijzerman, Melissa~Sue John, Jennifer~A. Joy-Gaba,
  Heather~Barry Kappes, Lacy~E. Krueger, Jaime Kurtz, Carmel~A. Levitan,
  Robyn~K. Mallett, Wendy~L. Morris, Anthony~J. Nelson, Jason~A. Nier, Grant
  Packard, Ronaldo Pilati, Abraham~M. Rutchick, Kathleen Schmidt, Jeanine~L.
  Skorinko, Robert Smith, Troy~G. Steiner, Justin Storbeck, Lyn~M. {Van Swol},
  Donna Thompson, A.~E. {Van 'T Veer}, Leigh~Ann Vaughn, Marek Vranka, Aaron~L.
  Wichman, Julie~A. Woodzicka, and Brian~A. Nosek.
\newblock Investigating variation in replicability: A ``{Many Labs}''
  replication project.
\newblock {\em Social Psychology}, 45\penalty0 (3):\penalty0 142--152, 2014.

\bibitem[Simmons et~al.(2011)Simmons, Nelson, and Simonsohn]{Simmons2011}
Joseph~P. Simmons, Leif~D. Nelson, and Uri Simonsohn.
\newblock False-positive psychology: Undisclosed flexibility in data collection
  and analysis allows presenting anything as significant.
\newblock {\em Psychological Science}, 22\penalty0 (11):\penalty0 1359--1366,
  2011.

\bibitem[Gelman and Loken(2014)]{Gelman2014}
Andrew Gelman and Eric Loken.
\newblock The statistical crisis in science: data-dependent analysis---a
  ``garden of forking paths''---explains why many statistically significant
  comparisons don't hold up.
\newblock {\em American Scientist}, 102\penalty0 (6):\penalty0 460--466, 2014.

\bibitem[Durante et~al.(2013)Durante, Rae, and Griskevicius]{Durante2013}
Kristina~M Durante, Ashley Rae, and Vladas Griskevicius.
\newblock The fluctuating female vote: Politics, religion, and the ovulatory
  cycle.
\newblock {\em Psychological Science}, 24\penalty0 (6):\penalty0 1007--1016,
  2013.

\bibitem[Donnelly et~al.(2019)Donnelly, Brooks, and Homer]{Donnelly2019}
Seamus Donnelly, Patricia~J Brooks, and Bruce~D Homer.
\newblock Is there a bilingual advantage on interference-control tasks? {A}
  multiverse meta-analysis of global reaction time and interference cost.
\newblock {\em Psychonomic bulletin \& review}, 26:\penalty0 1122--1147, 2019.

\bibitem[Kalokerinos et~al.(2019)Kalokerinos, Erbas, Ceulemans, and
  Kuppens]{Kalokerinos2019}
Elise~K. Kalokerinos, Yasemin Erbas, Eva Ceulemans, and Peter Kuppens.
\newblock Differentiate to regulate: Low negative emotion differentiation is
  associated with ineffective use but not selection of emotion-regulation
  strategies.
\newblock {\em Psychological Science}, 30\penalty0 (6):\penalty0 863--879,
  2019.

\bibitem[Modecki et~al.(2020)Modecki, Low-Choy, Uink, Vernon, Correia, and
  Andrews]{Modecki2020}
Kathryn~L. Modecki, Samantha Low-Choy, Bep~N. Uink, Lynette Vernon, Helen
  Correia, and Kylie Andrews.
\newblock Tuning into the real effect of smartphone use on parenting: a
  multiverse analysis.
\newblock {\em Journal of Child Psychology and Psychiatry}, 61\penalty0
  (8):\penalty0 855--865, 2020.

\bibitem[Lonsdorf et~al.(2022)Lonsdorf, Gerlicher, Klingelh{\"o}fer-Jens, and
  Krypotos]{Lonsdorf2022}
Tina~B. Lonsdorf, Anna Gerlicher, Maren Klingelh{\"o}fer-Jens, and
  Angelos-Miltiadis Krypotos.
\newblock Multiverse analyses in fear conditioning research.
\newblock {\em Behaviour Research and Therapy}, 153:\penalty0 104072, 2022.

\bibitem[Dafflon et~al.(2022)Dafflon, F~Da~Costa, V{\'a}{\v{s}}a, Monti, Bzdok,
  Hellyer, Turkheimer, Smallwood, Jones, and Leech]{Dafflon2022}
Jessica Dafflon, Pedro F~Da~Costa, Franti{\v{s}}ek V{\'a}{\v{s}}a, Ricardo~Pio
  Monti, Danilo Bzdok, Peter~J Hellyer, Federico Turkheimer, Jonathan
  Smallwood, Emily Jones, and Robert Leech.
\newblock A guided multiverse study of neuroimaging analyses.
\newblock {\em Nature {C}ommunications}, 13\penalty0 (3758), 2022.

\bibitem[Ioffe and Szegedy(2015)]{Ioffe2015}
Sergey Ioffe and Christian Szegedy.
\newblock Batch normalization: Accelerating deep network training by reducing
  internal covariate shift.
\newblock In {\em Proceedings of the 32nd International Conference on Machine
  Learning, ICML}, pages 448--456, 2015.

\bibitem[Wilson et~al.(2017)Wilson, Roelofs, Stern, Srebro, and
  Recht]{Wilson2017}
Ashia~C. Wilson, Rebecca Roelofs, Mitchell Stern, Nati Srebro, and Benjamin
  Recht.
\newblock The marginal value of adaptive gradient methods in machine learning.
\newblock In {\em Advances in Neural Information Processing Systems},
  volume~30, 2017.

\bibitem[Keskar et~al.(2017)Keskar, Nocedal, Tang, Mudigere, and
  Smelyanskiy]{Keskar2017}
Nitish~Shirish Keskar, Jorge Nocedal, Ping Tak~Peter Tang, Dheevatsa Mudigere,
  and Mikhail Smelyanskiy.
\newblock {On large-batch training for deep learning: Generalization gap and
  sharp minima}.
\newblock In {\em Proceedings of the 5th International Conference on Learning
  Representations, ICLR 2017}, 2017, arXiv:1609.04836.

\bibitem[Chaloner and Verdinelli(1995)]{Chaloner1995}
Kathryn Chaloner and Isabella Verdinelli.
\newblock Bayesian experimental design: A review.
\newblock {\em Statistical Science}, pages 273--304, 1995.

\bibitem[Bergstra and Bengio(2012)]{Bergstra2012}
James Bergstra and Yoshua Bengio.
\newblock Random search for hyper-parameter optimization.
\newblock {\em {Journal of Machine Learning Research}}, 13:\penalty0 281--305,
  2012.

\bibitem[Williams and Rasmussen(2006)]{Williams2006}
Christopher~K. Williams and Carl~Edward Rasmussen.
\newblock {\em Gaussian Processes for Machine Learning}.
\newblock The MIT Press, 2006.
\newblock ISBN 978-0-262-18253-9.

\bibitem[Snoek et~al.(2012)Snoek, Larochelle, and Adams]{Snoek2012}
Jasper Snoek, Hugo Larochelle, and Ryan~P Adams.
\newblock Practical bayesian optimization of machine learning algorithms.
\newblock In {\em Advances in Neural Information Processing Systems},
  volume~25, 2012.

\bibitem[Sacks et~al.(1989)Sacks, Welch, Mitchell, and Wynn]{Sacks1989}
Jerome Sacks, William~J Welch, Toby~J Mitchell, and Henry~P Wynn.
\newblock Design and analysis of computer experiments.
\newblock {\em Statistical Science}, 4\penalty0 (4):\penalty0 409--423, 1989.

\bibitem[Sobol(2001)]{Sobol2001}
I.~M. Sobol.
\newblock Global sensitivity indices for nonlinear mathematical models and
  their {Monte Carlo} estimates.
\newblock {\em Mathematics and Computers in Simulation}, 55\penalty0
  (1):\penalty0 271--280, 2001.

\bibitem[Saltelli(2002)]{Saltelli2002}
Andrea Saltelli.
\newblock Making best use of model evaluations to compute sensitivity indices.
\newblock {\em Computer Physics Communications}, 145\penalty0 (2):\penalty0
  280--297, 2002.

\bibitem[GPy(2012)]{GPy2012}
{GPy: A {Gaussian Process} framework in python}.
\newblock \url{http://github.com/SheffieldML/GPy}, 2012.

\bibitem[Paleyes et~al.(2019)Paleyes, Pullin, Mahsereci, McCollum, Lawrence,
  and González]{Emukit2019}
Andrei Paleyes, Mark Pullin, Maren Mahsereci, Cliff McCollum, Neil Lawrence,
  and Javier González.
\newblock Emulation of physical processes with {Emukit}.
\newblock In {\em 2nd Workshop on Machine Learning and the Physical Sciences,
  NeurIPS}, 2019.

\bibitem[Paszke et~al.(2017)Paszke, Gross, Chintala, Chanan, Yang, DeVito, Lin,
  Desmaison, Antiga, and Lerer]{Paszke2017}
Adam Paszke, Sam Gross, Soumith Chintala, Gregory Chanan, Edward Yang, Zachary
  DeVito, Zeming Lin, Alban Desmaison, Luca Antiga, and Adam Lerer.
\newblock Automatic differentiation in {PyTorch}.
\newblock In {\em Autodiff Workshop, NeurIPS}, 2017.

\bibitem[Pineau et~al.(2020)Pineau, Vincent-Lamarre, Sinha, Larivière,
  Beygelzimer, d'Alché Buc, Fox, and Larochelle]{Pineau2020}
Joelle Pineau, Philippe Vincent-Lamarre, Koustuv Sinha, Vincent Larivière,
  Alina Beygelzimer, Florence d'Alché Buc, Emily Fox, and Hugo Larochelle.
\newblock Improving reproducibility in machine learning research (a report from
  the {NeurIPS} 2019 reproducibility program).
\newblock 2020, arXiv:2003.12206.

\bibitem[Cox and John(1992)]{Cox1992}
D.~D. Cox and S.~John.
\newblock A statistical method for global optimization.
\newblock In {\em {Proceedings of the 1992 IEEE International Conference on
  Systems, Man, and Cybernetics}}, volume~2, pages 1241--1246, 1992.

\bibitem[Wolberg et~al.(1995)Wolberg, Street, and Mangasarian]{Wolberg1995}
William Wolberg, W.~Street, and Olvi Mangasarian.
\newblock {Breast Cancer Wisconsin (Diagnostic)}.
\newblock UCI Machine Learning Repository, 1995.

\bibitem[Kingma and Ba(2014)]{Kingma2014}
Diederik~P. Kingma and Jimmy Ba.
\newblock Adam: A method for stochastic optimization.
\newblock 2014, arXiv:1412.6980.

\bibitem[Robbins and Monro(1951)]{Robbins1951}
Herbert Robbins and Sutton Monro.
\newblock A stochastic approximation method.
\newblock {\em The annals of mathematical statistics}, pages 400--407, 1951.

\bibitem[Polyak(1964)]{Polyak1964}
Boris~T. Polyak.
\newblock Some methods of speeding up the convergence of iteration methods.
\newblock {\em {USSR} computational mathematics and mathematical physics},
  4\penalty0 (5):\penalty0 1--17, 1964.

\bibitem[Krizhevsky(2009)]{Krizhevsky2009}
Alex Krizhevsky.
\newblock Learning multiple layers of features from tiny images.
\newblock Master's thesis, University of Toronto, 2009.

\bibitem[Choi et~al.(2019)Choi, Shallue, Nado, Lee, Maddison, and
  Dahl]{Choi2019}
Dami Choi, Christopher~J. Shallue, Zachary Nado, Jaehoon Lee, Chris~J.
  Maddison, and George~E. Dahl.
\newblock On empirical comparisons of optimizers for deep learning, 2019,
  arXiv:1910.05446.

\bibitem[Simonyan and Zisserman(2014)]{Simonyan2014}
Karen Simonyan and Andrew Zisserman.
\newblock Very deep convolutional networks for large-scale image recognition,
  2014, arXiv:1409.1556.

\bibitem[Srivastava et~al.(2014)Srivastava, Hinton, Krizhevsky, Sutskever, and
  Salakhutdinov]{Srivastava2014}
Nitish Srivastava, Geoffrey Hinton, Alex Krizhevsky, Ilya Sutskever, and Ruslan
  Salakhutdinov.
\newblock Dropout: a simple way to prevent neural networks from overfitting.
\newblock {\em {Journal of Machine Learning Research}}, 15\penalty0
  (1):\penalty0 1929--1958, 2014.

\bibitem[Golmant et~al.(2018)Golmant, Vemuri, Yao, Feinberg, Gholami, Rothauge,
  Mahoney, and Gonzalez]{Golmant2018}
Noah Golmant, Nikita Vemuri, Zhewei Yao, Vladimir Feinberg, Amir Gholami, Kai
  Rothauge, Michael~W. Mahoney, and Joseph Gonzalez.
\newblock On the computational inefficiency of large batch sizes for stochastic
  gradient descent.
\newblock 2018, arXiv:1811.12941.

\bibitem[Masters and Luschi(2018)]{Masters2018}
Dominic Masters and Carlo Luschi.
\newblock Revisiting small batch training for deep neural networks.
\newblock 2018, arXiv:1804.07612.

\bibitem[Hochreiter and Schmidhuber(1997)]{Hochreiter1997}
Sepp Hochreiter and J{\"{u}}rgen Schmidhuber.
\newblock {Flat minima}.
\newblock {\em Neural Computation}, 9\penalty0 (1):\penalty0 1--42, 1997.

\bibitem[Li et~al.(2018)Li, Xu, Taylor, Studer, and Goldstein]{Li2017}
Hao Li, Zheng Xu, Gavin Taylor, Christoph Studer, and Tom Goldstein.
\newblock Visualizing the loss landscape of neural nets.
\newblock In {\em Advances in Neural Information Processing Systems},
  volume~31, 2018.

\bibitem[Jastrz{\k{e}}bski et~al.(2017)Jastrz{\k{e}}bski, Kenton, Arpit,
  Ballas, Fischer, Bengio, and Storkey]{Jastrzkebski2017}
Stanis{\l}aw Jastrz{\k{e}}bski, Zachary Kenton, Devansh Arpit, Nicolas Ballas,
  Asja Fischer, Yoshua Bengio, and Amos Storkey.
\newblock Three factors influencing minima in {SGD}.
\newblock 2017, arXiv:1711.04623.

\bibitem[Chaudhari et~al.(2019)Chaudhari, Choromanska, Soatto, LeCun, Baldassi,
  Borgs, Chayes, Sagun, and Zecchina]{Chaudhari2019}
Pratik Chaudhari, Anna Choromanska, Stefano Soatto, Yann LeCun, Carlo Baldassi,
  Christian Borgs, Jennifer Chayes, Levent Sagun, and Riccardo Zecchina.
\newblock Entropy-{SGD}: Biasing gradient descent into wide valleys.
\newblock {\em Journal of Statistical Mechanics: Theory and Experiment},
  2019\penalty0 (12):\penalty0 124018, 2019.

\bibitem[Dinh et~al.(2017)Dinh, Pascanu, Bengio, and Bengio]{Dinh2017}
Laurent Dinh, Razvan Pascanu, Samy Bengio, and Yoshua Bengio.
\newblock Sharp minima can generalize for deep nets.
\newblock In {\em Proceedings of the 34th International Conference on Machine
  Learning, ICML}, 2017.

\bibitem[Devarakonda et~al.(2017)Devarakonda, Naumov, and
  Garland]{Devarakonda2017}
Aditya Devarakonda, Maxim Naumov, and Michael Garland.
\newblock {AdaBatch}: Adaptive batch sizes for training deep neural networks.
\newblock 2017, arXiv:1712.02029.

\bibitem[Bottou et~al.(2018)Bottou, Curtis, and Nocedal]{Bottou2018}
L{\'e}on Bottou, Frank~E. Curtis, and Jorge Nocedal.
\newblock Optimization methods for large-scale machine learning.
\newblock {\em SIAM Review}, 60\penalty0 (2):\penalty0 223--311, 2018.

\bibitem[Hoffer et~al.(2017)Hoffer, Hubara, and Soudry]{Hoffer2017}
Elad Hoffer, Itay Hubara, and Daniel Soudry.
\newblock Train longer, generalize better: Closing the generalization gap in
  large batch training of neural networks.
\newblock In {\em Advances in Neural Information Processing Systems},
  volume~30, 2017.

\bibitem[Smith et~al.(2017)Smith, Kindermans, Ying, and Le]{Smith2017}
Samuel~L. Smith, Pieter-Jan Kindermans, Chris Ying, and Quoc~V. Le.
\newblock Don't decay the learning rate, increase the batch size.
\newblock 2017, arXiv:1711.00489.

\bibitem[Goyal et~al.(2017)Goyal, Doll{\'a}r, Girshick, Noordhuis, Wesolowski,
  Kyrola, Tulloch, Jia, and He]{Goyal2017}
Priya Goyal, Piotr Doll{\'a}r, Ross Girshick, Pieter Noordhuis, Lukasz
  Wesolowski, Aapo Kyrola, Andrew Tulloch, Yangqing Jia, and Kaiming He.
\newblock Accurate, large minibatch {SGD}: Training {ImageNet} in 1 hour.
\newblock 2017, arXiv:1706.02677.

\bibitem[You et~al.(2017)You, Gitman, and Ginsburg]{You2017}
Yang You, Igor Gitman, and Boris Ginsburg.
\newblock Large batch training of convolutional networks.
\newblock 2017, arXiv:1708.03888.

\bibitem[Shallue et~al.(2019)Shallue, Lee, Antognini, Sohl-Dickstein, Frostig,
  and Dahl]{Shallue2019}
Christopher~J. Shallue, Jaehoon Lee, Joseph Antognini, Jascha Sohl-Dickstein,
  Roy Frostig, and George~E. Dahl.
\newblock Measuring the effects of data parallelism on neural network training.
\newblock {\em {Journal of Machine Learning Research}}, 20:\penalty0 1--49,
  2019.

\bibitem[Krizhevsky et~al.(2012)Krizhevsky, Sutskever, and
  Hinton]{Krizhevsky2012}
Alex Krizhevsky, Ilya Sutskever, and Geoffrey~E. Hinton.
\newblock {ImageNet} classification with deep convolutional neural networks.
\newblock In {\em Advances in Neural Information Processing Systems},
  volume~25, 2012.

\bibitem[He et~al.(2016)He, Zhang, Ren, and Sun]{He2016}
Kaiming He, Xiangyu Zhang, Shaoqing Ren, and Jian Sun.
\newblock Deep residual learning for image recognition.
\newblock In {\em Proceedings of the IEEE Conference on Computer Vision and
  Pattern Recognition}, pages 770--778, 2016.

\bibitem[Tin()]{TinyImageNet}
{Tiny ImageNet}.
\newblock \url{https://tiny-imagenet.herokuapp.com/}.

\bibitem[Russakovsky et~al.(2015)Russakovsky, Deng, Su, Krause, Satheesh, Ma,
  Huang, Karpathy, Khosla, and Bernstein]{Russakovsky2015}
Olga Russakovsky, Jia Deng, Hao Su, Jonathan Krause, Sanjeev Satheesh, Sean Ma,
  Zhiheng Huang, Andrej Karpathy, Aditya Khosla, and Michael Bernstein.
\newblock {ImageNet} large scale visual recognition challenge.
\newblock {\em International Journal of Computer Vision}, 115\penalty0
  (3):\penalty0 211--252, 2015.

\bibitem[Helterbrand and Cressie(1994)]{Helterbrand1994}
Jeffrey~D. Helterbrand and Noel Cressie.
\newblock Universal cokriging under intrinsic coregionalization.
\newblock {\em Mathematical Geology}, 26\penalty0 (2):\penalty0 205--226, 1994.

\bibitem[Alvarez et~al.(2012)Alvarez, Rosasco, and Lawrence]{Alvarez2012}
Mauricio~A. Alvarez, Lorenzo Rosasco, and Neil~D. Lawrence.
\newblock Kernels for vector-valued functions: A review.
\newblock {\em Foundations and Trends in Machine Learning}, 4\penalty0
  (3):\penalty0 195--266, 2012.

\bibitem[Lacoste et~al.(2019)Lacoste, Luccioni, Schmidt, and
  Dandres]{Lacoste2019}
Alexandre Lacoste, Alexandra Luccioni, Victor Schmidt, and Thomas Dandres.
\newblock Quantifying the carbon emissions of machine learning.
\newblock 2019, arXiv:1910.09700.

\bibitem[Strubell et~al.(2019)Strubell, Ganesh, and McCallum]{Strubell2019}
Emma Strubell, Ananya Ganesh, and Andrew McCallum.
\newblock Energy and policy considerations for deep learning in {NLP}.
\newblock 2019, arXiv:1906.02243.

\bibitem[Schwartz et~al.(2020)Schwartz, Dodge, Smith, and
  Etzioni]{Schwartz2020}
Roy Schwartz, Jesse Dodge, Noah~A. Smith, and Oren Etzioni.
\newblock {Green AI}.
\newblock {\em Communications of the ACM}, 63\penalty0 (12):\penalty0 54--63,
  2020.

\bibitem[Goldberg et~al.(1997)Goldberg, Williams, and Bishop]{Goldberg1997}
Paul Goldberg, Christopher Williams, and Christopher Bishop.
\newblock Regression with input-dependent noise: A {Gaussian Process}
  treatment.
\newblock volume~10, 1997.

\bibitem[Binois et~al.(2018)Binois, Gramacy, and Ludkovski]{Binois2018}
Micka\"{e}l Binois, Robert~B. Gramacy, and Mike Ludkovski.
\newblock Practical heteroscedastic {Gaussian Process} modeling for large
  simulation experiments.
\newblock {\em Journal of Computational and Graphical Statistics}, 27\penalty0
  (4):\penalty0 808--821, 2018.

\bibitem[Binois et~al.(2019)Binois, Huang, Gramacy, and Ludkovski]{Binois2019}
Micka{\"e}l Binois, Jiangeng Huang, Robert~B. Gramacy, and Mike Ludkovski.
\newblock Replication or exploration? {Sequential} design for stochastic
  simulation experiments.
\newblock {\em Technometrics}, 61\penalty0 (1):\penalty0 7--23, 2019.

\bibitem[Dragicevic et~al.(2019)Dragicevic, Jansen, Sarma, Kay, and
  Chevalier]{Dragicevic2019}
Pierre Dragicevic, Yvonne Jansen, Abhraneel Sarma, Matthew Kay, and Fanny
  Chevalier.
\newblock Increasing the transparency of research papers with explorable
  multiverse analyses.
\newblock In {\em Proceedings of the 2019 Conference on Human Factors in
  Computing Systems}, pages 1--15, 2019.

\bibitem[Pre()]{PreregistrationWorkshop}
{The pre-registration workshop}.
\newblock \url{https://preregister.science/}.

\bibitem[Nosek et~al.(2018)Nosek, Ebersole, DeHaven, and Mellor]{Nosek2018}
Brian~A. Nosek, Charles~R. Ebersole, Alexander~C. DeHaven, and David~T. Mellor.
\newblock The preregistration revolution.
\newblock {\em Proceedings of the National Academy of Sciences of the United
  States of America}, 115\penalty0 (11):\penalty0 2600--2606, 2018.

\end{thebibliography}
\normalsize

\section*{Checklist}

\begin{enumerate}

\item For all authors...
\begin{enumerate}
  \item Do the main claims made in the abstract and introduction accurately reflect the paper's contributions and scope?
    \answerYes{}
  \item Did you describe the limitations of your work?
    \answerYes{See \cref{sec:discussion}.}
  \item Did you discuss any potential negative societal impacts of your work?
    \answerYes{See \cref{sec:compute-cost} for compute cost/climate impact discussion.}
  \item Have you read the ethics review guidelines and ensured that your paper conforms to them?
    \answerYes{}
\end{enumerate}

\item If you are including theoretical results...
\begin{enumerate}
  \item Did you state the full set of assumptions of all theoretical results?
    \answerNA{}
        \item Did you include complete proofs of all theoretical results?
    \answerNA{}
\end{enumerate}

\item If you ran experiments...
\begin{enumerate}
  \item Did you include the code, data, and instructions needed to reproduce the main experimental results (either in the supplemental material or as a URL)?
    \answerYes{To be included in supplementary materials.}
  \item Did you specify all the training details (e.g., data splits, hyperparameters, how they were chosen)?
    \answerYes{}
        \item Did you report error bars (e.g., with respect to the random seed after running experiments multiple times)?
    \answerYes{}
        \item Did you include the total amount of compute and the type of resources used (e.g., type of GPUs, internal cluster, or cloud provider)?
    \answerYes{See \cref{sec:compute-cost}.}
\end{enumerate}

\item If you are using existing assets (e.g., code, data, models) or curating/releasing new assets...
\begin{enumerate}
  \item If your work uses existing assets, did you cite the creators?
    \answerYes{All datasets and models cited throughout.}
  \item Did you mention the license of the assets?
    \answerNo{}
  \item Did you include any new assets either in the supplemental material or as a URL?
    \answerNo{}
  \item Did you discuss whether and how consent was obtained from people whose data you're using/curating?
    \answerNA{}
  \item Did you discuss whether the data you are using/curating contains personally identifiable information or offensive content?
    \answerNo{}
\end{enumerate}

\item If you used crowdsourcing or conducted research with human subjects...
\begin{enumerate}
  \item Did you include the full text of instructions given to participants and screenshots, if applicable?
    \answerNA{}
  \item Did you describe any potential participant risks, with links to Institutional Review Board (IRB) approvals, if applicable?
    \answerNA{}
  \item Did you include the estimated hourly wage paid to participants and the total amount spent on participant compensation?
    \answerNA{}
\end{enumerate}

\end{enumerate}



\appendix

\pagebreak
\FloatBarrier

\section*{Appendix}

\renewcommand{\thefigure}{S\arabic{figure}}
\setcounter{figure}{0}


\begin{figure}[h]
    \centering
    \subfloat[Shared]{
        \includegraphics[trim={0 0 10cm 0},clip,height=5cm]{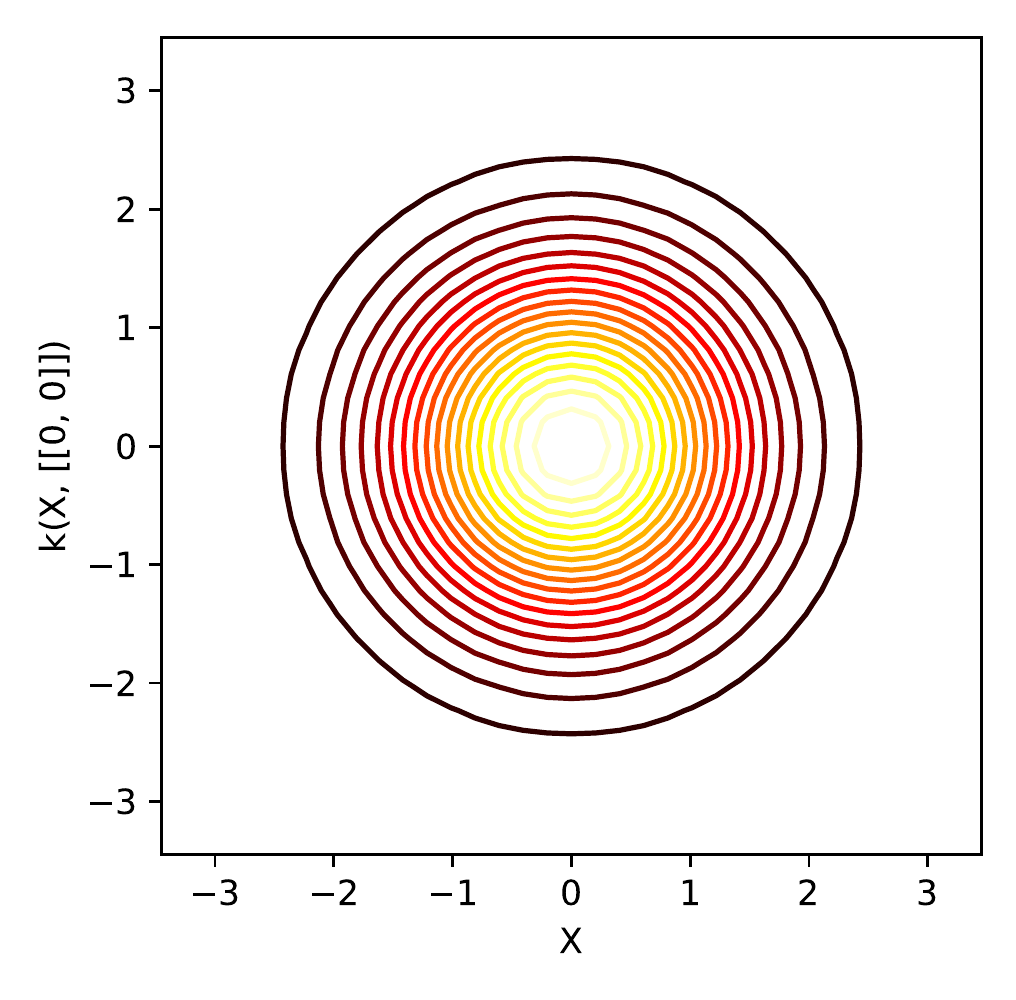}
    }\quad
    \subfloat[Additive]{
        \includegraphics[trim={10cm 0 0 0},clip,height=5cm]{kernel-examples}
    } 
    \caption{Illustration of \textbf{(a)} shared and \textbf{(b)} additive kernel. We use the additive kernel to model the absence of interaction effect, as it computes similarity along either (rather then both) dimension.}\label{fig:kernel-examples}
\end{figure}

\begin{figure}[h]
    \centering
    \subfloat[Learning rate]{
        \includegraphics[trim={0 0 17cm 0},clip,height=5cm]{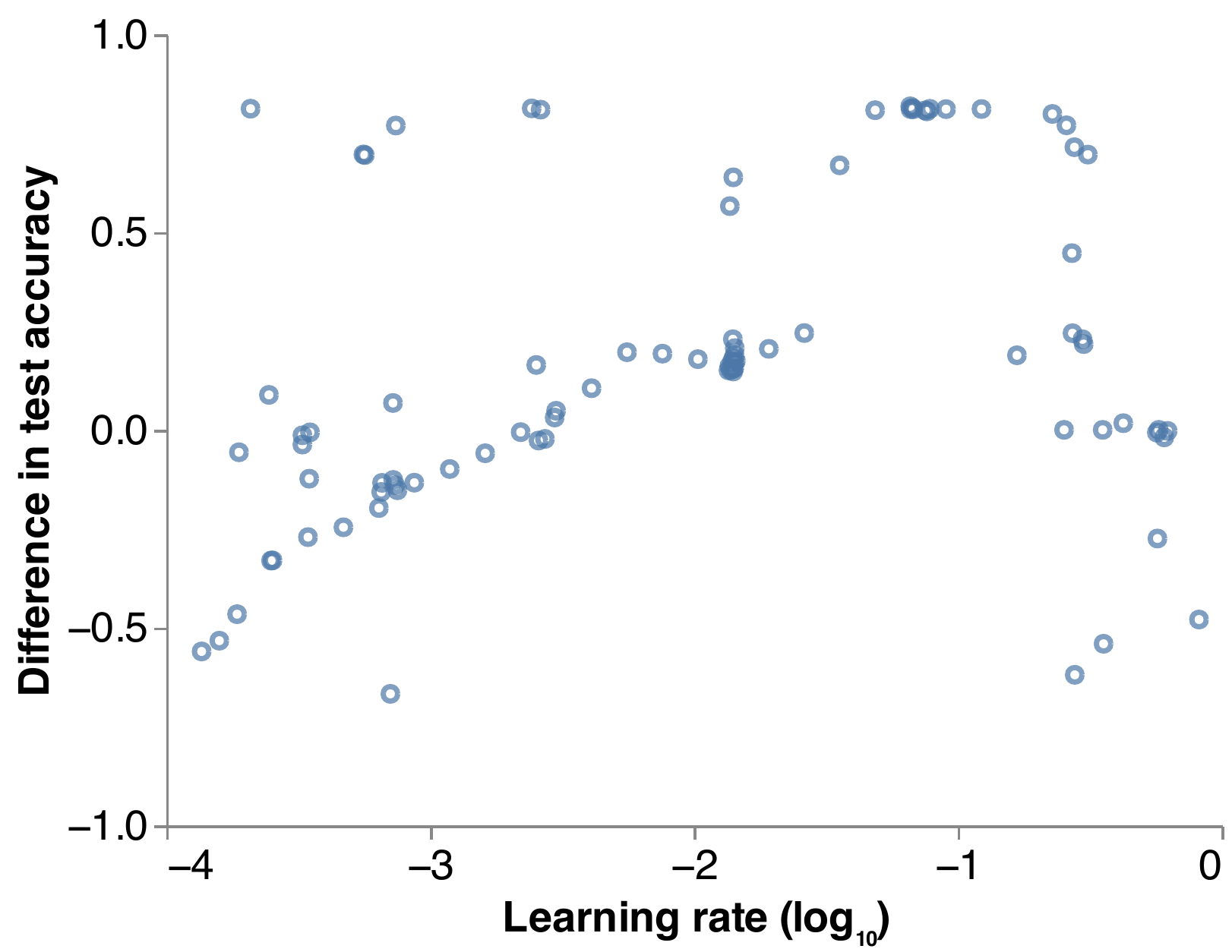}
    }\quad
    \subfloat[Epsilon]{
        \includegraphics[trim={17cm 0 0 0},clip,height=5cm]{adaptive-optimizers/adaptive-opt-scatters}
    } 
    \caption{Raw results for case study 1. Difference in test accuracy (SGD - Adam) by \textbf{(a)} learning rate and \textbf{(b)} \(\epsilon\). Orange line fit with linear regression for illustration purposes}\label{fig:adaptive-opt-scatters}
\end{figure}

\begin{figure}[h]
    \centering
    \subfloat[Learning rate]{
        \includegraphics[trim={0 1.5cm 30cm 0},clip,height=5cm]{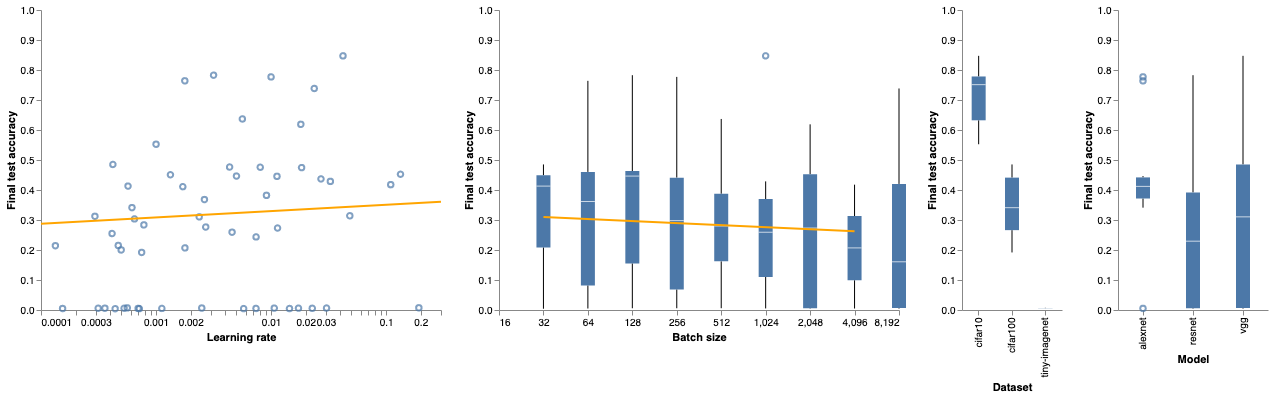}
    }\quad
    \subfloat[Batch size]{
        \includegraphics[trim={16.5cm 1.5cm 13.5cm 0},clip,height=5cm]{batch-size/batch-size-lr-scatters}
    } \\
    \subfloat[Dataset]{
        \includegraphics[trim={33cm 0.2cm 8cm 0},clip,height=6cm]{batch-size/batch-size-lr-scatters}
    }\qquad
    \subfloat[Model]{
        \includegraphics[trim={39cm 0.2cm 0 0},clip,height=6cm]{batch-size/batch-size-lr-scatters}
    }
    \caption{Raw results for case study 2. Test accuracy by \textbf{(a)} learning rate; \textbf{(b)} batch size; \textbf{(c)} dataset; \textbf{(d)} model. Orange lines fit with linear regression for illustration purposes}\label{fig:batch-size-scatters}
\end{figure}

\begin{figure}
    \centering
    \subfloat[AlexNet on Tiny ImageNet]{
        \includegraphics[trim={0 0 3.5cm 0},clip,width=0.30\textwidth]{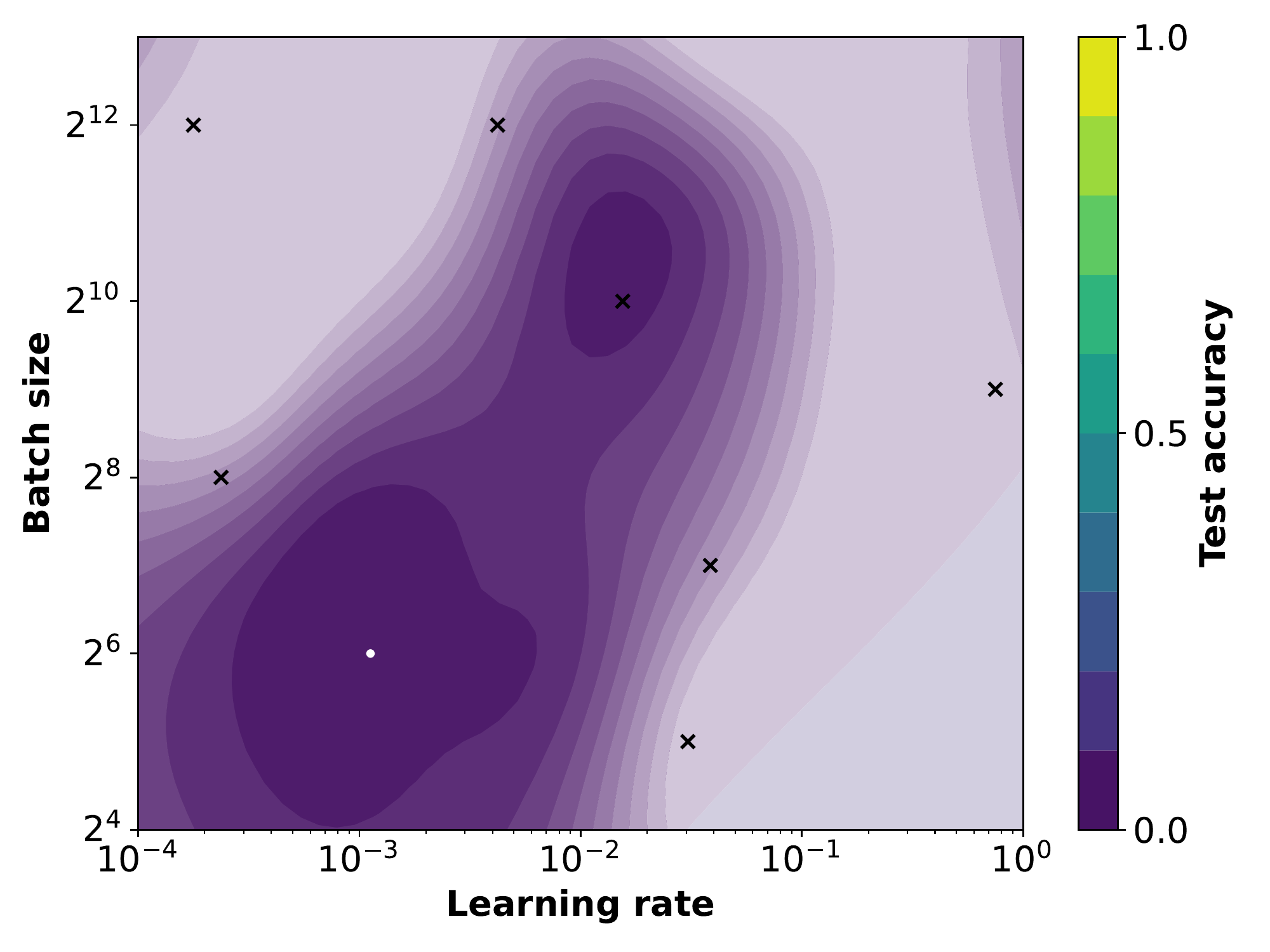}
    }
    \subfloat[ResNet on Tiny ImageNet]{
        \includegraphics[trim={0 0 3.5cm 0},clip,width=0.30\textwidth]{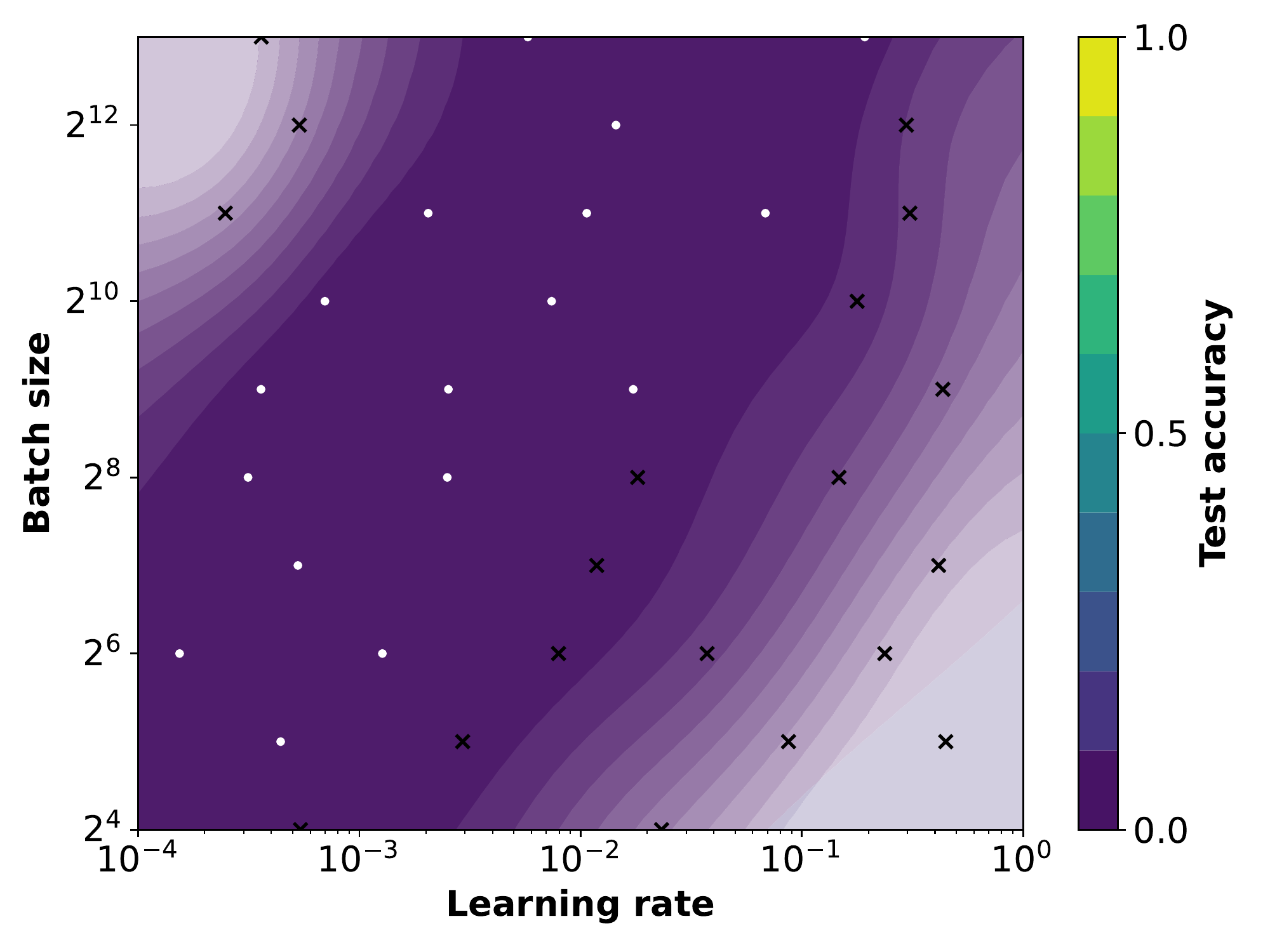}
    }
    \subfloat[VGG on Tiny ImageNet]{
        \includegraphics[trim={0 0 0 0},clip,width=0.3625\textwidth]{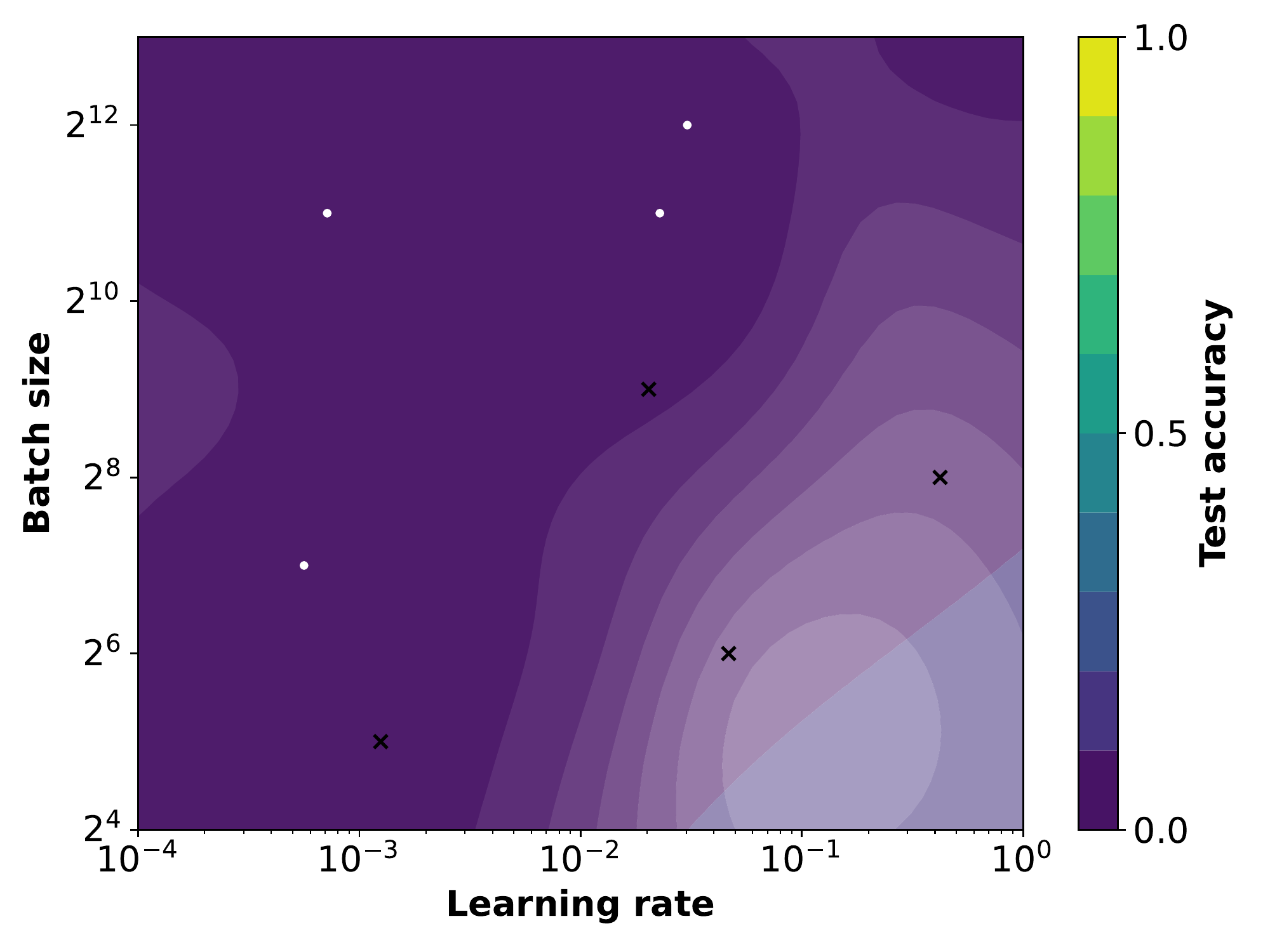}
    }
    \caption{Contour plot of GP-predicted mean test accuracy over the search space of learning rate, batch size, and model for Tiny ImageNet only. White points are trials with training accuracy \(\geq 0.99\); black crosses were excluded.}\label{fig:batch-size-contours-tiny-imagenet}
\end{figure}

\begin{figure}
    \centering
    \subfloat[AlexNet on CIFAR-10]{
        \includegraphics[trim={0 0 3.5cm 0},clip,width=0.30\textwidth]{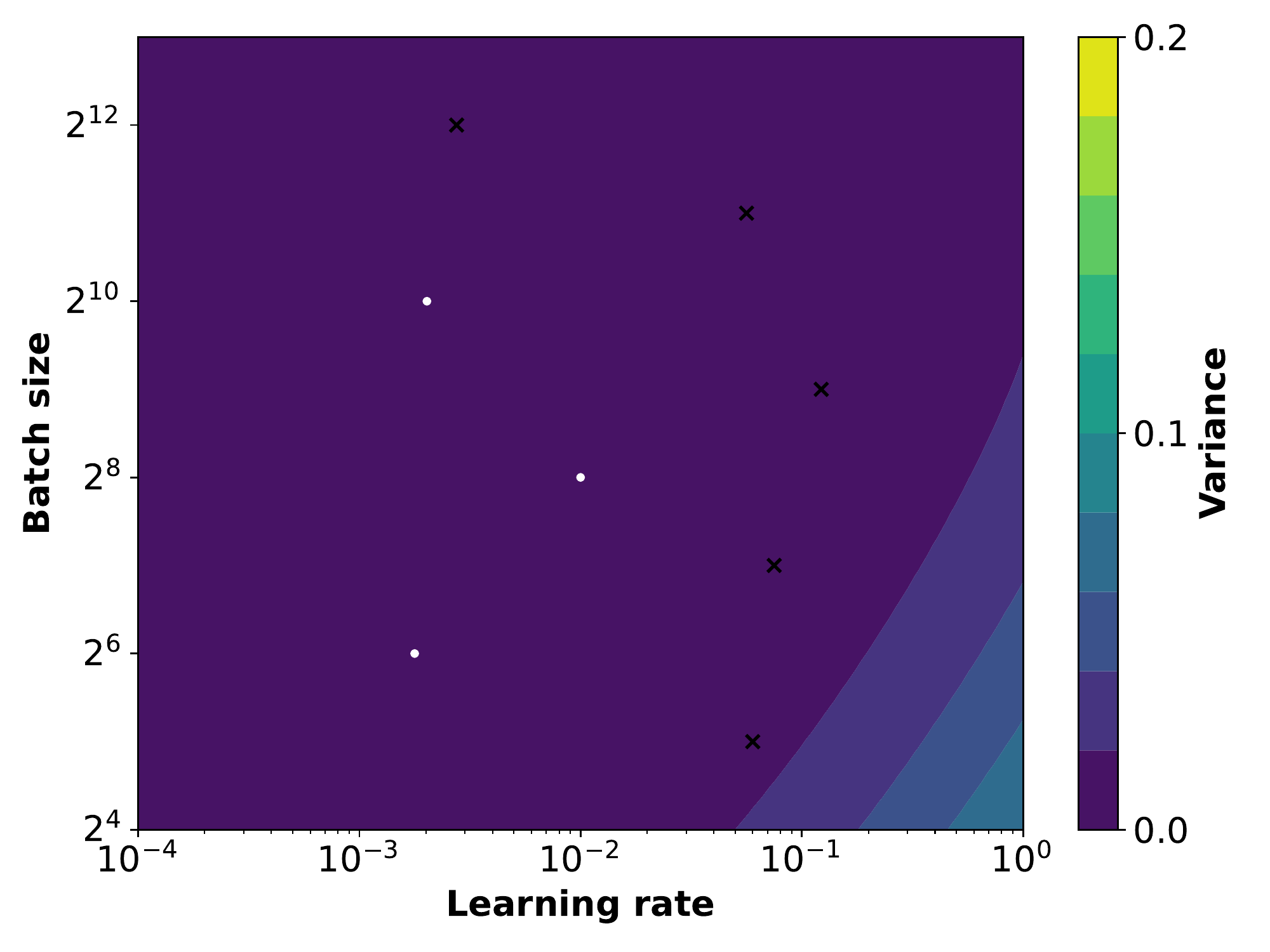}
    }
    \subfloat[ResNet on CIFAR-10]{
        \includegraphics[trim={0 0 3.5cm 0},clip,width=0.30\textwidth]{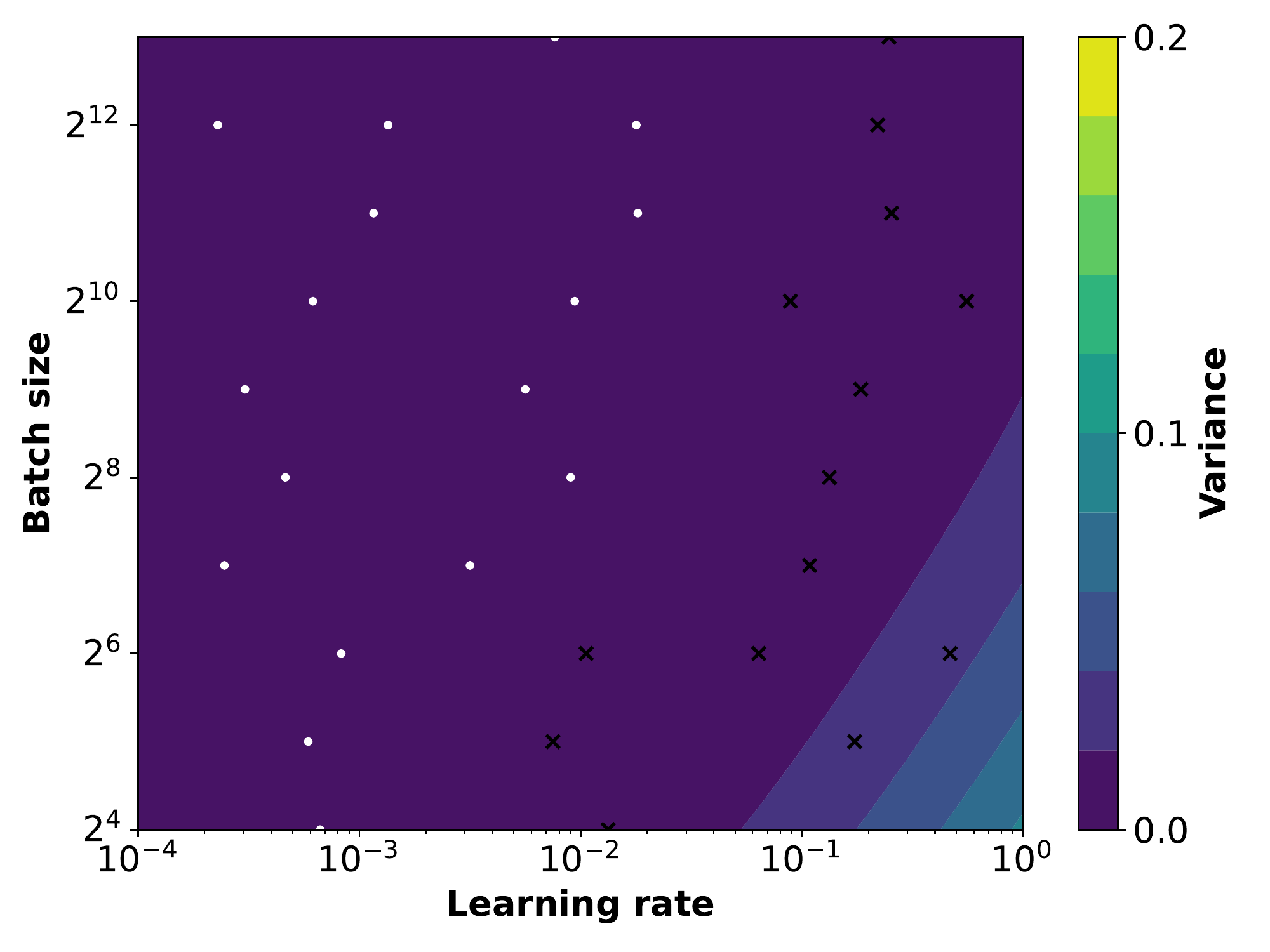}
    }
    \subfloat[VGG on CIFAR-10]{
        \includegraphics[trim={0 0 0 0},clip,width=0.3625\textwidth]{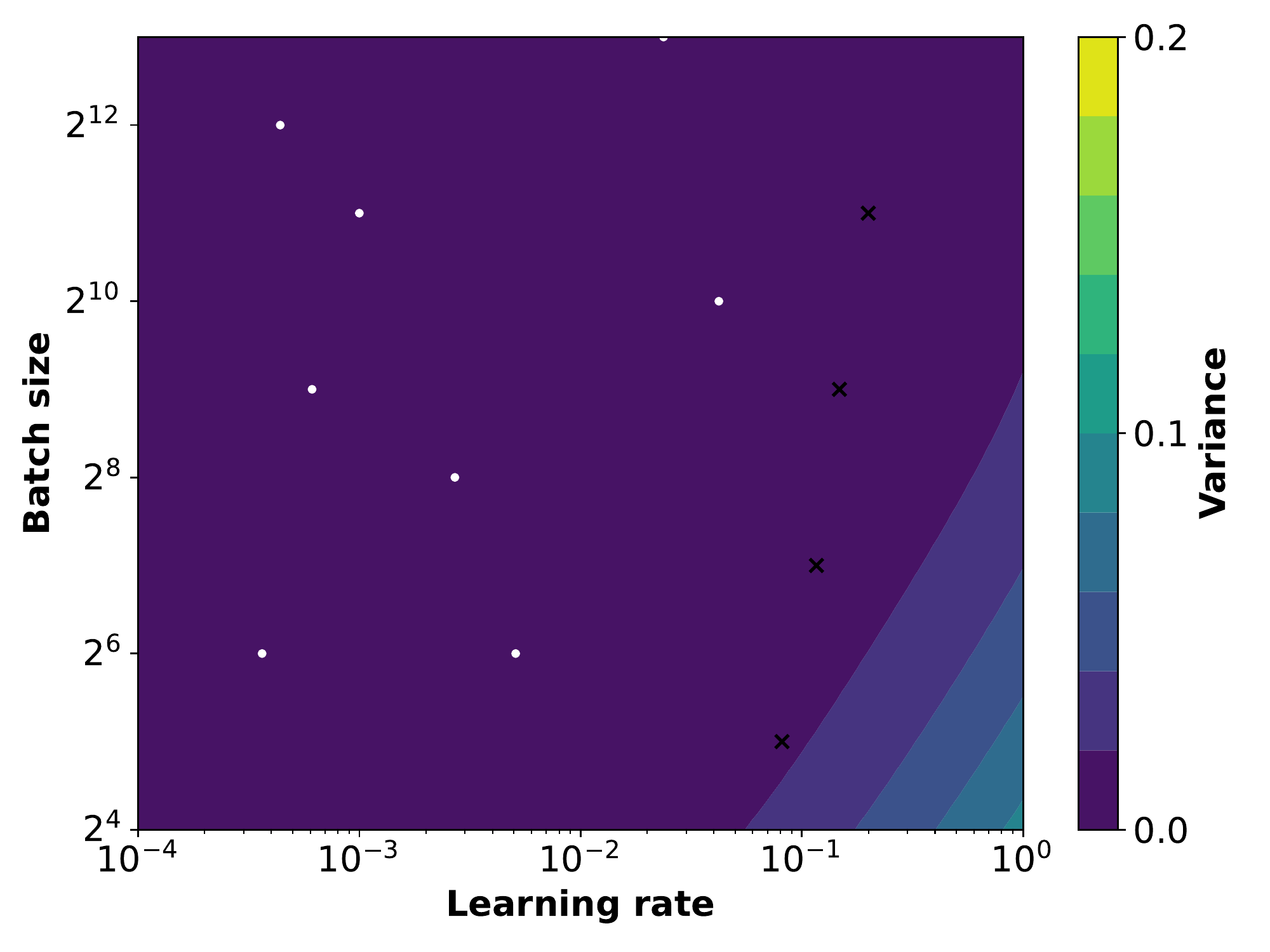}
    } \\ 
    \subfloat[AlexNet on CIFAR-100]{
        \includegraphics[trim={0 0 3.5cm 0},clip,width=0.30\textwidth]{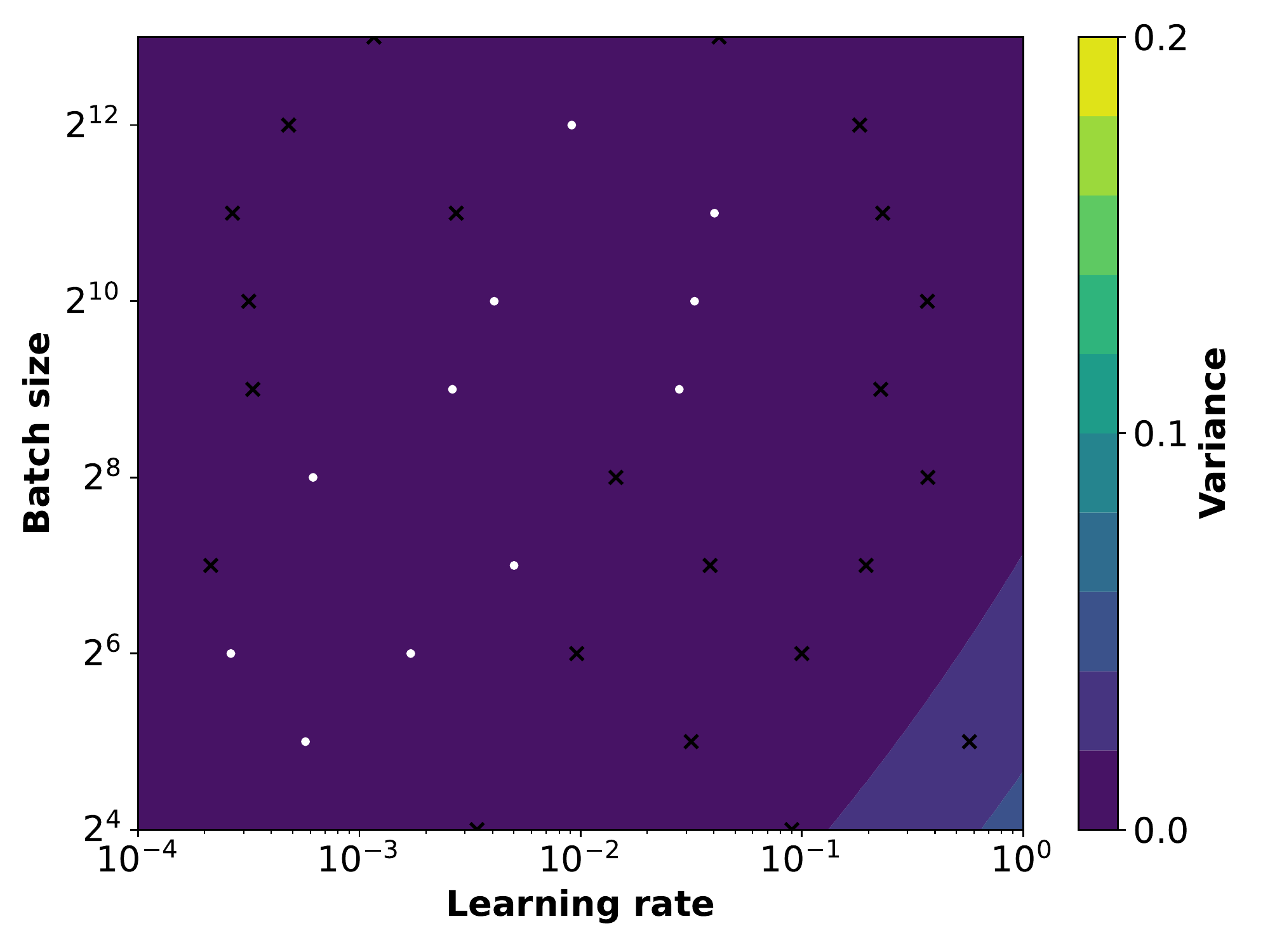}
    }
    \subfloat[ResNet on CIFAR-100]{
        \includegraphics[trim={0 0 3.5cm 0},clip,width=0.30\textwidth]{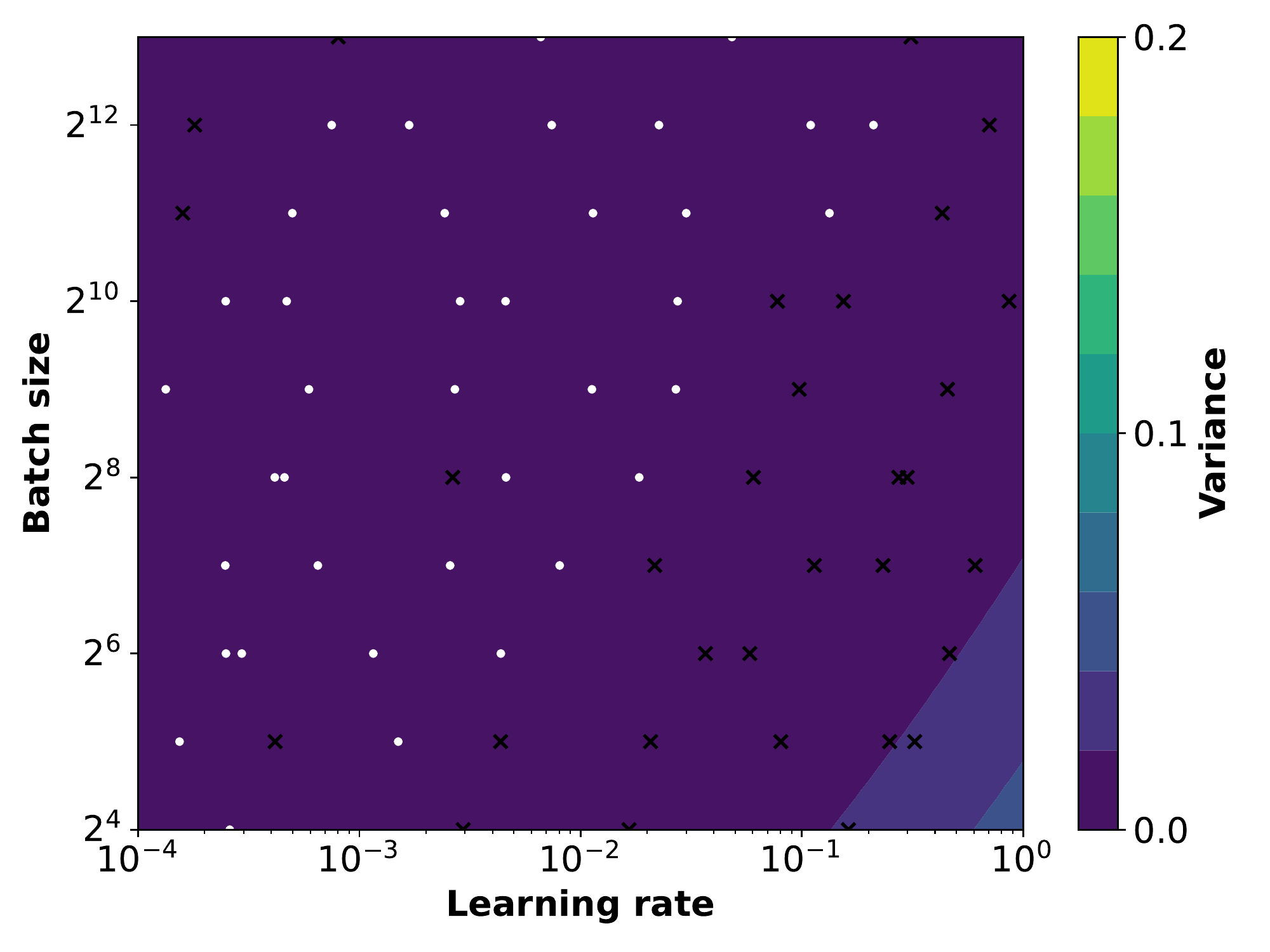}
    }
    \subfloat[VGG on CIFAR-100]{
        \includegraphics[trim={0 0 0 0},clip,width=0.3625\textwidth]{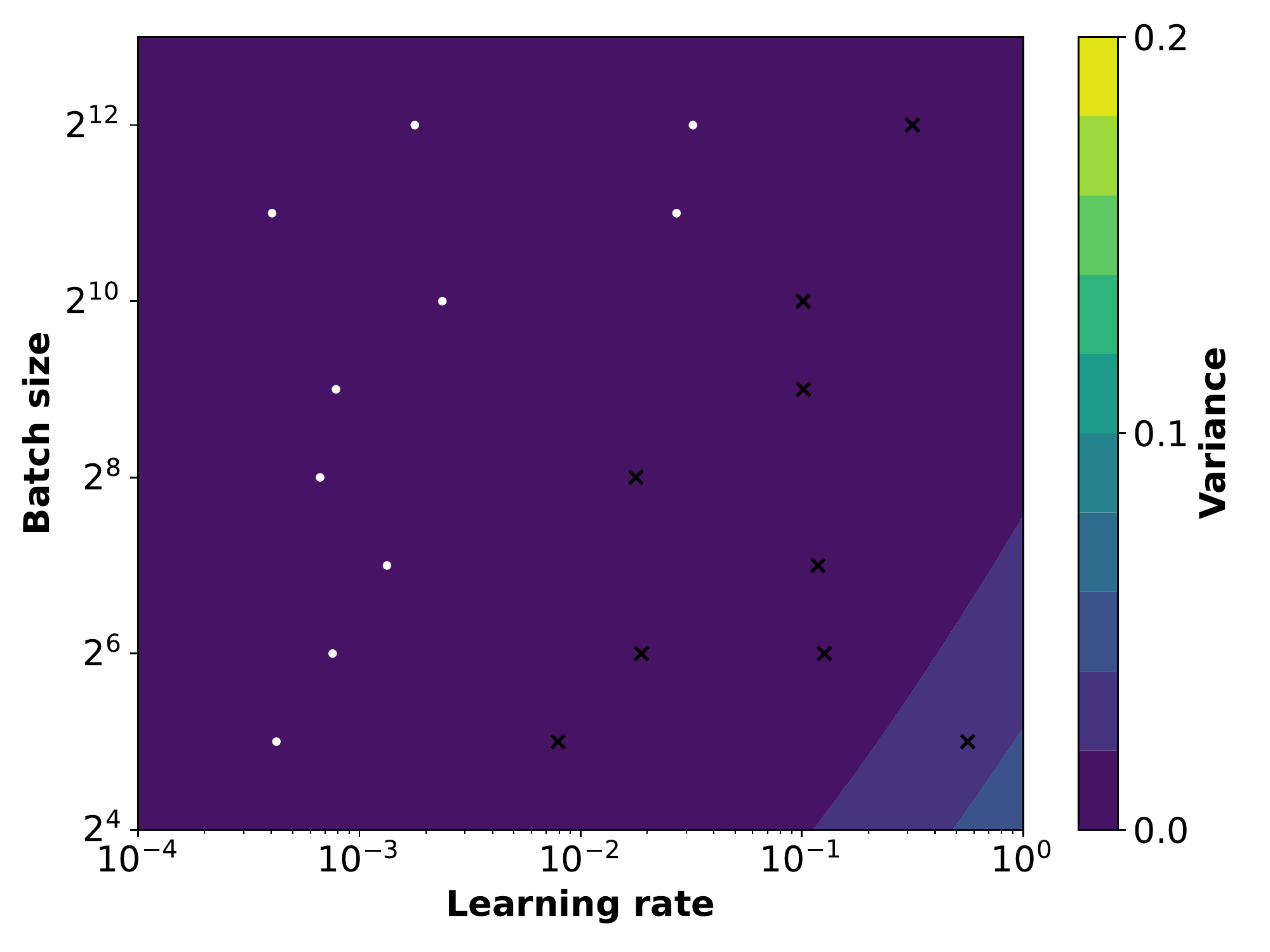}
    } \\
    \subfloat[AlexNet on Tiny ImageNet]{
        \includegraphics[trim={0 0 3.5cm 0},clip,width=0.30\textwidth]{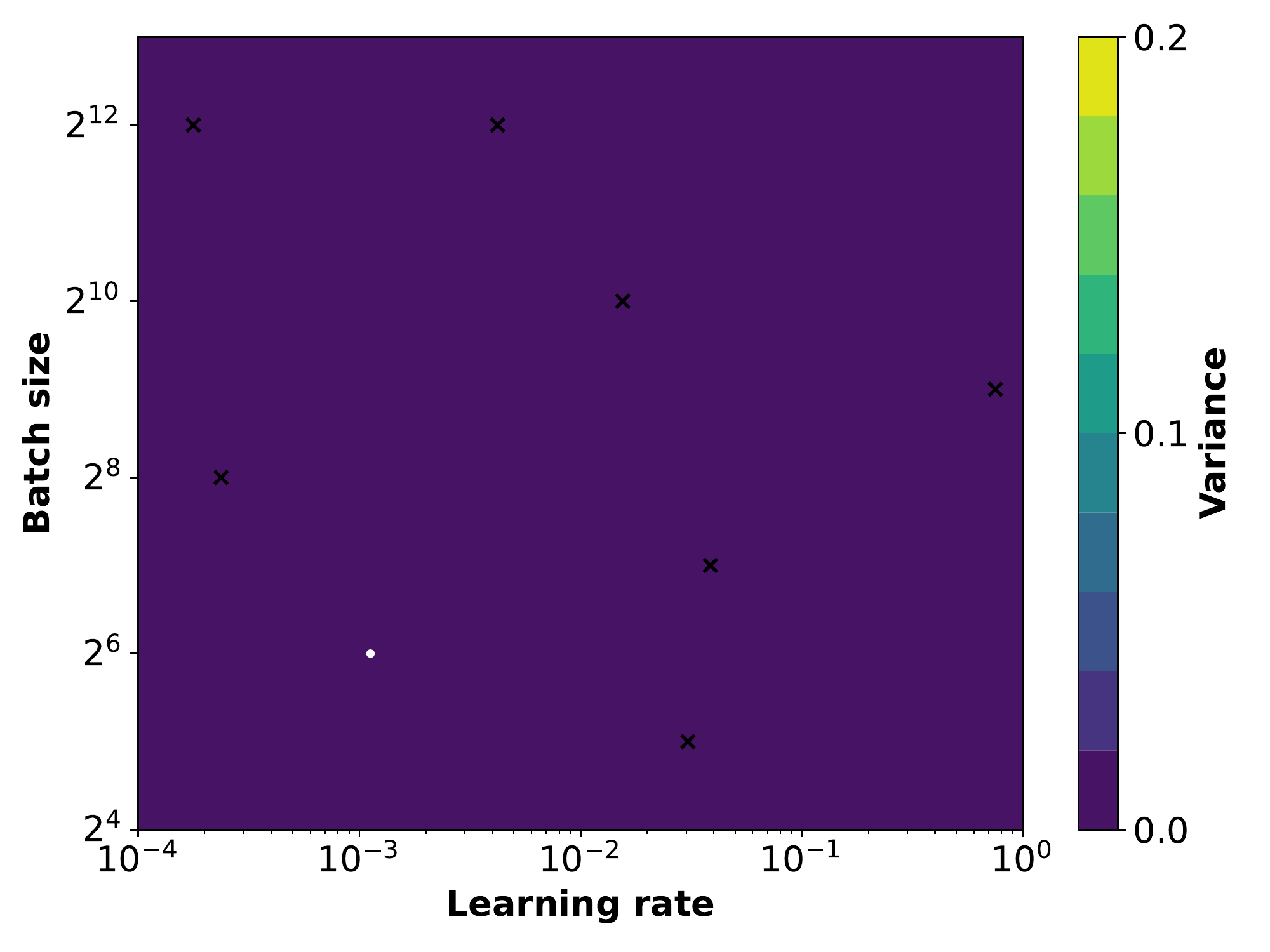}
    }
    \subfloat[ResNet on Tiny ImageNet]{
        \includegraphics[trim={0 0 3.5cm 0},clip,width=0.30\textwidth]{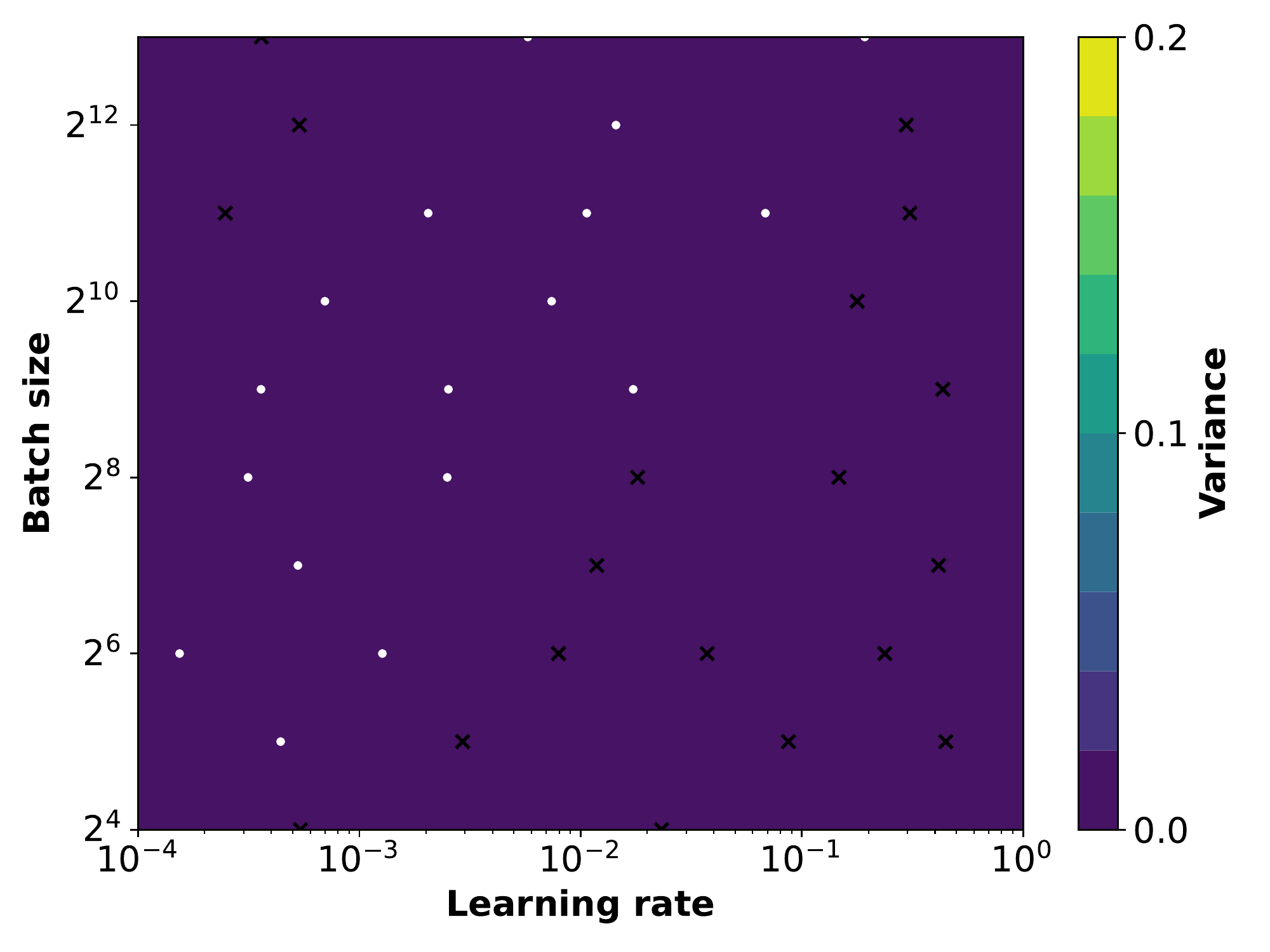}
    }
    \subfloat[VGG on Tiny ImageNet]{
        \includegraphics[trim={0 0 0 0},clip,width=0.3625\textwidth]{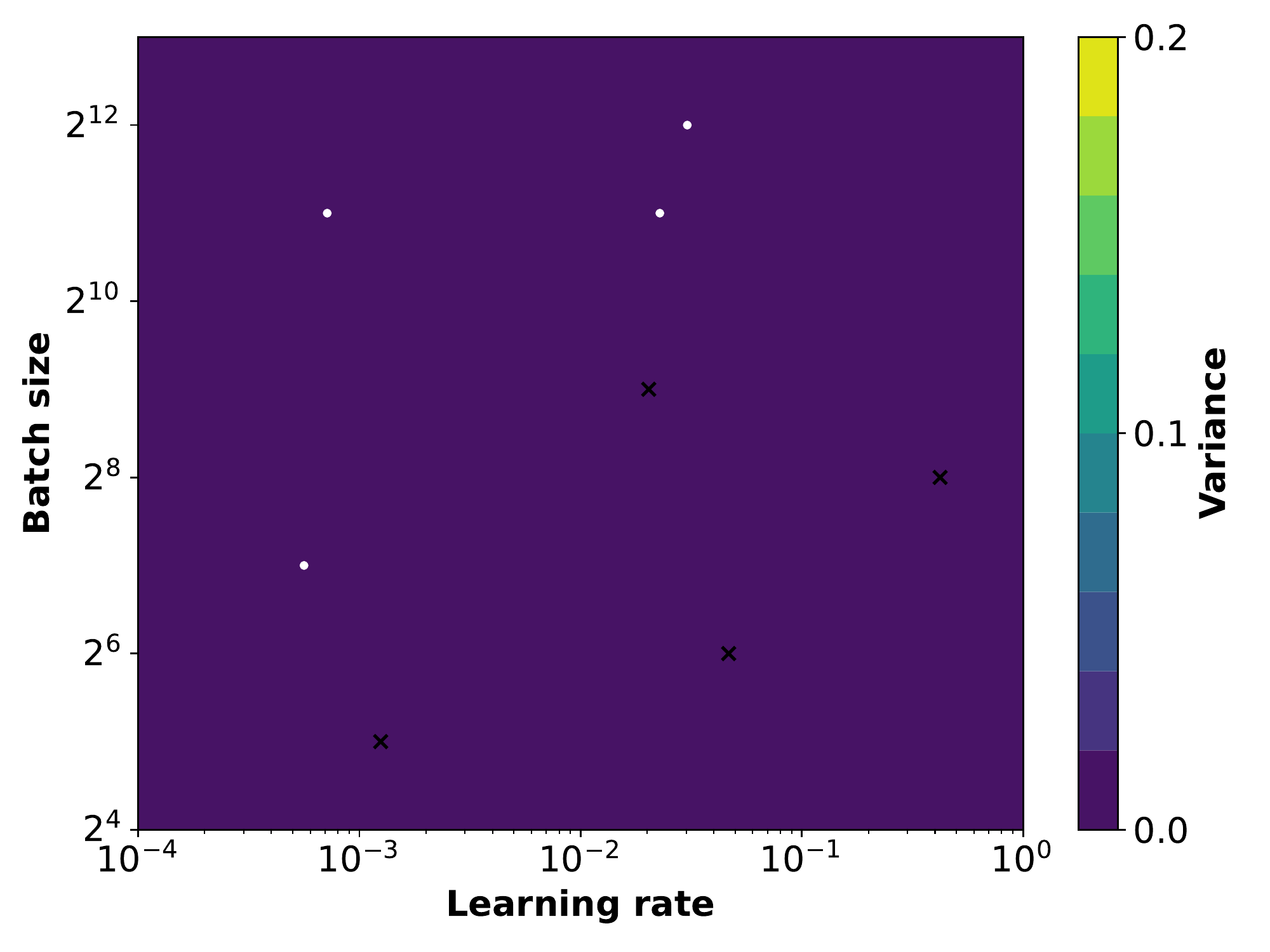}
    }

    \caption{Contour plot of test accuracy variance over the search space of learning rate, batch size, dataset and model. White points are trials with training accuracy \(\geq 0.99\); black crosses were excluded.}\label{fig:batch-size-contours-var}
\end{figure}

\begin{figure}
    \centering
    \subfloat[Mat\'{e}rn 5/2 with ARD]{
        \includegraphics[trim={0 0 0 0},clip,width=0.45\textwidth]{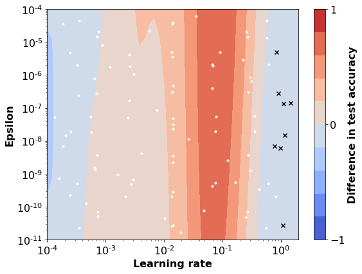}
    }
    \subfloat[RBF with ARD]{
        \includegraphics[trim={0 0 0 0},clip,width=0.45\textwidth]{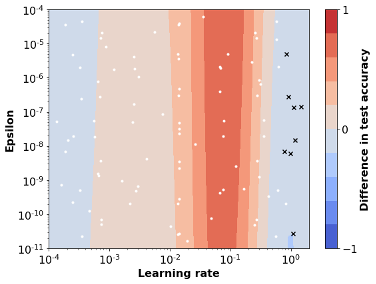}
    } \\ 
    \subfloat[Mat\'{e}rn 5/2 without ARD]{
        \includegraphics[trim={0 0 0 0},clip,width=0.45\textwidth]{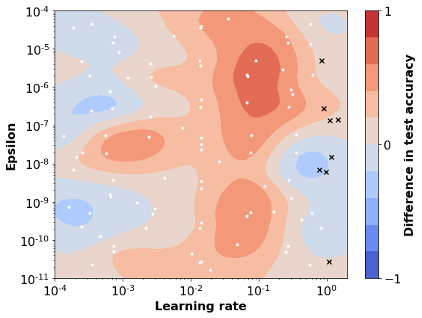}
    }
    \subfloat[RBF without ARD]{
        \includegraphics[trim={0 0 0 0},clip,width=0.45\textwidth]{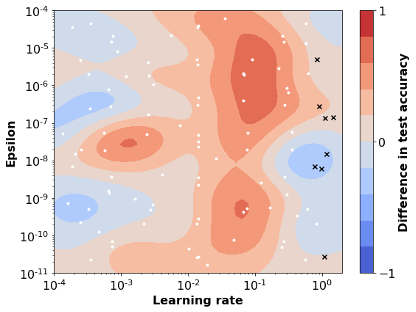}
    }
    \caption{Effect of GP kernel on predicted mean difference in test accuracy (SGD - Adam) over the search space of learning rate and \(\epsilon\). Different kernel choices may impact the contour visualization, but do not change the primary conclusion regarding learning rate sensitivity and \(\epsilon\) insensitivity. A complementary analysis over different initial kernel lengthscales and variances yielded Bayes factors consistent with our conclusions across all trials, showing that the conclusions we draw are insensitive to GP hyperparameters.}\label{fig:adaptive-opt-kernel-comparison}
\end{figure}

\begin{figure}[t]
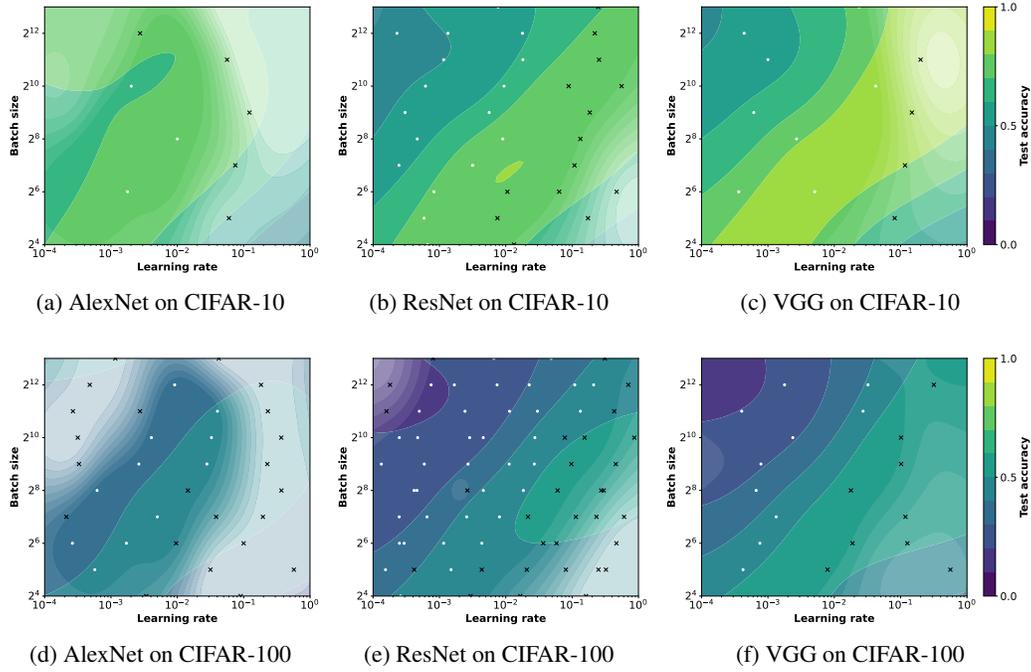

    \centering
    \subfloat[AlexNet on CIFAR-10]{
        \includegraphics[trim={0 0 3.5cm 0},clip,width=0.30\textwidth]{batch-size/batch-size-lr-final-test-acc-mean-cifar10-alexnet}
    }
    \subfloat[ResNet on CIFAR-10]{
        \includegraphics[trim={0 0 3.5cm 0},clip,width=0.30\textwidth]{batch-size/batch-size-lr-final-test-acc-mean-cifar10-resnet}
    }
    \subfloat[VGG on CIFAR-10]{
        \includegraphics[trim={0 0 0 0},clip,width=0.3625\textwidth]{batch-size/batch-size-lr-final-test-acc-mean-cifar10-vgg}
    } \\ 
    \subfloat[AlexNet on CIFAR-100]{
        \includegraphics[trim={0 0 3.5cm 0},clip,width=0.30\textwidth]{batch-size/batch-size-lr-final-test-acc-mean-cifar100-alexnet}
    }
    \subfloat[ResNet on CIFAR-100]{
        \includegraphics[trim={0 0 3.5cm 0},clip,width=0.30\textwidth]{batch-size/batch-size-lr-final-test-acc-mean-cifar100-resnet}
    }
    \subfloat[VGG on CIFAR-100]{
        \includegraphics[trim={0 0 0 0},clip,width=0.3625\textwidth]{batch-size/batch-size-lr-final-test-acc-mean-cifar100-vgg}
    }
    \caption{Effect of GP kernel on predicted test accuracy over the search space of batch size, learning rate, model and dataset: \textbf{Mat\'{e}rn 5/2 kernel with ARD}. Contours are broadly consistent across kernel choices as shown in \cref{fig:batch-size-kernel-comparison-matern-no-ard,fig:batch-size-kernel-comparison-rbf-ard,fig:batch-size-kernel-comparison-rbf-no-ard}.  This figure is repeated in the main text, but included here for easy comparison.}\label{fig:batch-size-kernel-comparison-matern-ard}
\end{figure}

\begin{figure}[t]
    \centering
    \subfloat[AlexNet on CIFAR-10]{
        \includegraphics[trim={0 0 3.5cm 0},clip,width=0.30\textwidth]{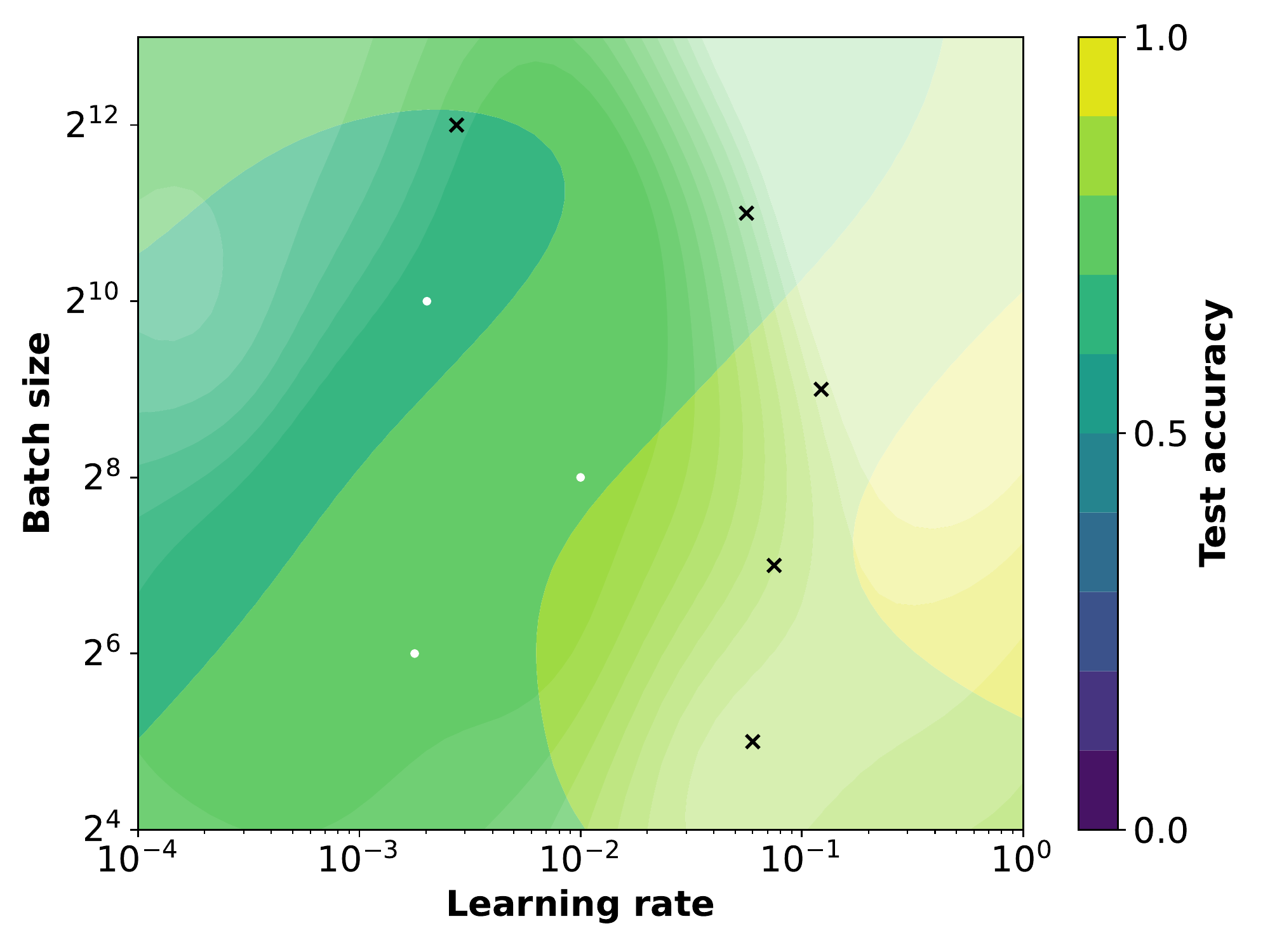}
    }
    \subfloat[ResNet on CIFAR-10]{
        \includegraphics[trim={0 0 3.5cm 0},clip,width=0.30\textwidth]{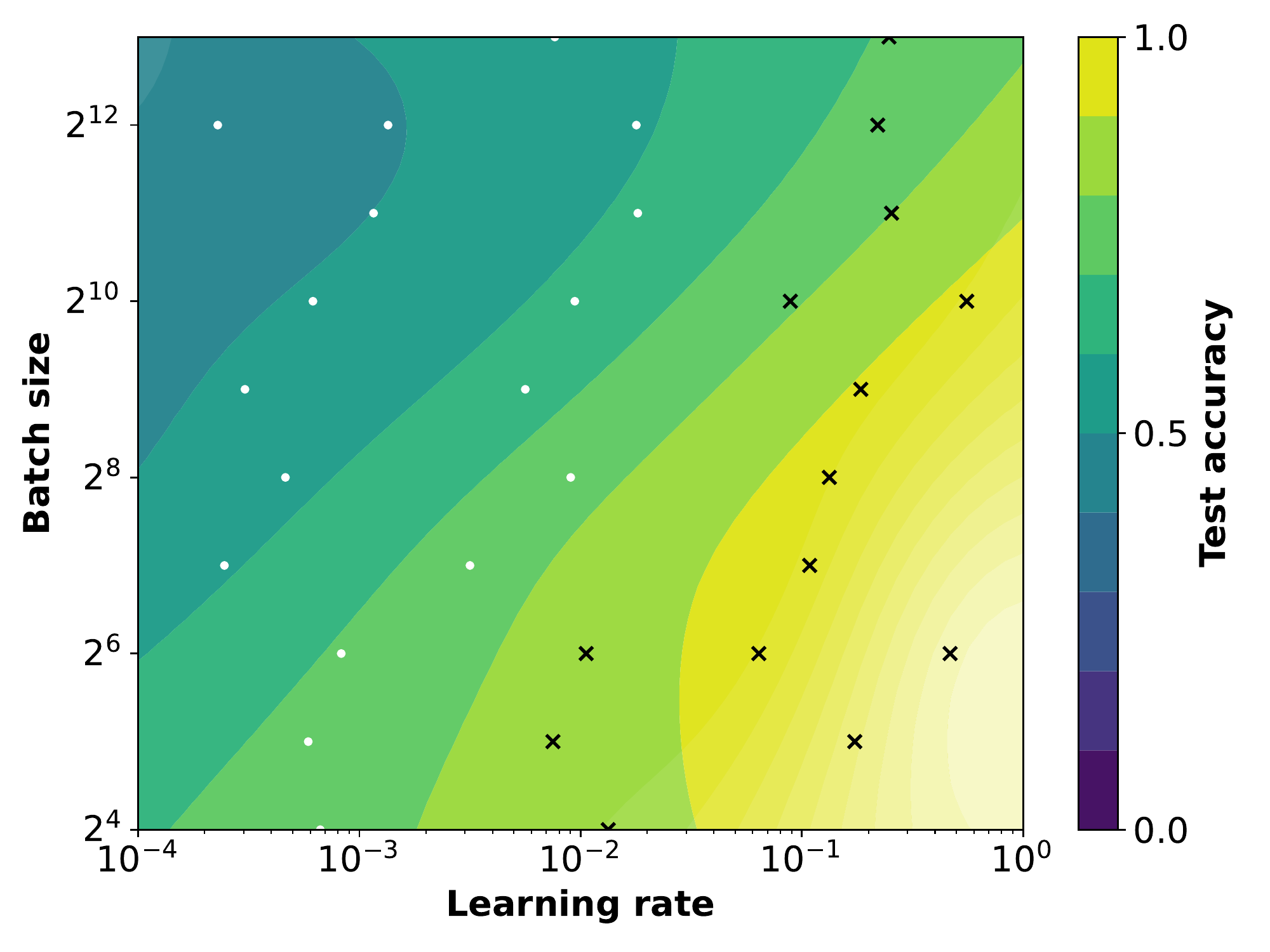}
    }
    \subfloat[VGG on CIFAR-10]{
        \includegraphics[trim={0 0 0 0},clip,width=0.3625\textwidth]{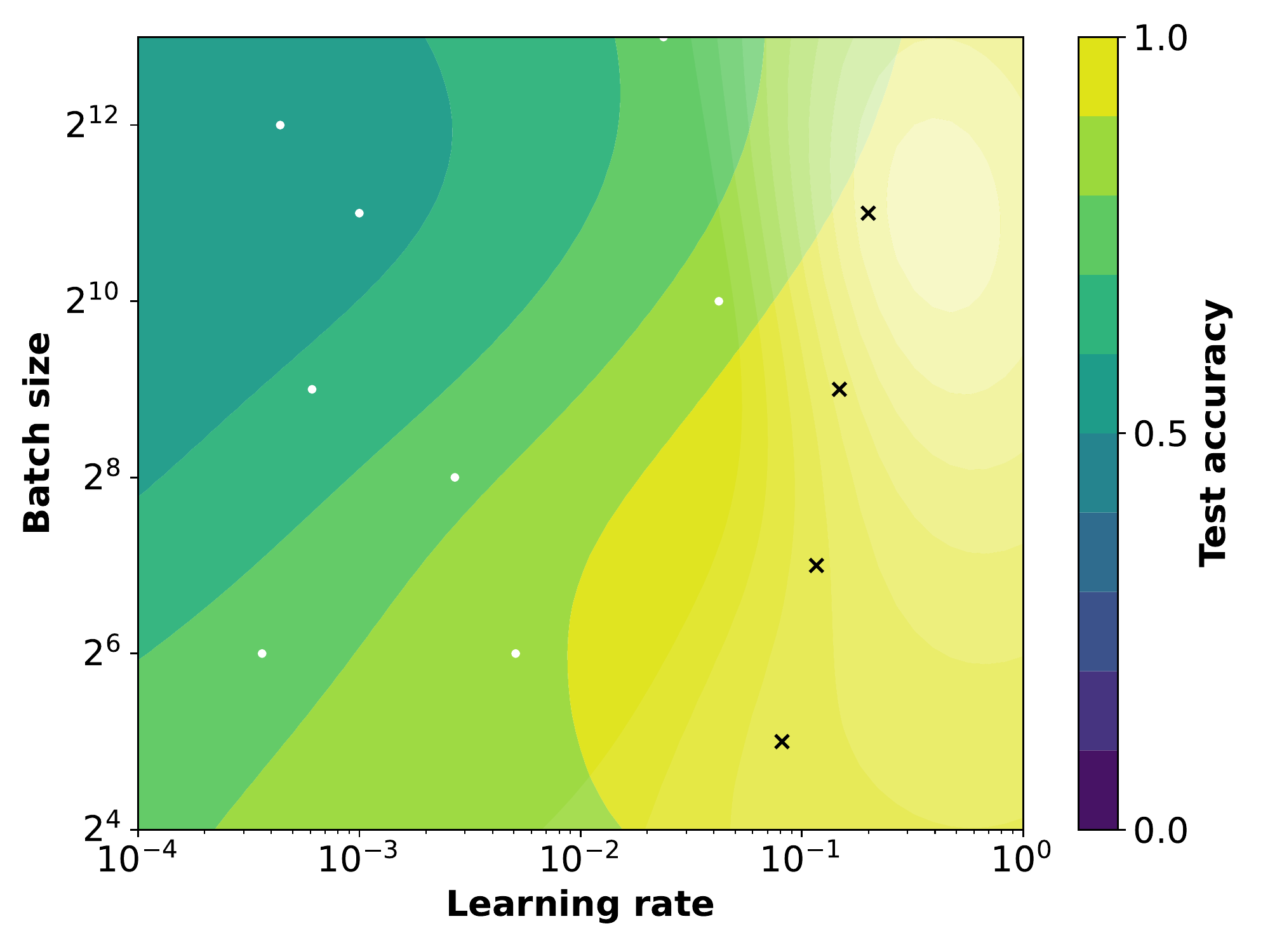}
    } \\ 
    \subfloat[AlexNet on CIFAR-100]{
        \includegraphics[trim={0 0 3.5cm 0},clip,width=0.30\textwidth]{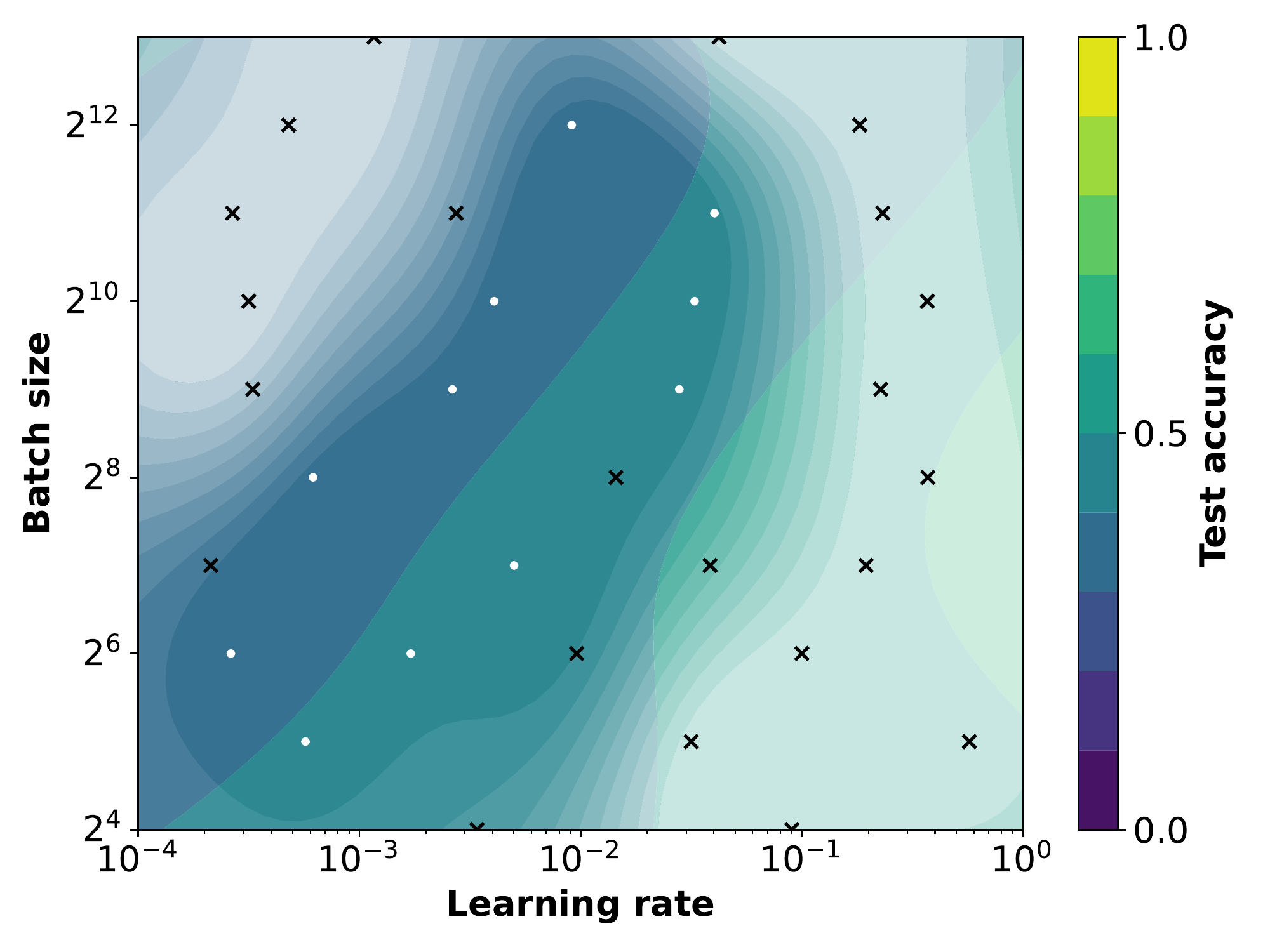}
    }
    \subfloat[ResNet on CIFAR-100]{
        \includegraphics[trim={0 0 3.5cm 0},clip,width=0.30\textwidth]{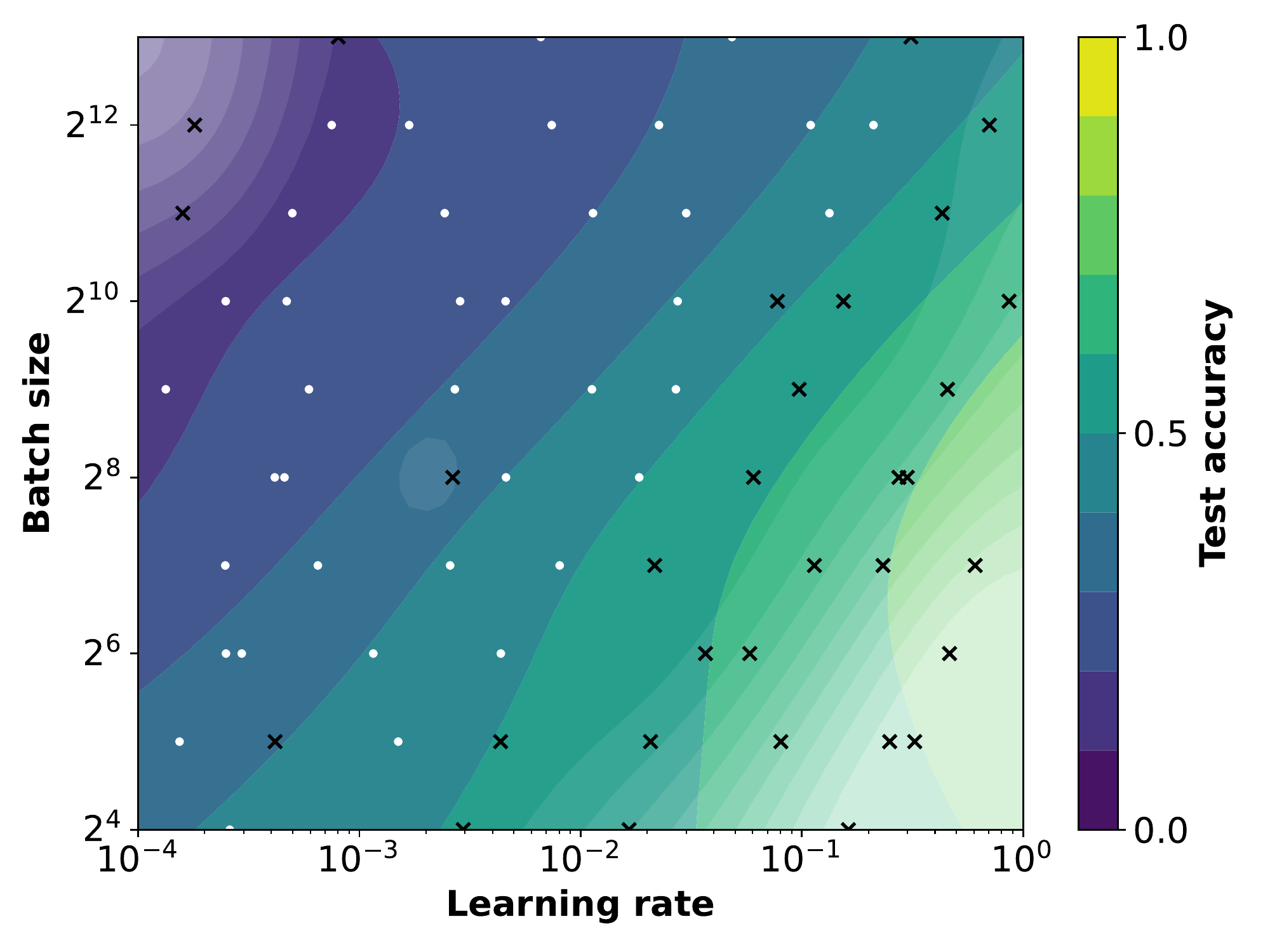}
    }
    \subfloat[VGG on CIFAR-100]{
        \includegraphics[trim={0 0 0 0},clip,width=0.3625\textwidth]{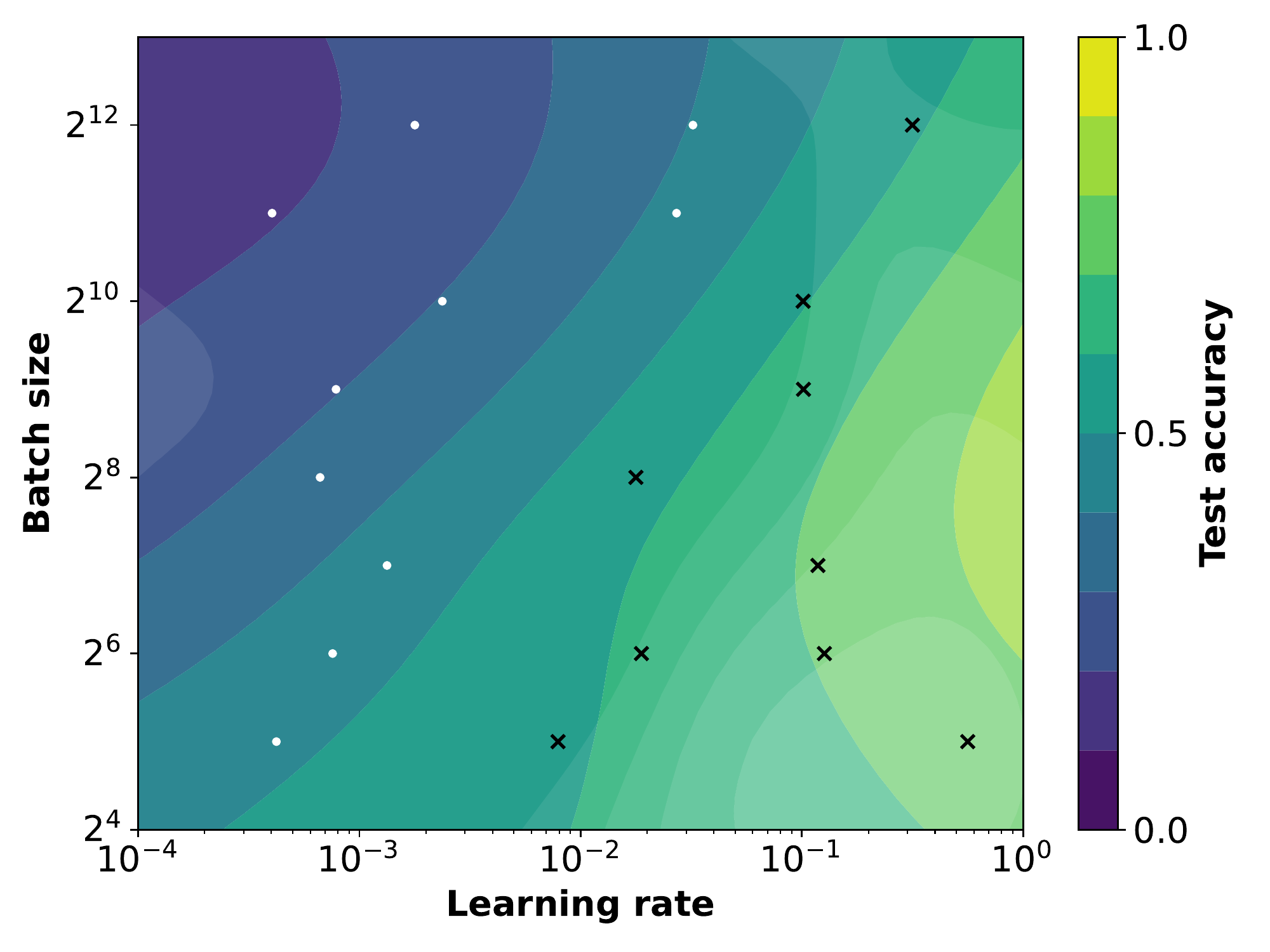}
    }
    \caption{Effect of GP kernel on predicted test accuracy over the search space of batch size, learning rate, model and dataset: \textbf{Mat\'{e}rn 5/2 kernel without ARD}.}\label{fig:batch-size-kernel-comparison-matern-no-ard}
\end{figure}

\begin{figure}[t]
    \centering
    \subfloat[AlexNet on CIFAR-10]{
        \includegraphics[trim={0 0 3.5cm 0},clip,width=0.30\textwidth]{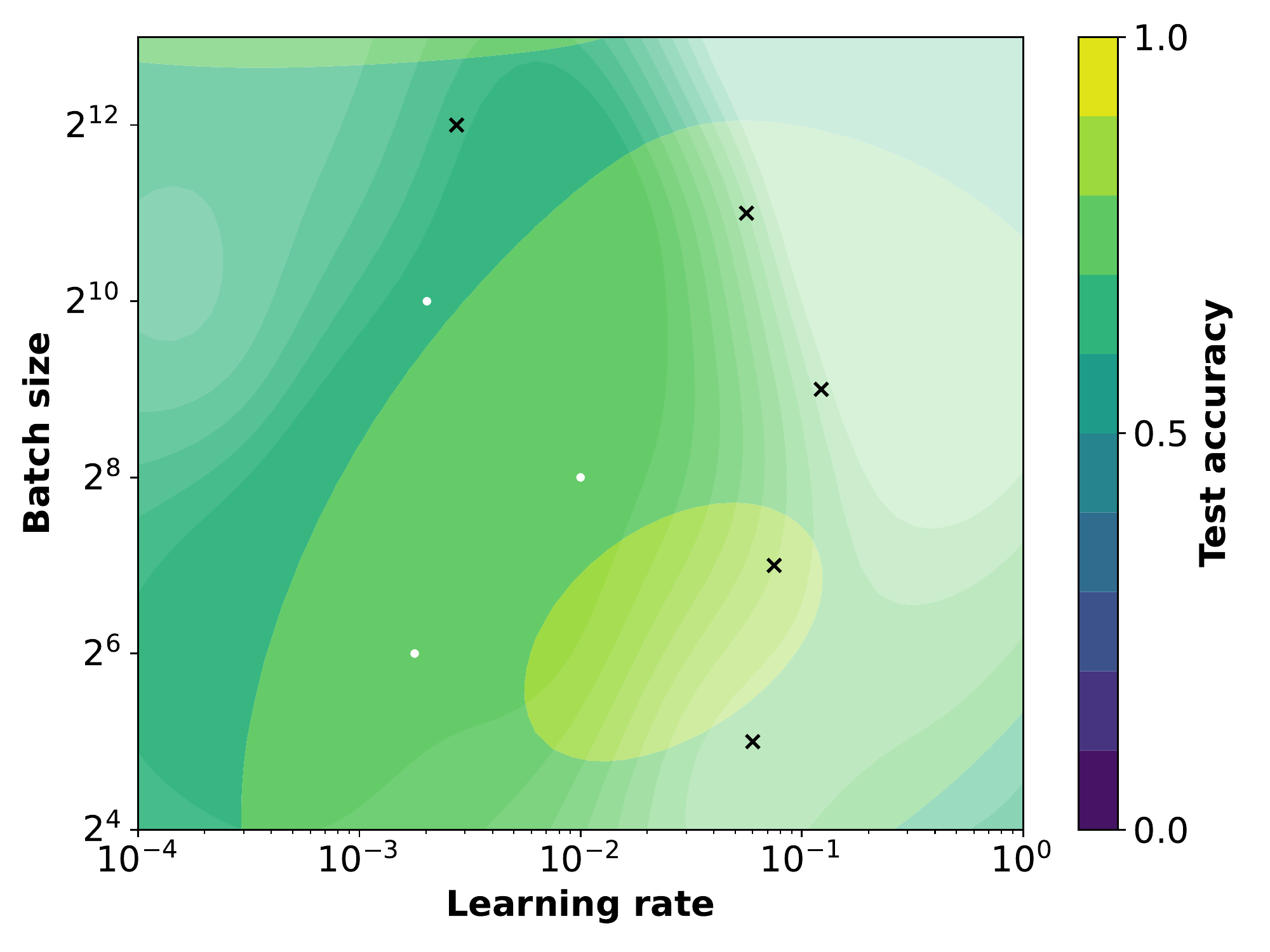}
    }
    \subfloat[ResNet on CIFAR-10]{
        \includegraphics[trim={0 0 3.5cm 0},clip,width=0.30\textwidth]{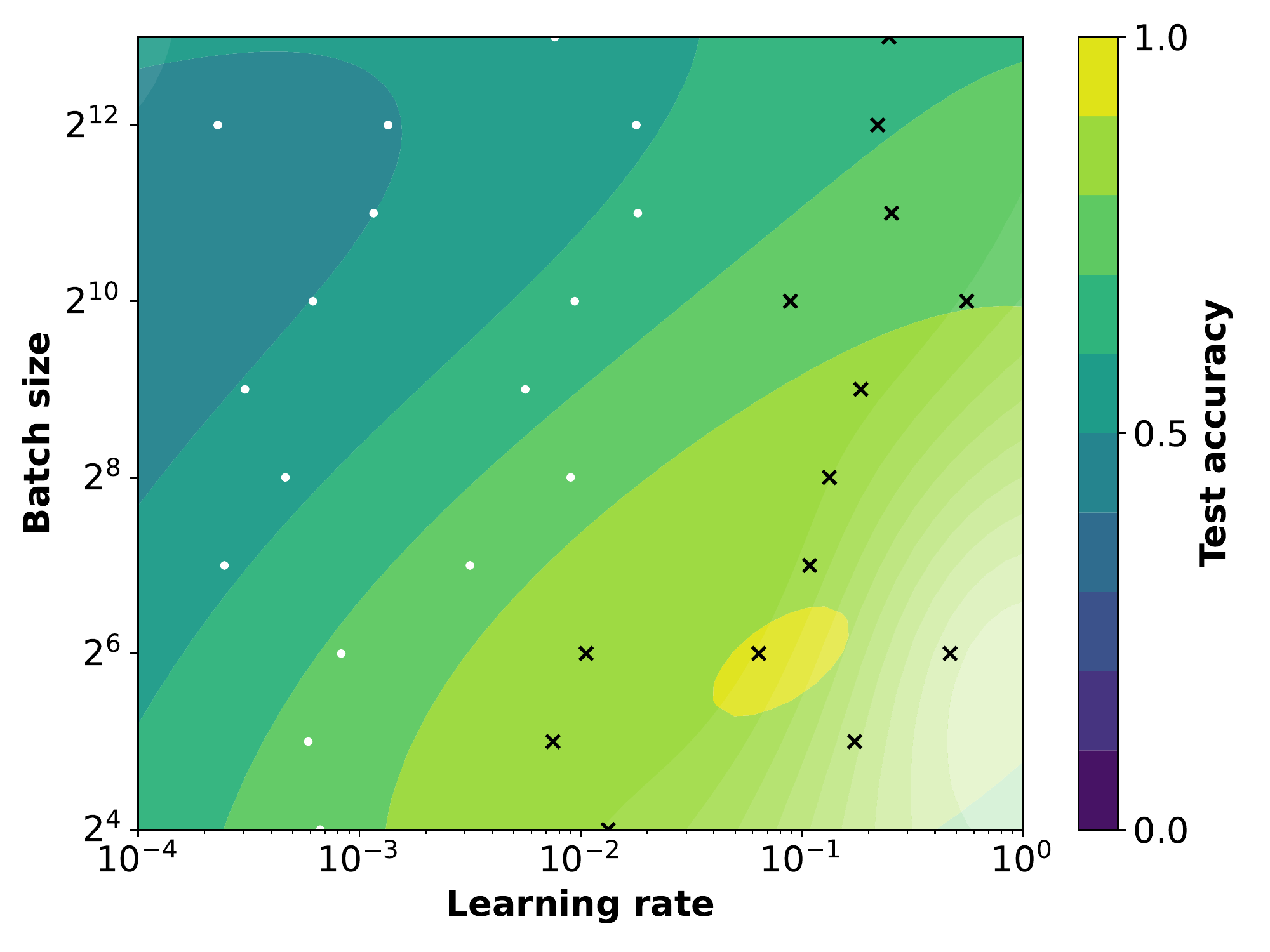}
    }
    \subfloat[VGG on CIFAR-10]{
        \includegraphics[trim={0 0 0 0},clip,width=0.3625\textwidth]{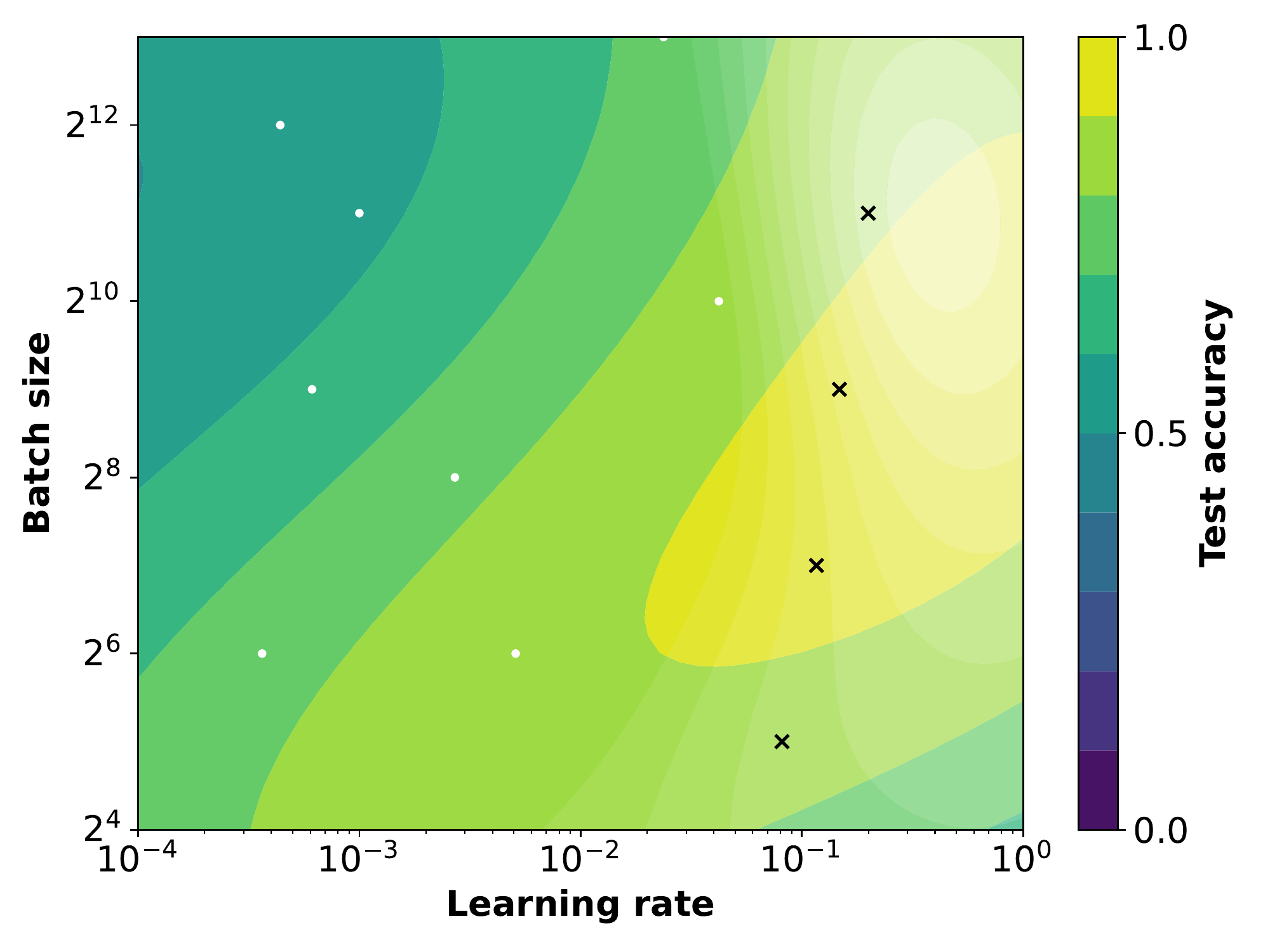}
    } \\ 
    \subfloat[AlexNet on CIFAR-100]{
        \includegraphics[trim={0 0 3.5cm 0},clip,width=0.30\textwidth]{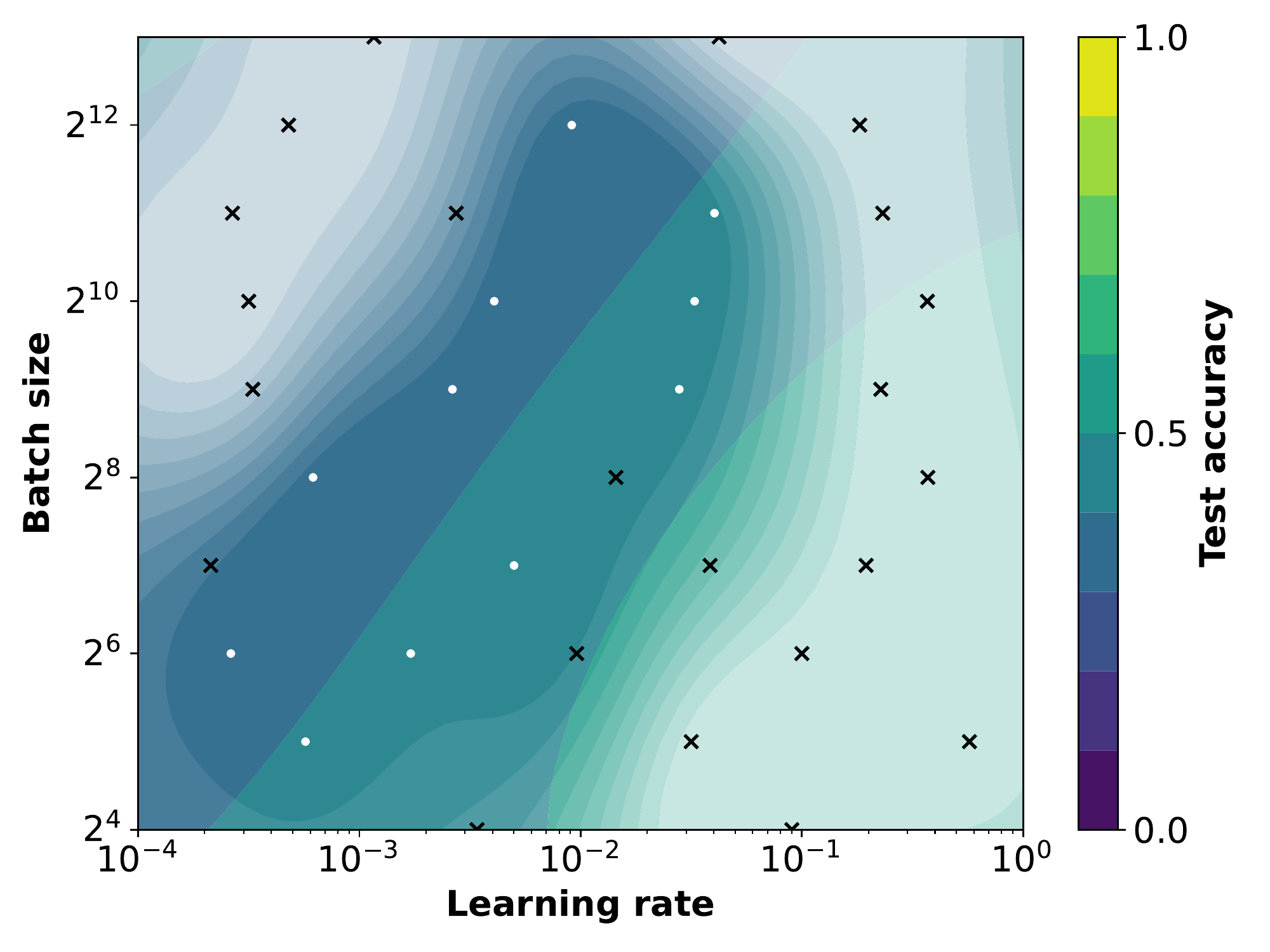}
    }
    \subfloat[ResNet on CIFAR-100]{
        \includegraphics[trim={0 0 3.5cm 0},clip,width=0.30\textwidth]{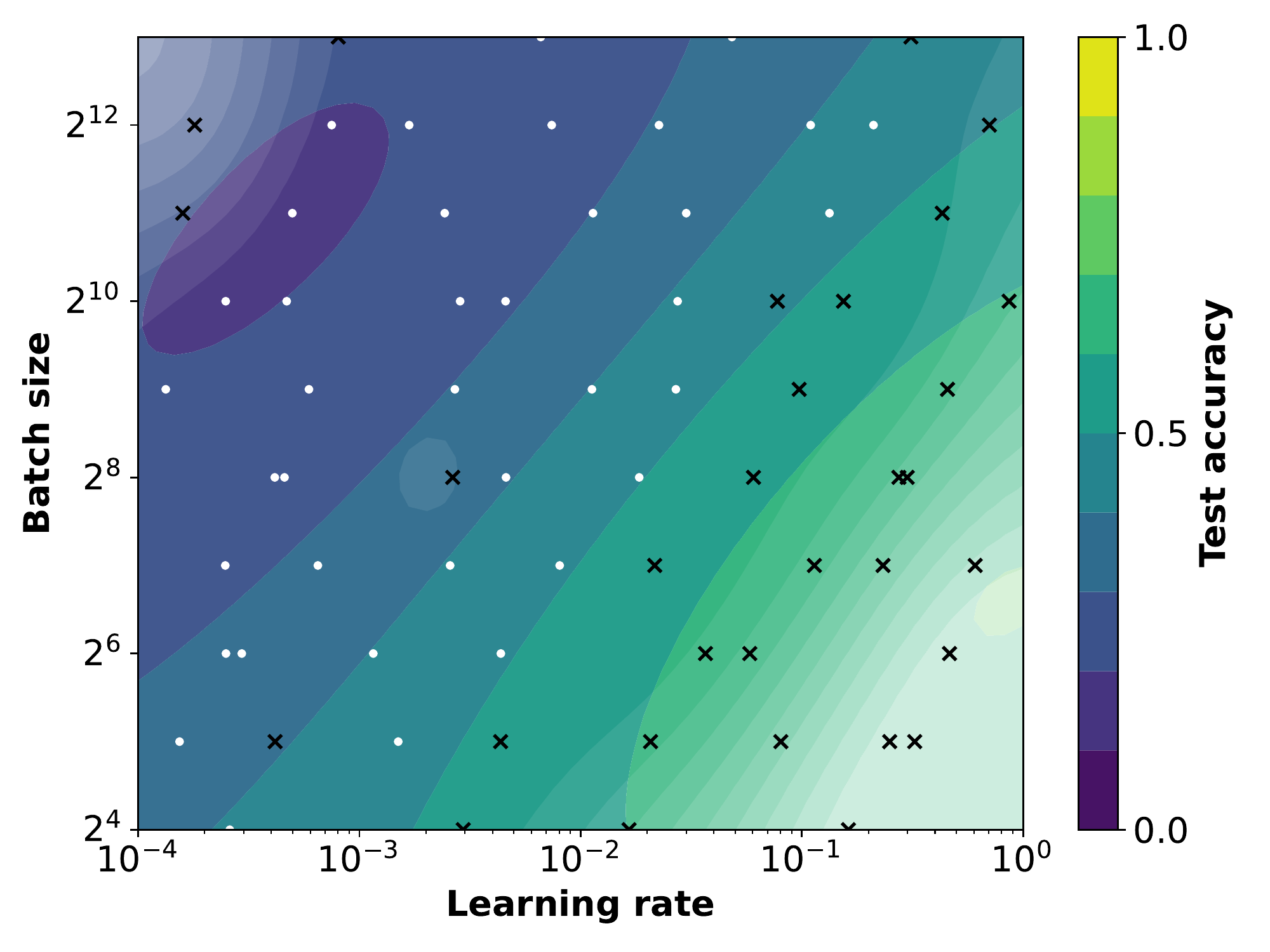}
    }
    \subfloat[VGG on CIFAR-100]{
        \includegraphics[trim={0 0 0 0},clip,width=0.3625\textwidth]{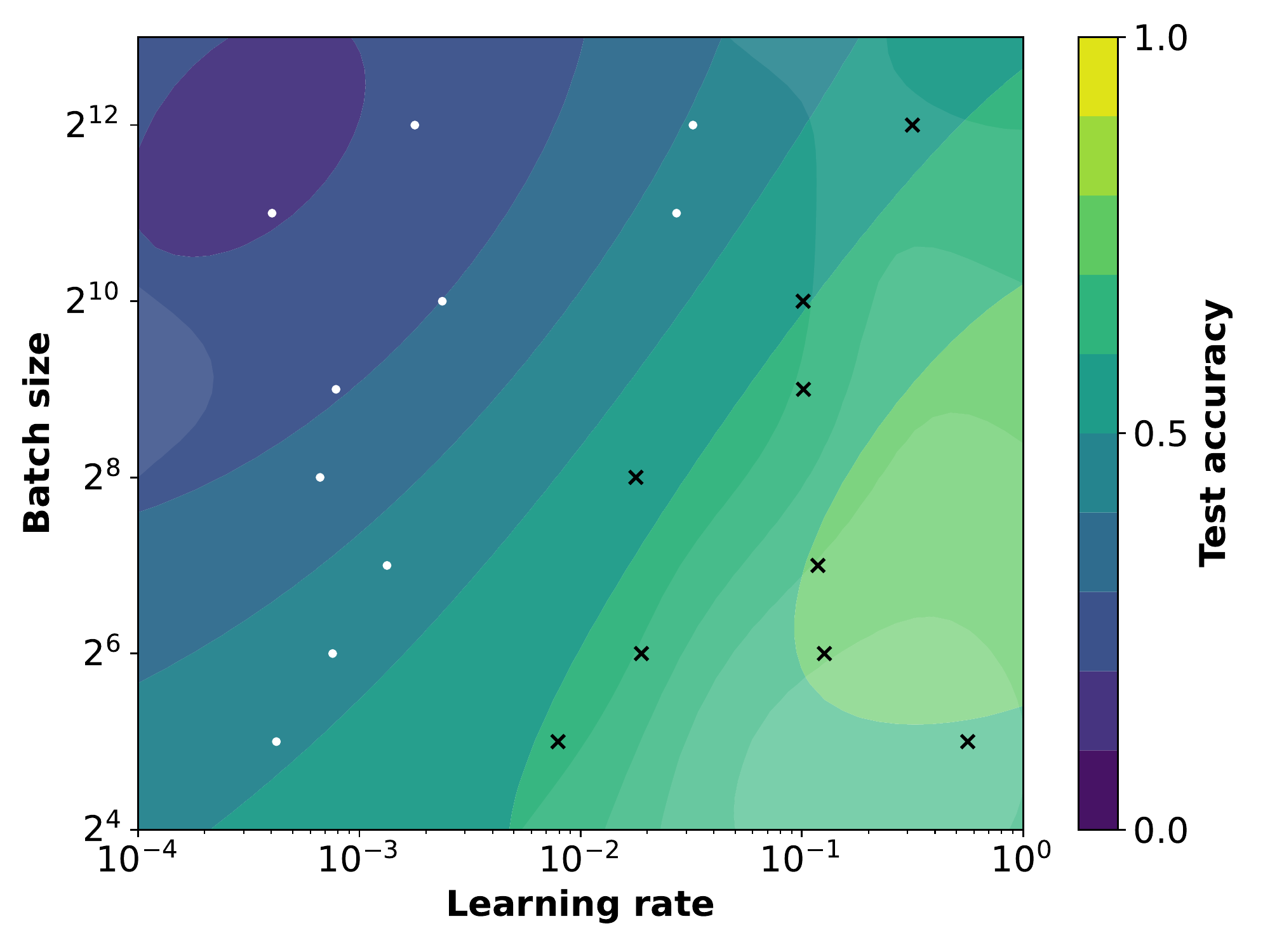}
    }
    \caption{Effect of GP kernel on predicted test accuracy over the search space of batch size, learning rate, model and dataset: \textbf{RBF kernel with ARD}.}\label{fig:batch-size-kernel-comparison-rbf-ard}
\end{figure}

\begin{figure}[t]
    \centering
    \subfloat[AlexNet on CIFAR-10]{
        \includegraphics[trim={0 0 3.5cm 0},clip,width=0.30\textwidth]{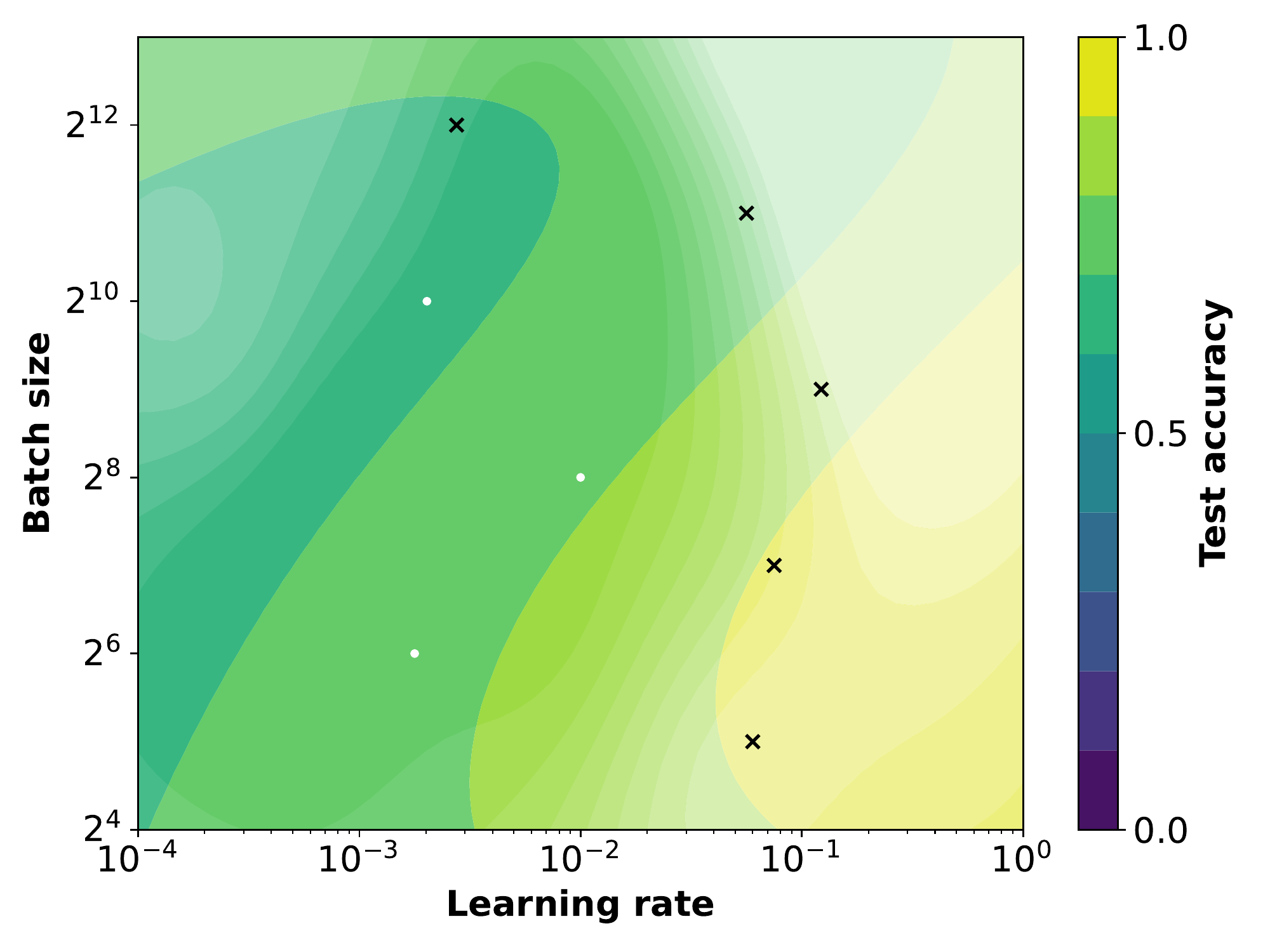}
    }
    \subfloat[ResNet on CIFAR-10]{
        \includegraphics[trim={0 0 3.5cm 0},clip,width=0.30\textwidth]{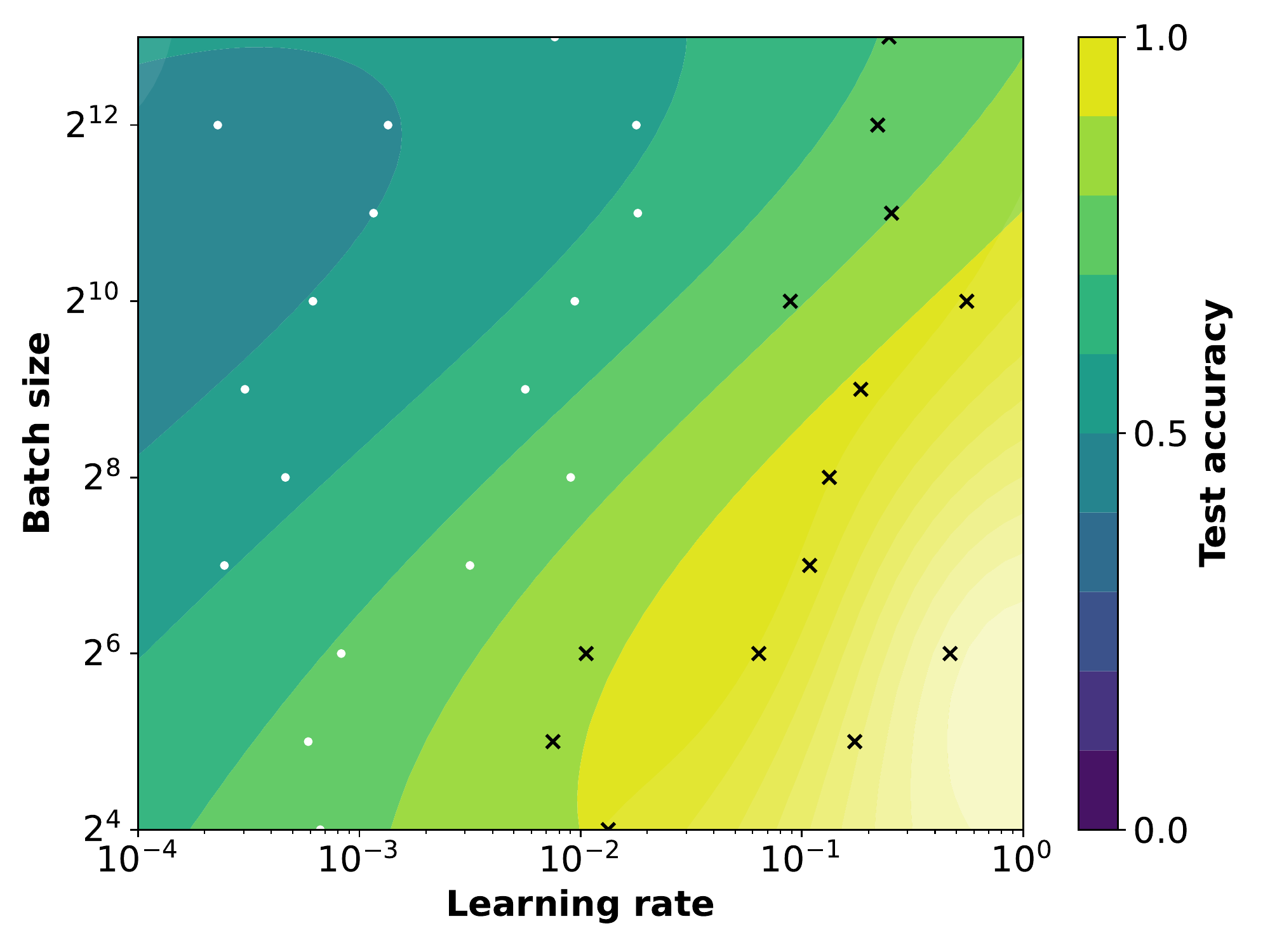}
    }
    \subfloat[VGG on CIFAR-10]{
        \includegraphics[trim={0 0 0 0},clip,width=0.3625\textwidth]{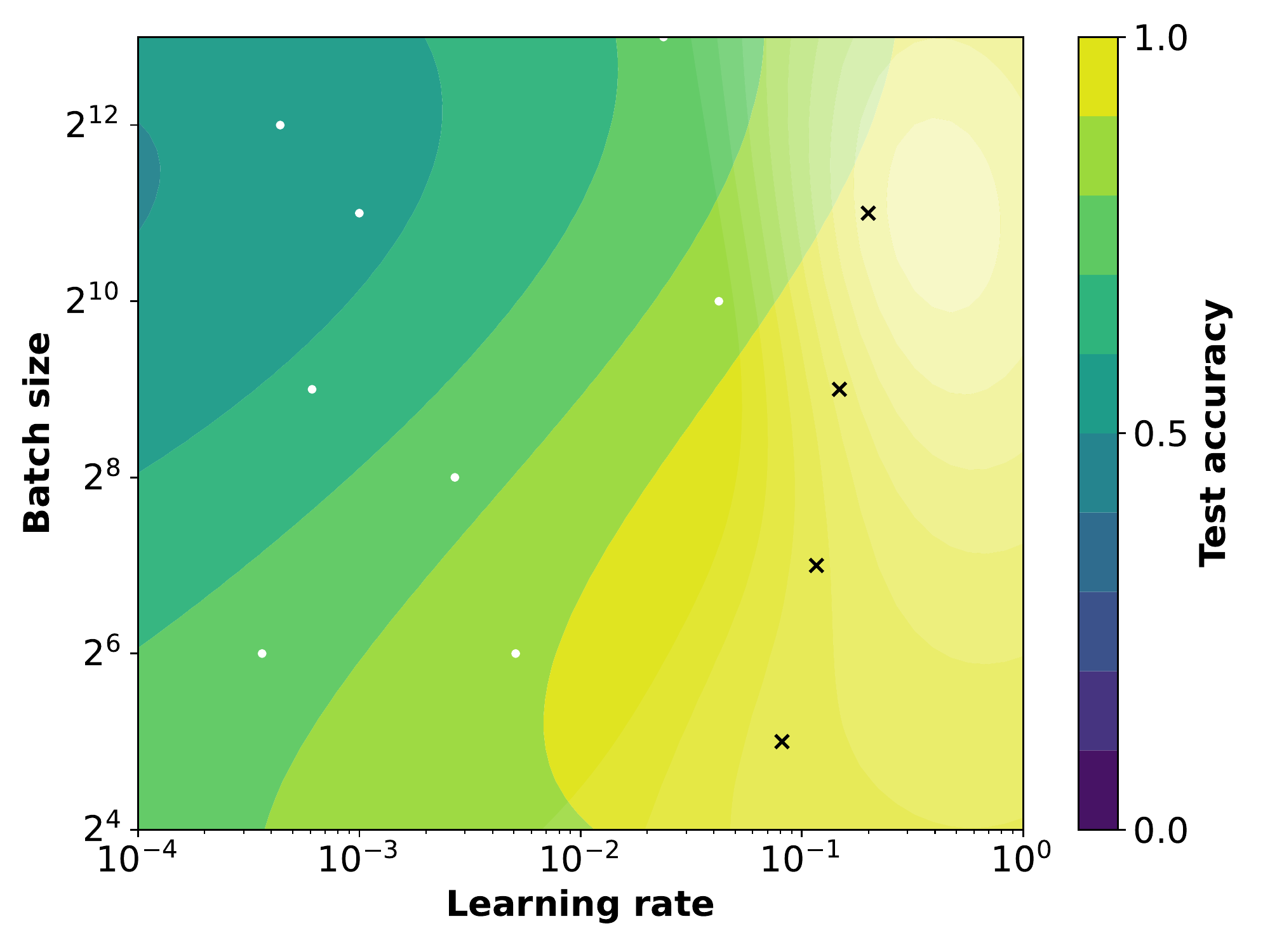}
    } \\ 
    \subfloat[AlexNet on CIFAR-100]{
        \includegraphics[trim={0 0 3.5cm 0},clip,width=0.30\textwidth]{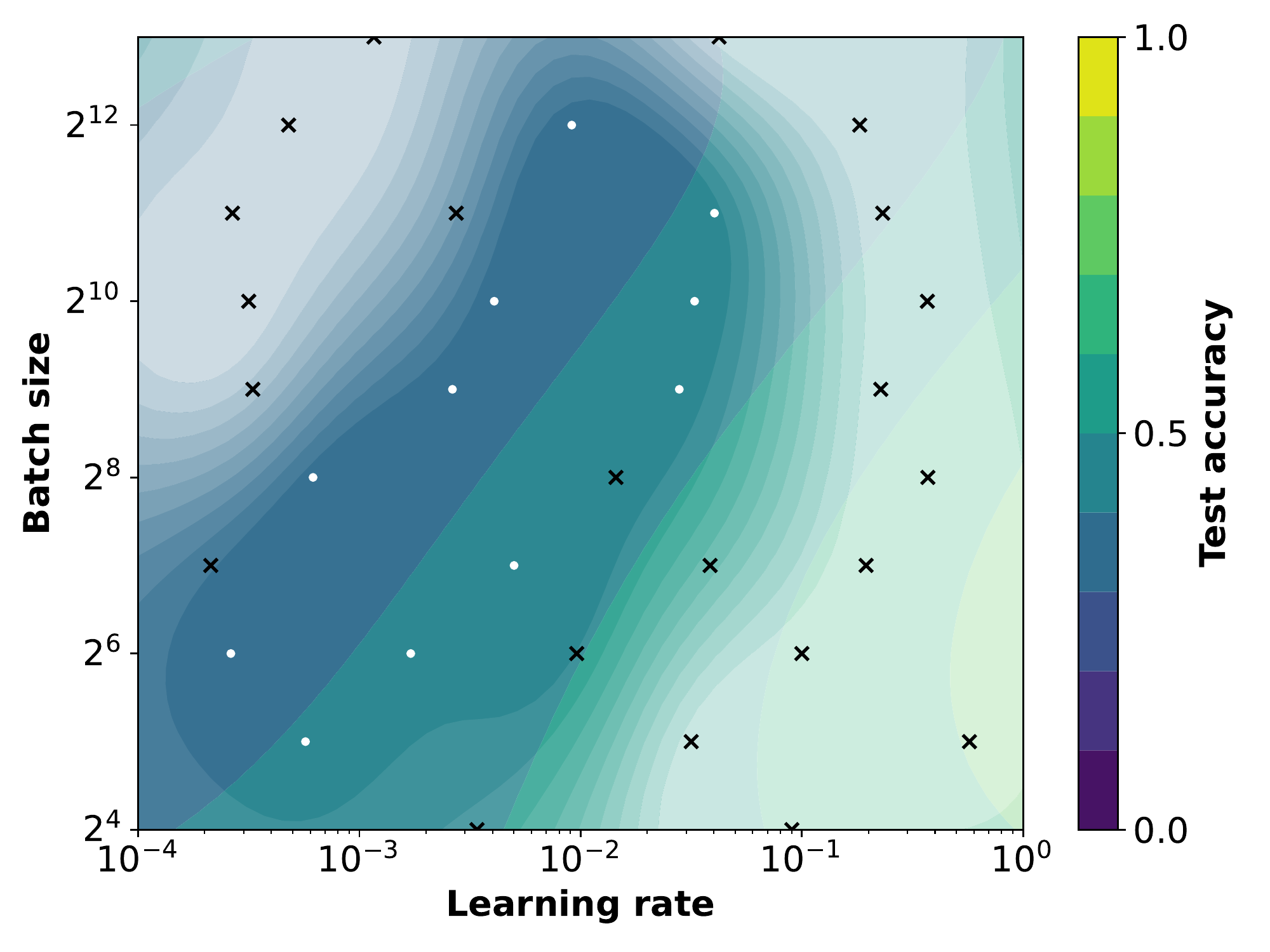}
    }
    \subfloat[ResNet on CIFAR-100]{
        \includegraphics[trim={0 0 3.5cm 0},clip,width=0.30\textwidth]{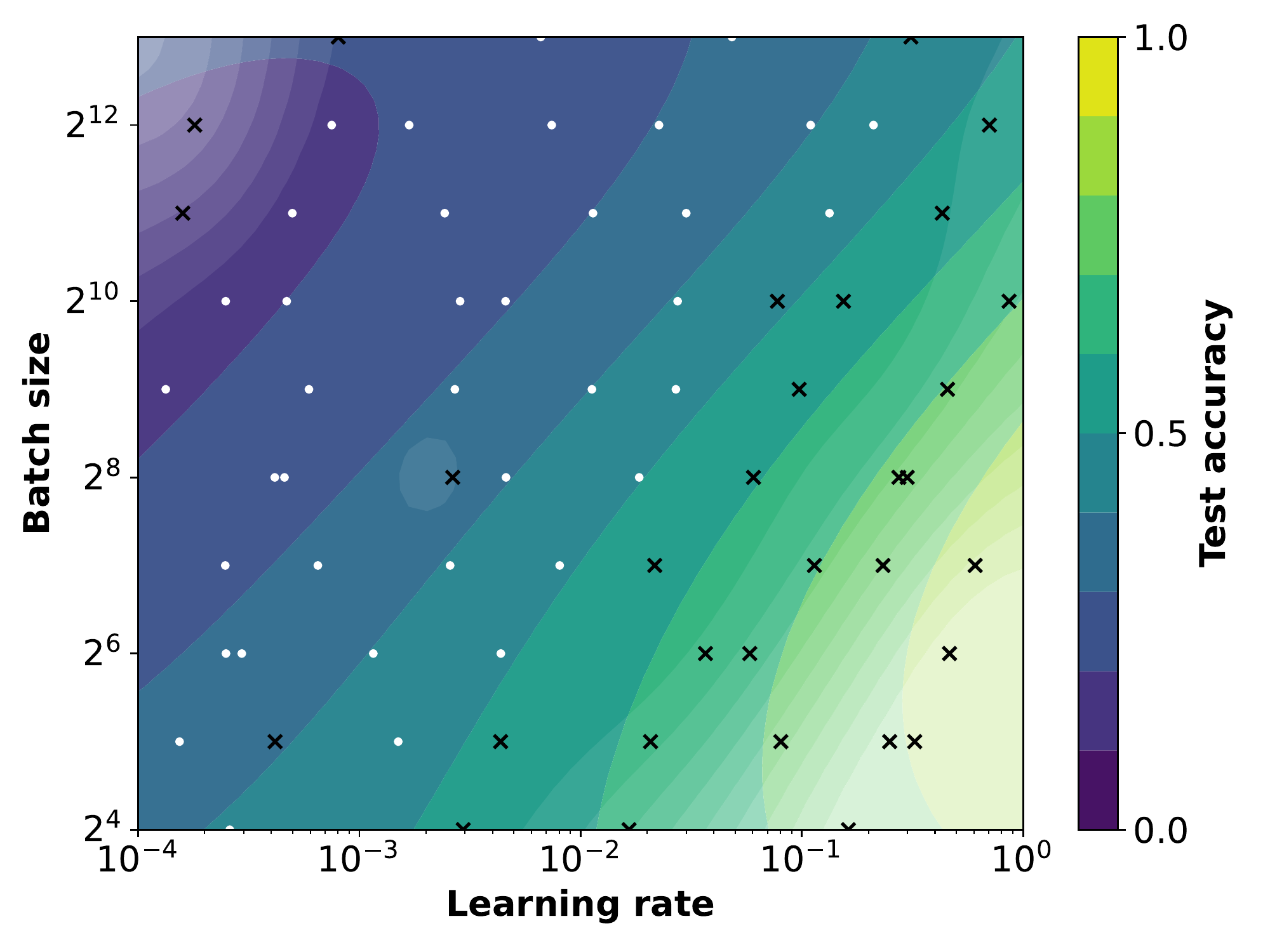}
    }
    \subfloat[VGG on CIFAR-100]{
        \includegraphics[trim={0 0 0 0},clip,width=0.3625\textwidth]{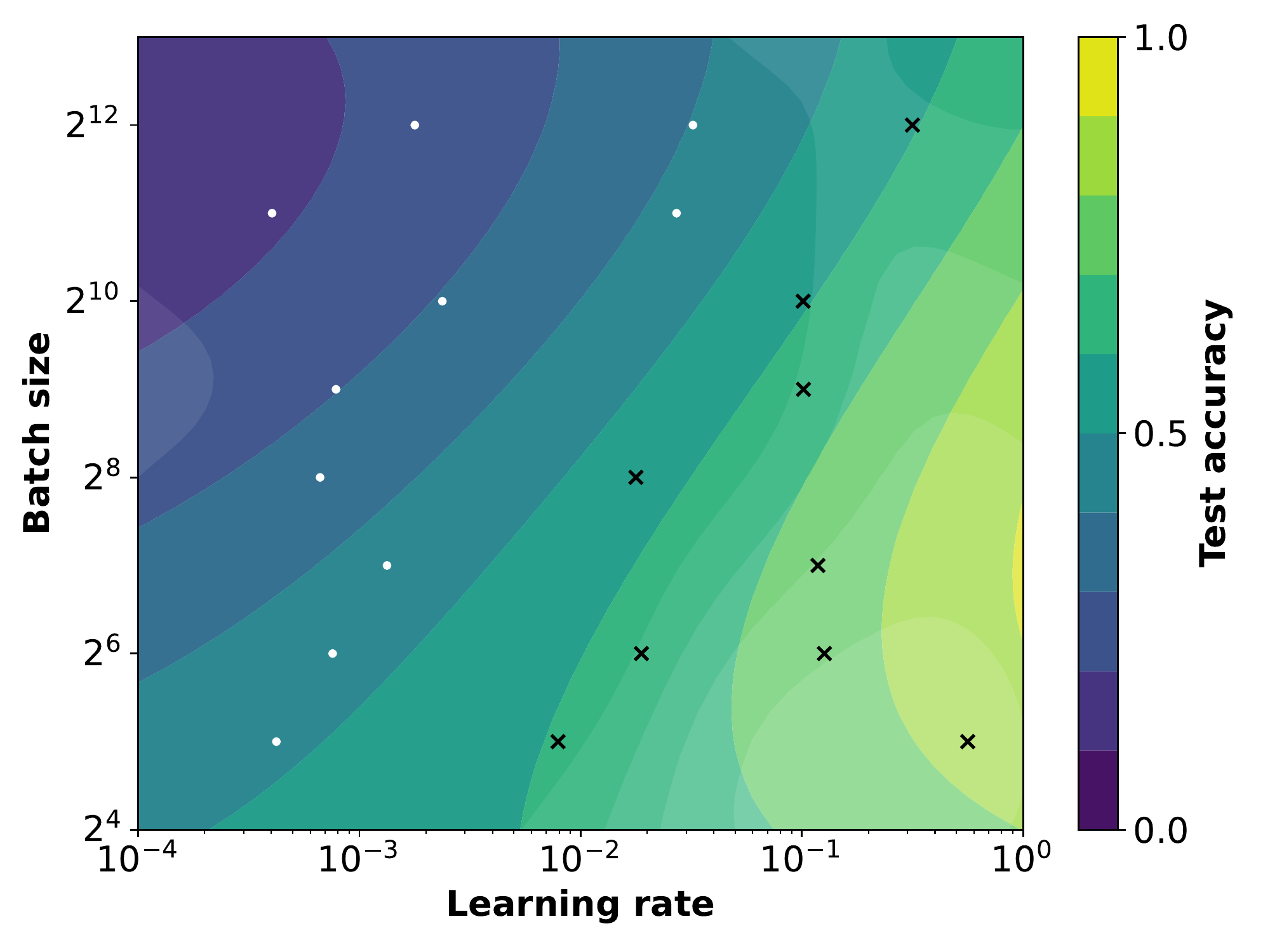}
    }
    \caption{Effect of GP kernel on predicted test accuracy over the search space of batch size, learning rate, model and dataset: \textbf{RBF kernel without ARD}.}\label{fig:batch-size-kernel-comparison-rbf-no-ard}
\end{figure}

\end{document}